%% file: main_arxiv.tex
\documentclass[11pt, a4paper, logo, copyright]{googledeepmind}

\usepackage[utf8]{inputenc} 
\usepackage[T1]{fontenc}    
\usepackage{hyperref}       
\usepackage{cleveref}
\usepackage{cite}
\usepackage{url}            
\usepackage{booktabs}       
\usepackage{amsfonts}       
\usepackage{nicefrac}       
\usepackage{microtype}      
\usepackage{xcolor}         

\input{math_commands.tex}

\usepackage{graphicx}
\usepackage{amsmath,bm}
\usepackage{amsthm}
\usepackage{enumerate}
\usepackage{wrapfig}
\usepackage{color, soul}
\usepackage{psfrag}
\usepackage{adjustbox}
\usepackage{xspace}
\usepackage{amssymb}
\usepackage{multirow}
\usepackage{afterpage}
\usepackage{colortbl}
\usepackage{float}
\usepackage{textcomp}
\usepackage{siunitx}
\usepackage{mdframed}
\usepackage[super]{nth}

\usepackage{enumitem}
\usepackage{bbding}
\usepackage{pifont}
\usepackage{bbm}
\usepackage{tcolorbox}
\usepackage{theoremref}
\usepackage{wrapfig}
\usepackage{subcaption}

\usepackage{soul}
\colorlet{soulblue}{blue!20}

\colorlet{darkgreen}{green!65!black}
\colorlet{darkblue}{blue!75!black}
\colorlet{darkred}{red!80!black}
\definecolor{yellow}{HTML}{f7c600}
\definecolor{lightblue}{HTML}{0071bc}
\definecolor{lightgreen}{HTML}{39b54a}
\definecolor{deemph}{gray}{0.55}

\definecolor{baselinecolor}{gray}{.95}
\definecolor{graycolor}{gray}{.95}

\newcommand{\tablestyle}[2]{\setlength{\tabcolsep}{#1}\renewcommand{\arraystretch}{#2}\centering\footnotesize}
\newlength\savewidth
\newcolumntype{x}[1]{>{\centering\arraybackslash}p{#1pt}}
\newcolumntype{y}[1]{>{\raggedright\arraybackslash}p{#1pt}}
\newcolumntype{z}[1]{>{\raggedleft\arraybackslash}p{#1pt}}

\newcommand{\name}{\texttt{GlucoFM}\xspace}
\newcommand{\chronos}{\texttt{Chronos-2}\xspace}
\newcommand{\chronosone}{\texttt{Chronos}\xspace}
\newcommand{\moment}{\texttt{MOMENT}\xspace}
\newcommand{\mantis}{\texttt{Mantis}\xspace}
\newcommand{\mantisv}{\texttt{MantisV2}\xspace}
\newcommand{\cgmformer}{\texttt{CGMformer}\xspace}
\newcommand{\cgmjepa}{\texttt{CGM-JEPA}\xspace}
\newcommand{\xcgm}{\texttt{X-CGM-JEPA}\xspace}
\newcommand{\gluformer}{\texttt{GluFormer}\xspace}
\newcommand{\cgmlsm}{\texttt{CGM-LSM}\xspace}

\newcommand{\wearcgm}{\texttt{Wear-CGM}\xspace}
\newcommand{\shanghai}{\texttt{ShanghaiT2DM}\xspace}
\newcommand{\stanford}{\texttt{Stanford}\xspace}
\newcommand{\bigideas}{\texttt{BIG IDEAs}\xspace}
\newcommand{\colas}{\texttt{Colas}\xspace}
\newcommand{\cgmacros}{\texttt{CGMacros}\xspace}
\newcommand{\hall}{\texttt{Hall}\xspace}

\newcommand{\hlfirst}[1]{\colorbox[HTML]{CFE2FF}{#1}}   

\newcommand{\cmark}{\textcolor{green!70!black}{\ding{51}}}
\newcommand{\xmark}{\textcolor{red!70!black}{\ding{55}}}
\newcommand{\gup}[1]{\textcolor{green!60!black}{\scriptsize~\textbf{#1}}}
\newcommand{\ggup}[1]{\textcolor{green!60!black}{\textbf{#1}}}
\newcommand{\gdown}[1]{\textcolor{red!60}{\scriptsize~#1}}
\newcommand{\pho}{\hphantom{\scriptsize~(+0.0)}}

\title{GlucoFM: A Dual-Stream Foundation Model for Continuous Glucose Monitoring}

\reportnumber{}

\author[1,2,*]{Zechen Li}
\author[1]{Keerthana Natarajan}
\author[1]{Weizhi Zhang}
\author[1]{Menglian Zhou}
\author[1]{Simon A. Lee}
\author[1]{Yuwei Zhang}
\author[1]{Maxwell A. Xu}
\author[1]{Zeinab Esmaeilpour}
\author[2]{Flora D. Salim}
\author[1]{Mark Malhotra}
\author[1]{Lindsey Sunden}
\author[1]{Shwetak Patel}
\author[1,$\dagger$]{Yuzhe Yang}
\author[1,$\dagger$]{Ahmed A. Metwally}

\affil[1]{Google Research}
\affil[2]{University of New South Wales, Sydney}
\affil[$\dagger$]{Co-last authors} 
\affil[*]{Work done during an internship at Google}

\correspondingauthor{Correspondence to: \{zechenl, yuzheyang, aametwally\}@google.com.}

\input{sections/0_abstract}

\begin{document}

\maketitle

\newenvironment{Itemize}{
    \begin{itemize}[leftmargin=*]
    \setlength{\itemsep}{0pt}
    \setlength{\topsep}{0pt}
    \setlength{\partopsep}{0pt}
    \setlength{\parskip}{1pt}}
{\end{itemize}}
\setlength{\leftmargini}{9pt}

\input{sections/1_intro}

\input{sections/2_related}
\input{sections/3_method}
\input{sections/4_exp}

\input{sections/5_ablation}

\input{sections/6_discussion}

\bibliography{ref}
\bibliographystyle{plain}

\newpage
\appendix

\input{sections/7_appendix}

\end{document}

%% file: math_commands.tex
\usepackage{amsmath,amsfonts,bm}

\def\eqref#1{equation~\ref{#1}}

\def\1{\bm{1}}

\DeclareMathAlphabet{\mathsfit}{\encodingdefault}{\sfdefault}{m}{sl}
\SetMathAlphabet{\mathsfit}{bold}{\encodingdefault}{\sfdefault}{bx}{n}



%% file: sections/0_abstract.tex
\begin{abstract}
Continuous glucose monitoring (CGM) provides a dense view of daily metabolic physiology, yet existing generic time-series and CGM-specific foundation models often encode glucose traces as entangled single-stream sequences, leaving their multiscale temporal structure only implicitly modeled. We present \name, a lightweight CGM foundation model that aligns irregular recordings to a 24-hour chronological grid, preserves observation masks, and decomposes glucose dynamics into slow-varying glycemic trend and short-term deviation streams. \name is pretrained on 109,066 hours of unlabeled CGM recordings from 477 subjects with masked contextual latent prediction over fused daily representations and temporal dynamics prediction over the two streams. Frozen pre-fusion probes confirm distinct temporal emphasis: state tokens preferentially preserve hourly glucose level, whereas event tokens better retain short-term change. Across four diverse cohorts and seven phenotype-classification tasks, \name achieves the strongest subject-disjoint linear-probing performance among evaluated baselines, improving average PR-AUC by 4.1 points over the best CGM-specific foundation model, while also supporting strong cross-dataset transfer and few-shot adaptation. Its gains are most pronounced on core metabolic outcomes, leading PR-AUC on all diabetes-risk and $\beta$-cell dysfunction tasks and on 3 of 4 insulin-resistance tasks. Beyond phenotyping, when combined with recent CGM, meal nutrition, and subject context, the same frozen encoder achieves the lowest two-hour postprandial glycemic response errors among evaluated methods for trajectory, incremental area under the curve, peak rise, and peak timing. Together, these results demonstrate that \name learns reusable frozen CGM representations spanning metabolic phenotyping and context-conditioned glucose-response prediction.
\end{abstract}

%% file: sections/1_intro.tex
\section{Introduction}
\label{intro}

\begin{figure*}[!t]
\centering
\includegraphics[width=1\textwidth]{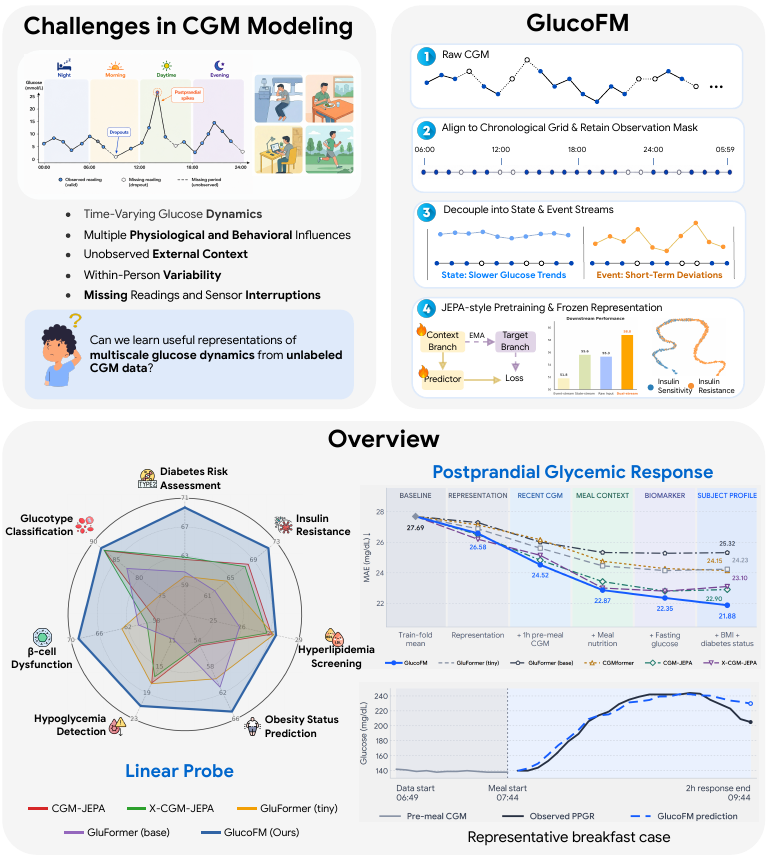} 
\caption{Overview of \name, a lightweight foundation model for continuous glucose monitoring.}
\label{fig:overview}
\end{figure*}

Continuous glucose monitoring (CGM) provides a dense window into daily metabolic physiology, reflecting fasting baselines, nocturnal patterns, postprandial excursions, and fluctuations influenced by behavior, physiology, and sensor conditions. Its use has expanded from type 1 diabetes management, including hypoglycemia prevention, time-in-range monitoring, and closed-loop insulin delivery, to type 2 diabetes, prediabetes, and normoglycemic cohorts, where it helps characterize glycemic variability, treatment response, early metabolic dysfunction, and heterogeneous glucose phenotypes \cite{metwally2025prediction,metwally2025usecontinuousglucosemonitoring, Wu2025GlycemicCarbsPhysiology, Park2025LifestyleT2DSubphenotypes, klonoff2025continuous}. These signals are increasingly used for metabolic phenotyping, including diabetes risk, insulin resistance, $\beta$-cell dysfunction, hypoglycemia, obesity, hyperlipidemia, and glucotypes~\cite{hall2018glucotypes}.

Despite this promise, high-quality clinical labels remain expensive, laborious to obtain, sparse, and cohort-specific, limiting fully supervised modeling~\cite{metwally2025prediction}. This motivates self-supervised CGM foundation models that learn reusable representations from unlabeled recordings and transfer them across metabolic prediction tasks. Beyond phenotype classification, a useful CGM representation should also compose with temporally available event and subject context to predict future glucose responses without fine-tuning the encoder. Recent time-series and CGM-specific foundation models have advanced representation learning through forecasting, autoregressive generation, masked modeling, and latent prediction~\cite{lutsker2026foundation,luo2025cgmlsm,lu2025pretrained,muhammad2026cgmjepalearningconsistentcontinuous}. However, many existing approaches still encode CGM as a single sequence representation, leaving physiological and sensing structure only implicitly modeled. CGM dynamics are inherently multi-scale: slow glycemic trends coexist with short-term deviations potentially influenced by meals, activity, stress, or sensor artifacts. Meanwhile, heterogeneous sampling densities and missingness patterns from device protocols motivate sensing-aware modeling beyond naive interpolation and raw-value reconstruction. These challenges call for a CGM foundation model that explicitly incorporates multiscale decomposition and sensing-aware design.

\input{tables/table_intro_comparison}

Motivated by the multi-scale dynamics of CGM, we propose \textbf{\name}, a lightweight self-supervised foundation model for CGM representation learning. \name is a research prototype designed solely for retrospective physiological representation learning. \name aligns irregular recordings to a 24-hour chronological grid while preserving observation masks, and uses a dual-stream state--event encoder to separately model slow glycemic trends and transient deviations before fusion. Rather than reconstructing raw glucose values, \name is pretrained with JEPA-style latent objectives~\cite{assran2023self} that combine masked contextual representation learning and temporal dynamics modeling. It further incorporates CGM-aware augmentations that simulate value perturbations, heterogeneous sampling rates, and sensor dropouts, encouraging robust daily representations that capture both global glycemic context and local temporal changes. As summarized in Table~\ref{tab:intro_comparison}, \name differs from prior TS/CGM foundation models through explicit state--event decomposition and CGM-aware augmentation.

We pretrain \name on 109,066 hours of unlabeled CGM recordings from 477 dataset-defined subjects and evaluate its frozen representations across four downstream cohorts and seven phenotype-classification tasks. For the controlled linear-probing comparison in Figure~\ref{fig:overview}, CGM-specific baselines without public checkpoints are re-pretrained on the same unlabeled CGM corpus, enabling controlled comparison of architectural and objective-level differences. \name achieves the strongest average subject-disjoint linear-probing performance among evaluated baselines and further demonstrates strong cross-dataset transfer and few-shot adaptation. Pre-fusion probes show that the two streams retain distinct slow--fast temporal content. Beyond phenotyping, the same frozen encoder, when combined with recent CGM, meal nutrition, and subject context, achieves the lowest two-hour postprandial glycemic response (PPGR) errors among evaluated methods for trajectory, incremental area under the curve (iAUC), peak rise, and peak timing.

Our contributions are threefold:
\begin{itemize}

    \item \textbf{Lightweight daily CGM foundation model.} \name learns daily glucose representations from unlabeled CGM while preserving observation masks and absolute time-of-day structure.

    \item \textbf{Multiscale and sensing-aware representation learning.} \name combines state--event dual-stream modeling, CGM-aware augmentation, masked contextual latent prediction, and temporal dynamics modeling to encode complementary glycemic dynamics.

    \item \textbf{Broad downstream utility.} Across four CGM cohorts and seven phenotype tasks, \name shows strong linear probing, cross-dataset transfer, few-shot adaptation, and multiday prediction. Combined with recent CGM, meal nutrition, and subject context, the same frozen encoder further supports two-hour PPGR prediction.

\end{itemize}

%% file: tables/table_intro_comparison.tex
\begin{table}[t]
\centering
\caption{Conceptual comparison between \name and representative TS/CGM foundation models. Checkmarks indicate components explicitly designed in the original model. \textit{Decomp.} denotes decomposition, and \textit{Aug.} denotes augmentation.}
\label{tab:intro_comparison}
\resizebox{\textwidth}{!}{
\begin{tabular}{r l c c c c}
\toprule
\textbf{Model} 
& \textbf{Pretraining Objective}
& \textbf{CGM-Specific}
& \textbf{Daily Modeling}
& \textbf{Signal Decomp.}
& \textbf{CGM-aware Aug.} \\
\midrule

\chronosone~\cite{ansari2024chronos}
& Quantized autoregressive forecasting
& \xmark & \xmark & \xmark & \xmark \\

\moment~\cite{goswami2024moment}
& Masked time-series modeling
& \xmark & \xmark & \xmark & \xmark \\

\mantis~\cite{feofanov2025mantislightweightcalibratedfoundation}
& Contrastive representation learning
& \xmark & \xmark & \xmark & \xmark \\

\midrule

\gluformer~\cite{lutsker2026foundation}
& Next-token prediction
& \cmark & \xmark & \xmark & \xmark \\

\cgmlsm~\cite{luo2025cgmlsm}
& Next-token prediction
& \cmark & \xmark & \xmark & \xmark \\

\cgmformer~\cite{lu2025pretrained}
& Masked token prediction
& \cmark & \cmark & \xmark & \xmark \\

\cgmjepa~\cite{muhammad2026cgmjepalearningconsistentcontinuous}
& Masked latent prediction
& \cmark & \cmark & \xmark & \xmark \\

\midrule
\rowcolor{gray!10}
\textbf{\name (ours)}
& \textbf{Contextual JEPA + Dynamics}
& \cmark & \cmark & \cmark & \cmark \\
\bottomrule
\end{tabular}
}
\end{table}

%% file: sections/2_related.tex
\section{Related Work}
\label{sec:related-work}

\textbf{Self-Supervised and Foundation Models for Time Series.}
Self-supervised time-series learning commonly uses contrastive learning, masked modeling, forecasting, or latent prediction~\cite{tonekaboni2021unsupervisedrepresentationlearningtime,kiyasseh2021clocs,yang2023simper,yu2024smart}. Recent time-series foundation models scale these ideas to heterogeneous corpora, including masked encoders such as \moment~\cite{goswami2024moment}, tokenized or decoder-only forecasters such as \chronos~\cite{ansari2024chronos,ansari2025chronos2} and TimesFM~\cite{Das2024timefm}, probabilistic forecasters such as Lag-Llama~\cite{rasul2024lagllama}, and multi-task models such as Moirai~\cite{woo2024moirai} and UniTS~\cite{gao2024building}. While these models provide strong general-purpose backbones, they typically treat time series as generic numerical sequences and are optimized for forecasting, reconstruction, or broad task transfer. \name instead targets CGM-specific daily representation learning, where temporal alignment, sensing irregularity, and glycemic structure are central.

\textbf{Clinical and Physiological Representation Learning.}
Clinical representation learning has produced pretrained models for longitudinal EHRs~\cite{li2019behrttransformerelectronichealth,rasmy2021medbert,landi2020deep,shmatko2025learning,jing2026one}, sleep physiology~\cite{xu2026sleeplm,shuai2026osf}, wearable sensing~\cite{xu2025lsm2learningincompletewearable,zhang2025sensorlmlearninglanguagewearable,li-etal-2025-sensorllm,narayanswamy2025scaling,lee2025himae,zhou2026physiology,li2026zara}, and broader physiological signals~\cite{chen2025comodo,10.1145/3450439.3451863,abbaspourazad2024largescale,li2026hearts}. These works demonstrate the value of domain-specific pretraining under label scarcity, patient heterogeneity, and noisy real-world measurement. However, CGM has distinct temporal structure: glucose traces combine slow basal regulation, circadian variation, and short-term excursions potentially influenced by meals, insulin, stress, activity, or sensor artifacts over hours to days~\cite{wagner2012continuous}. \name focuses on this glucose-specific structure by learning daily representations directly from CGM rather than relying on generic EHR, sleep, cardiovascular, or wearable encoders.

\textbf{CGM Representation Learning for Metabolic Phenotyping.}
CGM has been used to study metabolic heterogeneity through postprandial response prediction, glucotype discovery, cohort profiling, and prediction of insulin resistance, $\beta$-cell dysfunction, and metabolic risk~\cite{metwally2025prediction,klonoff2025continuous,hall2018glucotypes,metwally2026insulin,matabuena2025glucodensity,zeevi2015personalized,berry2020human,keshet2023cgmap,sugimoto2026use}. Much of this work relies on handcrafted metrics, clustering, supervised predictors, or study-specific protocols, limiting reusable representation learning across cohorts and tasks. Recent CGM foundation models, such as \cgmformer~\cite{lu2025pretrained}, \cgmlsm~\cite{luo2025cgmlsm}, \cgmjepa~\cite{muhammad2026cgmjepalearningconsistentcontinuous}, and \gluformer~\cite{lutsker2026foundation}, advance masked modeling, autoregressive forecasting, finetuning, or cohort-level prediction. In contrast, \name learns a compact frozen CGM encoder for subject-disjoint clinical prediction, cross-dataset transfer, and low-label adaptation.

\textbf{Missingness-Aware CGM Modeling.}
Missingness is central in clinical time-series modeling and is commonly addressed through imputation, mask-aware architectures, or continuous-time methods~\cite{che2018recurrent,cao2018brits,rubanova2019latent,shukla2019interpolation}. In CGM, gaps can arise from sensor warm-up, removal, dropout, calibration issues, or nonwear, and naive imputation may obscure meaningful glucose dynamics. \name therefore uses a mask-aware daily grid that preserves chronological structure while distinguishing observed measurements from missing positions, supporting representation learning under real-world CGM irregularity.

%% file: sections/3_method.tex
\section{Methodology}
\label{sec:method}

\begin{figure*}[!t]
\centering
\includegraphics[width=1\textwidth]{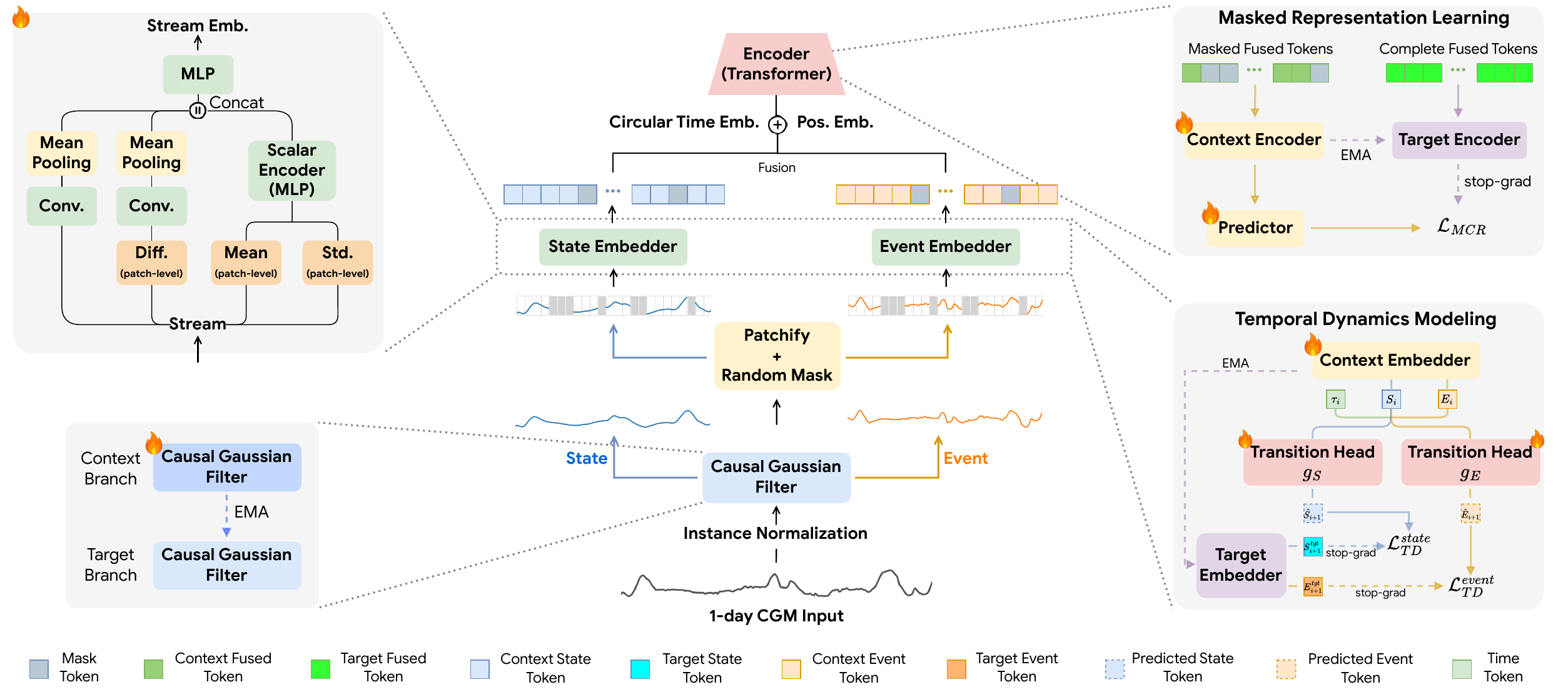} 
\caption{Model framework and pretraining objectives of \name.}
\label{fig:pretrain}
\vspace{-0.5em}
\end{figure*}

\subsection{Modeling Input CGM Data Streams}
\textbf{Chronological Grid Alignment.}
Given CGM readings $X=\{x_i\}_{i=1}^{N}$ with timestamps $T=\{t_i\}_{i=1}^{N}$, our goal is to learn representations that preserve both glucose dynamics and absolute time-of-day structure. Since CGM recordings can be irregular due to heterogeneous sampling rates and sensor dropouts, we align each recording segment to a fixed 24-hour chronological grid with $\Delta t=5$ minutes and $L=288$ positions. The first timestamp defines the absolute circadian start index, $s = \lfloor (60\cdot \mathrm{hour}(t_1) + \mathrm{minute}(t_1)) / \Delta t \rfloor$. 
The aligned input consists of a glucose sequence $\hat{X}\in\mathbb{R}^{L}$ and an observation mask $M\in\{0,1\}^{L}$, where $M_j=1$ only for physically observed measurements. Missing positions are filled only for tensor construction, while $M$ is preserved throughout representation learning. This separates chronological alignment from interpolation and allows \name to handle different sampling rates and sensor dropouts on a common daily grid.

\textbf{CGM-aware Data Augmentation.}
We use two families of on-the-fly augmentations during pretraining. \textit{Value perturbations} simulate realistic CGM signal variation, including low-frequency baseline wander and short compression-like drops, while preserving the observation mask. \textit{Structural sparsification} alters the observation pattern itself by randomly decimating dense 5-minute recordings into 15-minute-like sampling and masking short contiguous disconnection blocks. These augmentations expose \name to sensor drift, compression artifacts, heterogeneous sampling rates, and physical dropouts, while randomized ordering and decayed co-occurrence probabilities prevent over-corrupting individual sequences.

\subsection{State--Event Dual-Stream Encoder}

\textbf{Mask-aware Signal Decomposition.}
CGM signals naturally contain both low-frequency metabolic states and high-frequency transient events. 
Low-frequency components reflect stable glycemic baselines and slower regulatory patterns, while high-frequency components represent short-term deviations from the local trend, which may arise from transient physiology and behavior. To avoid encoding these heterogeneous dynamics as a single entangled sequence, \name decouples the aligned signal into state and event components before masked representation learning, as shown in Figure~\ref{fig:pretrain}.

Given the aligned sequence $\hat{X}$ and observation mask $M$, we first apply mask-aware normalization and compute patch-level statistics, including glucose summaries and rate-of-change features. We then estimate the low-frequency state using a learnable causal Gaussian filter. The filter bandwidth $\sigma$ is initialized at $\sigma=6.0$ and constrained within $[2.0,12.0]$ grid steps, corresponding to approximately 10--60 minutes on the 5-minute grid. This range allows the filter to separate short-term fluctuations from slower glycemic trends while adapting its smoothing scale during pretraining. Causality is enforced by using only current and past observations when estimating the local trend, preventing future glucose values from leaking into the representation. The filter is also mask-aware, normalizing only over valid observations to avoid dropout-induced boundary artifacts.

The filtered trend defines the state component, while the residual captures high-frequency event-level deviations:
\begin{equation}
\hat{X}_{\mathrm{state}} = \mathrm{Filter}(\hat{X}, M),
\qquad
\hat{X}_{\mathrm{event}} = (\hat{X} - \hat{X}_{\mathrm{state}}) \odot M .
\end{equation}
State and event signals are encoded by two parallel streams and fused into unified patch tokens.

\textbf{Patch Tokenization and Circadian Encoding.}
After state--event decomposition, \name tokenizes each sequence into 24 non-overlapping temporal patches, each covering 12 grid steps, corresponding to one hour. The state stream encodes the filtered trend together with mask-aware state statistics, while the event stream encodes residual deviations together with rate-of-change statistics. The two stream outputs are then fused into unified physiological patch tokens. To preserve absolute time-of-day information, each token is augmented with circular time features,
$\tau(i)=\left[\sin(2\pi i / L), \cos(2\pi i / L)\right]$,
where $i$ is the absolute grid index and $L=288$. This encoding allows the context encoder to model glucose dynamics with awareness of circadian phase.

\subsection{Latent Predictive Pretraining}
\name uses two complementary JEPA-style objectives to predict latent representations rather than reconstruct potentially interpolated or noisy glucose values.

\textbf{Masked Contextual Representation Learning.}
In our first pretraining objective, we randomly mask temporal patches in the online branch, with the masking ratio sampled from $[0.5,0.6]$. In the online branch, each selected patch is replaced by a learnable mask token, and the resulting full patch grid is encoded by a context encoder. In parallel, the full physically observed sequence is encoded by an EMA target branch to provide stable latent targets. The target branch is initialized from the online branch and updated as $\theta_{\mathrm{tgt}} \leftarrow m\theta_{\mathrm{tgt}} + (1-m)\theta_{\mathrm{online}}$, with $m$ cosine-scheduled from $0.997$ toward $1.0$. A lightweight predictor maps the context-branch tokens to the target latent space and predicts the representations of masked patches. Let $Z_i^{\mathrm{pred}}$ and $Z_i^{\mathrm{tgt}}$ denote the predicted and target latent tokens at patch $i$. The masked contextual representation loss is:
\begin{equation}
\mathcal{L}_{\mathrm{MCR}}
=
\frac{
\sum_{i=1}^{P}
w_i m_i^{\mathrm{mask}}
\,
\mathrm{SmoothL1}
\left(
Z_i^{\mathrm{pred}},
Z_i^{\mathrm{tgt}}
\right)
}{
\sum_{i=1}^{P}
w_i m_i^{\mathrm{mask}}
+
\epsilon
},
\end{equation}
where $P$ is the number of temporal patches, $m_i^{\mathrm{mask}}$ indicates whether patch $i$ is masked, and $w_i$ is the fraction of physically observed measurements within the patch.

\textbf{Temporal Dynamics Modeling.}
Masked contextual prediction learns daily contextual representations, but CGM phenotypes also depend on how glucose dynamics evolve over time. We therefore introduce a second JEPA-style objective that predicts next-patch state and event representations. Given context-branch state and event tokens $(S_i,E_i)$ at patch $i$, two lightweight transition heads $g_S$ and $g_E$ predict the next-patch state and event targets:
\begin{equation}
\hat{S}_{i+1}=S_i+g_S([S_i,E_i,\tau_i]),
\qquad
\hat{E}_{i+1}=E_i+g_E([E_i,S_i,\tau_i]),
\end{equation}
where $\tau_i$ is the temporal position embedding formed from patch position and gated circular time-of-day features. The residual update form encourages the heads to model temporal changes rather than directly copying the current representation.
The targets $(S_{i+1}^{\mathrm{tgt}},E_{i+1}^{\mathrm{tgt}})$ are taken from the EMA target branch before global Transformer self-attention, so the next-patch targets do not already encode information from later patches.
The temporal dynamics loss is:
\begin{equation}
\mathcal{L}_{\mathrm{TD}}
=
\frac{1}{2}
\left(
\frac{
\sum_{i=1}^{P-1}
q_i
\,
\mathrm{SmoothL1}
\left(
\hat{S}_{i+1},
S_{i+1}^{\mathrm{tgt}}
\right)
}{
\sum_{i=1}^{P-1} q_i+\epsilon
}
+
\frac{
\sum_{i=1}^{P-1}
q_i
\,
\mathrm{SmoothL1}
\left(
\hat{E}_{i+1},
E_{i+1}^{\mathrm{tgt}}
\right)
}{
\sum_{i=1}^{P-1} q_i+\epsilon
}
\right),
\end{equation}
where $q_i$ weights each transition by the observed support of adjacent patches and excludes patches hidden from the context branch. This objective encourages the state stream to encode gradual baseline shifts and the event stream to encode transient deviations, providing an explicit temporal constraint on the decoupled representations.

\textbf{Overall Objective.}
The final pretraining loss is:
\begin{equation}
\mathcal{L}
=
\lambda_{\mathrm{MCR}}\mathcal{L}_{\mathrm{MCR}}
+
\lambda_{\mathrm{TD}}\mathcal{L}_{\mathrm{TD}} .
\end{equation}
We set $\lambda_{\mathrm{MCR}}=\lambda_{\mathrm{TD}}=1.0$. \name uses compact $3$-layer Transformer context and EMA target encoders with hidden dimension $128$, $4$ attention heads, and feed-forward dimension $256$, together with a lightweight $1$-layer Transformer predictor. During downstream evaluation, we discard the EMA target branch and retain only the frozen online encoder. \name has 0.72M trainable parameters and 1.18M total parameters during pretraining, with the difference mainly due to the EMA target branch. Additional implementation details are provided in Appendix~\ref{apd:implementation}.

%% file: sections/4_exp.tex
\section{Experiments}
\label{subsec:exp}

\subsection{Experimental Setup}

\input{tables/wraptable_dataset}
\textbf{Datasets.} Table~\ref{tab:pretrain_data} summarizes the datasets used in this study. \shanghai counts denote recording-visit subject-entries. We use dataset-defined group separation between pretraining and downstream evaluation, ensuring no overlap between pretraining subjects and downstream test groups. For self-supervised pretraining, we aggregate five CGM cohorts spanning 5-minute and 15-minute sampling rates, totaling 477 dataset-defined subjects and 109,066 hours of recordings. This mixture exposes \name to diverse glucose dynamics, sensor densities, and population characteristics. For downstream evaluation, we use four CGM cohorts with 203 dataset-defined participants and 71,669 hours of CGM data, covering seven phenotype-classification tasks: diabetes risk assessment, insulin resistance, $\beta$-cell dysfunction, glucotype, hyperlipidemia, hypoglycemia, and obesity classification. All downstream evaluations use subject-disjoint splits to prevent overlap between training and test subjects, enabling robust assessment across heterogeneous CGM cohorts. Additional dataset details are provided in Appendix~\ref{apd:dataset}.

\textbf{Baselines.}
We compare \name with \textit{three groups of baselines}: general-purpose time-series foundation models, a CGM-specific open-weight model pretrained on a separate cohort, and CGM-specific models retrained on our pretraining corpus. Implementation details are provided in Appendix~\ref{apd:baseline}. General-purpose baselines include \chronos~\cite{ansari2025chronos2}, \moment~\cite{goswami2024moment}, \mantis~\cite{feofanov2025mantislightweightcalibratedfoundation}, and \mantisv~\cite{feofanov2026mantisv2closingzeroshotgap}. Controlled CGM-specific comparisons use study-retrained implementations of \cgmjepa and \xcgm~\cite{muhammad2026cgmjepalearningconsistentcontinuous}, and \gluformer~\cite{lutsker2026foundation}, because official pretrained checkpoints were unavailable at the time of evaluation. These implementations follow the published methods and are trained on the same corpus as \name. We use \gluformer as the representative autoregressive CGM baseline because \cgmlsm~\cite{luo2025cgmlsm} has a closely related architecture and next-token objective. We also evaluate \cgmformer~\cite{lu2025pretrained}, an open-weight CGM-specific model pretrained on an external 964-participant cohort with non-public raw data.

\input{tables/table_linear_probe}
\textbf{Setup.}
We report ROC-AUC, PR-AUC, and Macro-F1. ROC-AUC measures global discriminative ability, PR-AUC emphasizes performance under class imbalance, and Macro-F1 evaluates class-balanced decision quality across phenotype classes. \name is pretrained for 120 epochs on a single NVIDIA H100 GPU with batch size 128, learning rate $10^{-4}$, weight decay $10^{-2}$, and a separate learning rate of $10^{-3}$ for the learnable Gaussian bandwidth. Detailed model training setup is provided in Appendix~\ref{apd:implementation}.

\subsection{Experimental Results}
We evaluate \name along four primary downstream axes that reflect different clinically relevant utility scenarios: subject-disjoint linear probing, postprandial glycemic response forecasting, few-shot adaptation, and cross-dataset transferability. Additional setup details and full results with fold-level variance are provided in Appendix~\ref{apd:full_results}.

\textbf{Subject-Disjoint Linear Probing.}
We first evaluate whether \name learns phenotype-informative 24-hour representations that generalize to unseen subjects while capturing day-to-day variability. For each method, we freeze the encoder, extract embeddings from non-overlapping 24-hour windows, and train the same logistic regression classifier on the extracted representations. We use subject-grouped cross-validation, ensuring that all windows from the same subject appear only in either the training or test fold. Each held-out window is predicted independently, with evaluation performed over all held-out test windows. As shown in Table~\ref{tab:downstream_results}, \name achieves the strongest overall performance, obtaining the best task-averaged PR-AUC, ROC-AUC, and Macro-F1, while ranking within the top two on 11/14, 11/14, and 9/14 task--dataset evaluations, respectively. Across the 14 task--dataset evaluations, the paired task-averaged gains over \cgmjepa are $+4.11$ points for PR-AUC (95\% CI $[2.40,5.81]$), $+4.34$ for ROC-AUC ($[2.26,6.41]$), and $+2.57$ for Macro-F1 ($[0.91,4.23]$). Against \cgmformer, the corresponding gains are $+4.74$ ($[2.47,7.02]$), $+4.67$ ($[2.02,7.32]$), and $+2.34$ points ($[0.55,4.14]$); all six Nadeau--Bengio-corrected CIs~\cite{NIPS1999_7d12b66d} exclude zero and remain significant after Holm adjustment (adjusted $p\leq0.012$; Appendix~\ref{app:linear_probe_details}).

The gains are most evident on core metabolic phenotyping outcomes. In PR-AUC, \name leads all diabetes-risk and $\beta$-cell dysfunction evaluations, and 3/4 insulin-resistance evaluations, suggesting that its frozen representations capture clinically relevant glycemic structure rather than only generic temporal patterns. \name also outperforms \cgmformer in PR-AUC on 11/14 evaluations despite \cgmformer being pretrained on a larger external 964-participant cohort, and exceeds CGM-specific baselines re-pretrained on the same corpus on 11/14 evaluations. This suggests that the improvements are not solely explained by pretraining scale, but are consistent with the benefit of CGM-aware pretraining objectives and temporal inductive biases. The main exception is \shanghai, which may reflect a device- and sampling-rate shift: it is Libre-derived with 15-minute sampling, while our pretraining corpus contains relatively fewer real Libre/15-minute recordings. Overall, \name provides the strongest frozen representation across the benchmark, with particularly clear advantages on clinically central metabolic tasks.

\textbf{Postprandial Glycemic Response Forecasting.}
We next evaluate whether frozen CGM representations support prediction of glucose dynamics following a meal. We use paired Dexcom and Libre recordings from \cgmacros, retaining 874 matched meal events per sensor from 34 participants with no additional logged meal during the two-hour prediction window. For each event, a strictly causal 24-hour pre-meal CGM window is encoded into a frozen representation, and the task predicts the subsequent two-hour glucose change relative to the observed meal-start glucose. Dexcom and Libre are modeled separately using the same MLP with two hidden layers at their native target cadences: 24 five-minute outputs for Dexcom and 8 fifteen-minute outputs for Libre. The representation is combined cumulatively with one hour of pre-meal CGM, meal nutrition, fasting glucose, and BMI and diabetes status. All encoders remain frozen and are evaluated with five subject-disjoint folds and 10 head initializations. We report MAE for the trajectory, positive incremental area under the curve (iAUC), peak rise, and peak time, computing each endpoint separately by sensor and then averaging Dexcom and Libre with equal weight.

Table~\ref{tab:ppgr_multimetric_progression} shows how each source of context changes PPGR performance. Once recent CGM is included, \name achieves the lowest trajectory error at every enriched stage, decreasing from 26.58 to 21.88\,mg/dL across the full progression. With complete context, \name achieves the lowest error on all four endpoints for both Dexcom and Libre. In the equally weighted two-sensor mean, \name reduces trajectory, positive-iAUC, peak-rise, and peak-time MAE by 4.5\%, 5.8\%, 2.2\%, and 3.4\%, respectively, relative to \cgmjepa, the strongest comparator across all four endpoints. These gains show that \name's frozen representation composes effectively with recent glucose, meal nutrition, and subject information. Sensor-stratified results are provided in Appendix~\ref{app:ppgr_details}.
\input{tables/table_ppgr}

\textbf{Few-Shot Adaptation.}
We evaluate \name under two low-label adaptation settings. In \textit{limited-subject adaptation}, we restrict the number of labeled support subjects per class within the training fold. In \textit{limited-observation adaptation}, we retain all training subjects but use only a fraction of each subject's 24-hour windows for classifier training. We compare against strong baselines, including \cgmjepa, \xcgm, \gluformer (tiny), \cgmformer, and \mantisv. We use 5-fold subject-grouped cross-validation with 10 repeated iterations and 5 random support samplings per split.

\input{tables/table_cross_dataset}

\begin{figure}[!t]
    \centering
    \includegraphics[width=\textwidth]{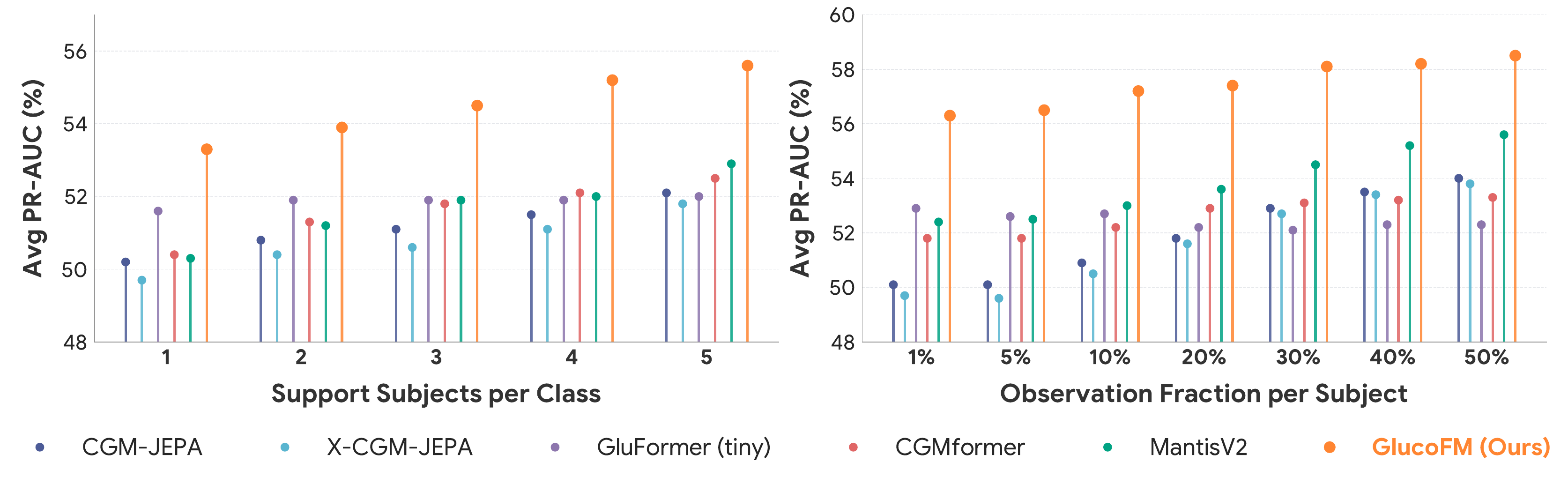}
    \caption{Few-shot adaptation under limited labeled subjects (left) and limited per-subject observations (right), reported as task-averaged PR-AUC.}
    \label{fig:few_shot}
\end{figure}

Figure~\ref{fig:few_shot} shows that \name achieves the highest task-averaged PR-AUC at every support-subject count and observation fraction. Performance improves as additional support subjects are introduced, while the consistent lead with only one or two subjects per class demonstrates effective adaptation from very few labeled individuals. With all training subjects retained, \name also remains strongest across observation fractions from 1\% to 50\% and improves as additional observations are included, demonstrating effective adaptation even with sparse per-subject data. Together, these results establish \name as a label-efficient representation under both limited-subject and limited-observation supervision.

\textbf{Cross-Dataset Generalization.}
We evaluate cohort transfer by training a linear probe on one source dataset and testing it directly on another using frozen representations. We focus on diabetes risk assessment and insulin resistance, which are shared across \cgmacros, \stanford, and \hall. To isolate representation transferability, we compare with CGM-specific baselines pretrained on the same CGM corpus and use the target dataset only for final evaluation.

As shown in Table~\ref{tab:cross_dataset}, \name achieves the best overall transfer performance, ranking first on 21 of 24 PR-AUC and ROC-AUC evaluations. The gains are strongest for transfers involving \hall and for \cgmacros$\rightarrow$\stanford diabetes-risk assessment, indicating that \name captures metabolic structure that transfers across cohorts and devices. The few non-leading cases mainly occur for \stanford$\rightarrow$\cgmacros insulin resistance, where \name remains close to the best baseline.

\begin{figure}[!t]
    \centering
    \includegraphics[width=\textwidth]{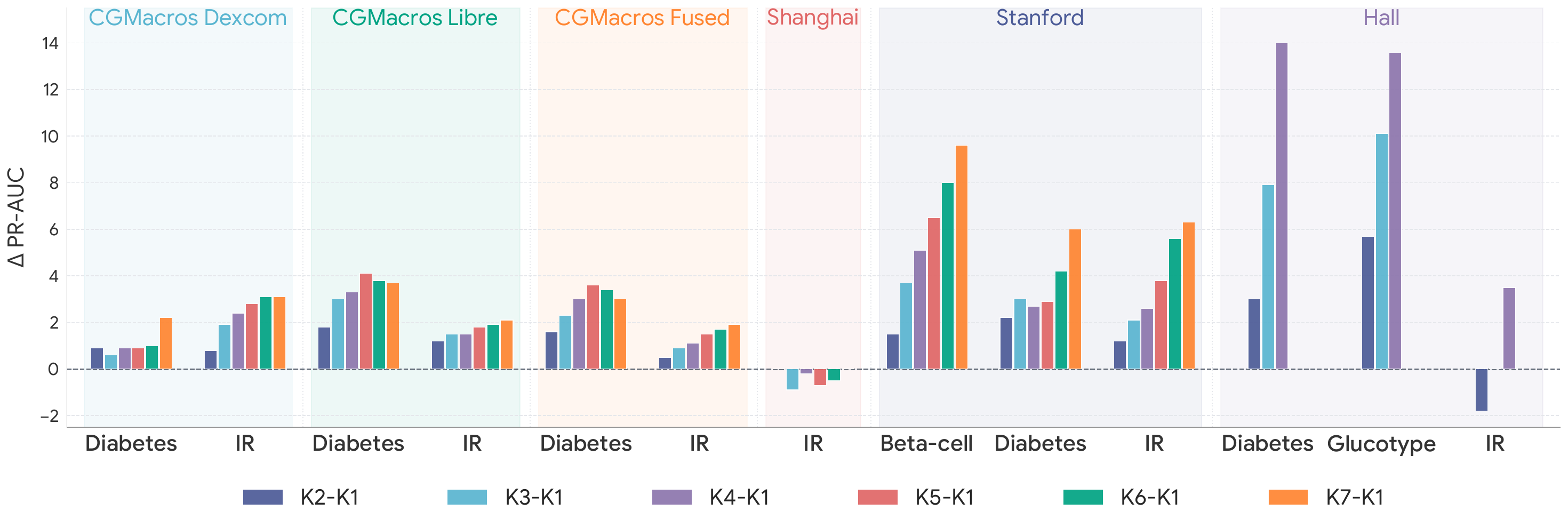}
    \caption{
    \textbf{Effect of multiday CGM observation.}
    Paired PR-AUC change from the one-day subject representation to $K$-day representations, 
    $\Delta_K=\mathrm{PR\text{-}AUC}(K)-\mathrm{PR\text{-}AUC}(1)$. Positive values indicate improvement over $K=1$. \hall uses $K\leq4$; other cohorts use $K\leq7$.
    }
    \label{fig:multiday_observation}
\end{figure}

\textbf{Multiday Representation Observation.}
We evaluate whether longer CGM observation improves subject-level prediction using frozen \name representations. For each subject, we select one fixed eligible $K_{\max}$-day CGM anchor episode and enumerate adjacent $K$-day subwindows within the same anchor for $K=1,\ldots,K_{\max}$. Each day is independently encoded by the frozen \name encoder, and embeddings within each subwindow are mean-pooled into one subject-level representation. For each $K$ and start position, every subject contributes one representation and each test subject receives one prediction, so subwindows from the same subject are not treated as independent test samples. Metrics are computed by start position and averaged within each repeated evaluation. We train linear probes with 10 iterations of 5-fold subject-level cross-validation and report the paired PR-AUC change.

\begin{wrapfigure}{r}{0.48\textwidth}
\centering
\vspace{-1em}
\includegraphics[width=0.48\textwidth]{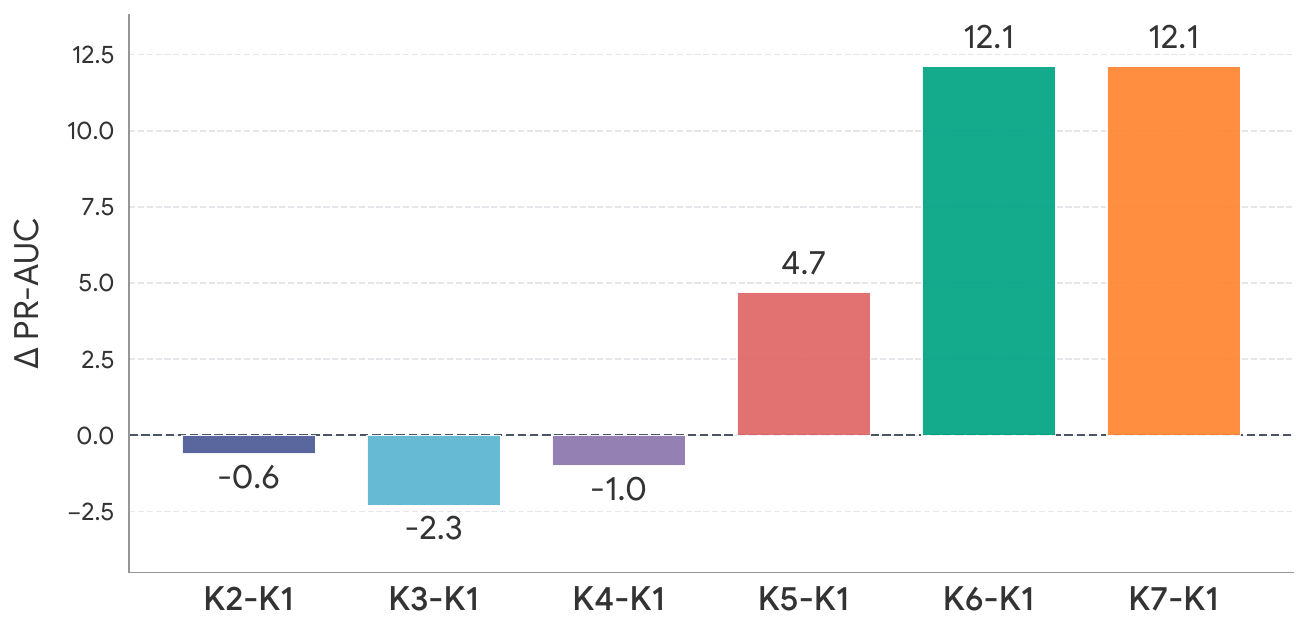} 
\caption{7-day \shanghai IR under concat(mean, max) aggregation.}
\label{fig:sh_multiday}
\vspace{-0.5em}
\end{wrapfigure}
Figure~\ref{fig:multiday_observation} shows that longer CGM observation often improves subject-level prediction. The gains are strongest on \stanford and \hall, suggesting that multiple daily CGM profiles provide a more stable estimate of subject-level metabolic phenotypes. \cgmacros further provides a paired-sensor setting, where Dexcom and Libre are recorded from the same subjects over the same period. We evaluate each sensor separately and include a matched fused setting, where same-subject same-day Dexcom and Libre embeddings are averaged before multiday aggregation. The mostly positive gains across these settings suggest that frozen \name embeddings capture subject-level signals that are reproducible across sensor views. \shanghai IR is an exception under simple mean pooling, but concat(mean, max) pooling (Figure~\ref{fig:sh_multiday}) yields clear gains at longer horizons, especially at $K=6$ and $7$. Overall, these results show that \name can support multiday subject-level prediction, while the best aggregation strategy may depend on the cohort, sensor type, and clinical label.

\begin{figure*}[!t]
\captionsetup[subfigure]{justification=centering,singlelinecheck=true}
    \centering
    \begin{subfigure}[t]{0.19\textwidth}
        \centering
        \includegraphics[width=\linewidth]{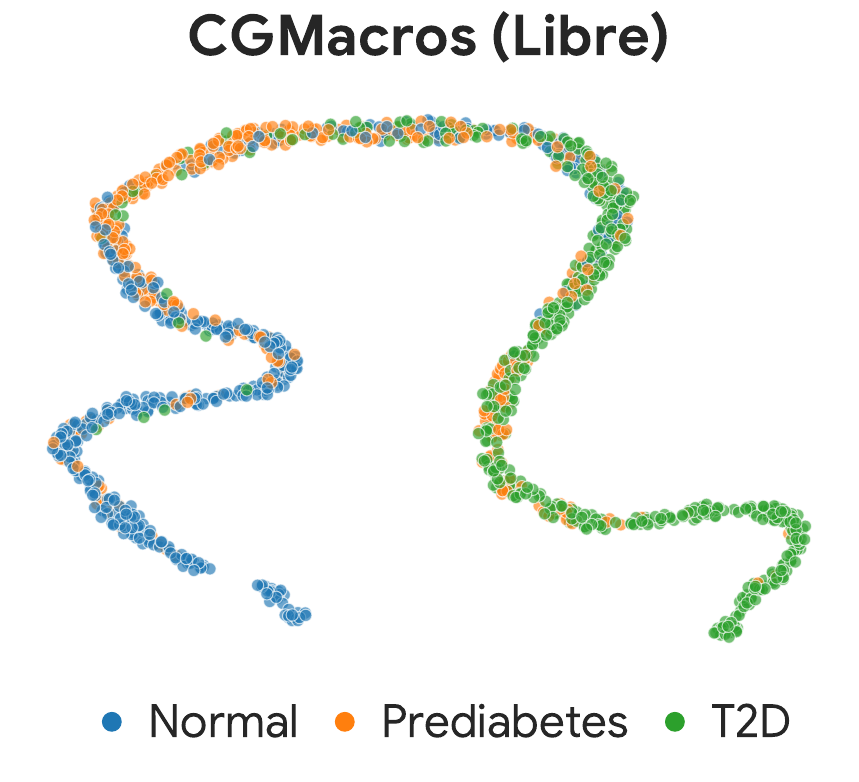}
        \caption{Diabetes}
        \label{fig:umap_diabetes}
    \end{subfigure}
    \hfill
    \begin{subfigure}[t]{0.19\textwidth}
        \centering
        \includegraphics[width=\linewidth]{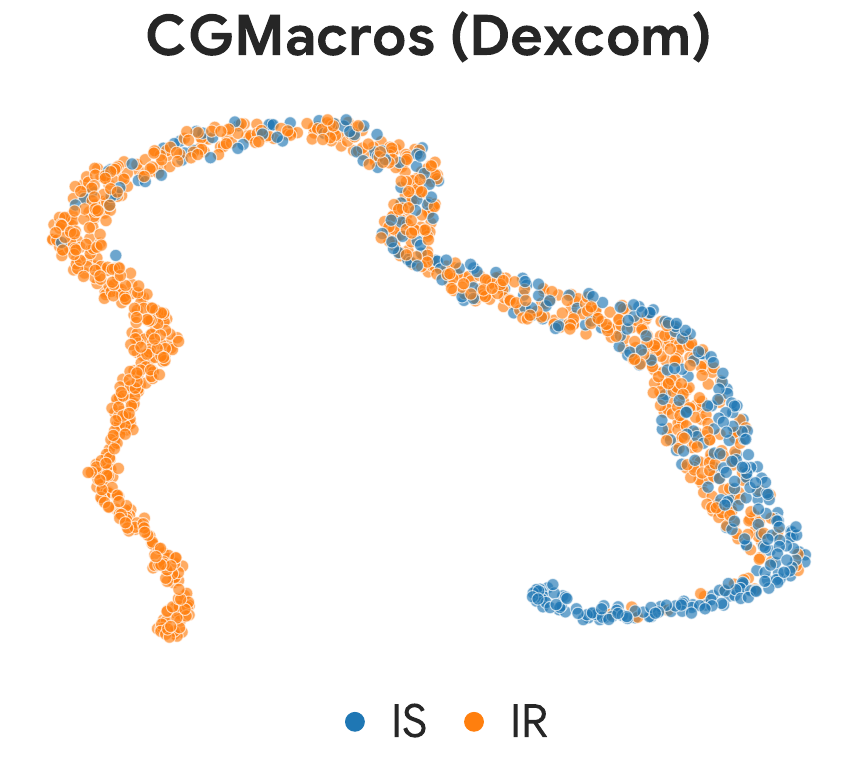}
        \caption{IR}
        \label{fig:umap_ir}
    \end{subfigure}
    \hfill
    \begin{subfigure}[t]{0.19\textwidth}
        \centering
        \includegraphics[width=\linewidth]{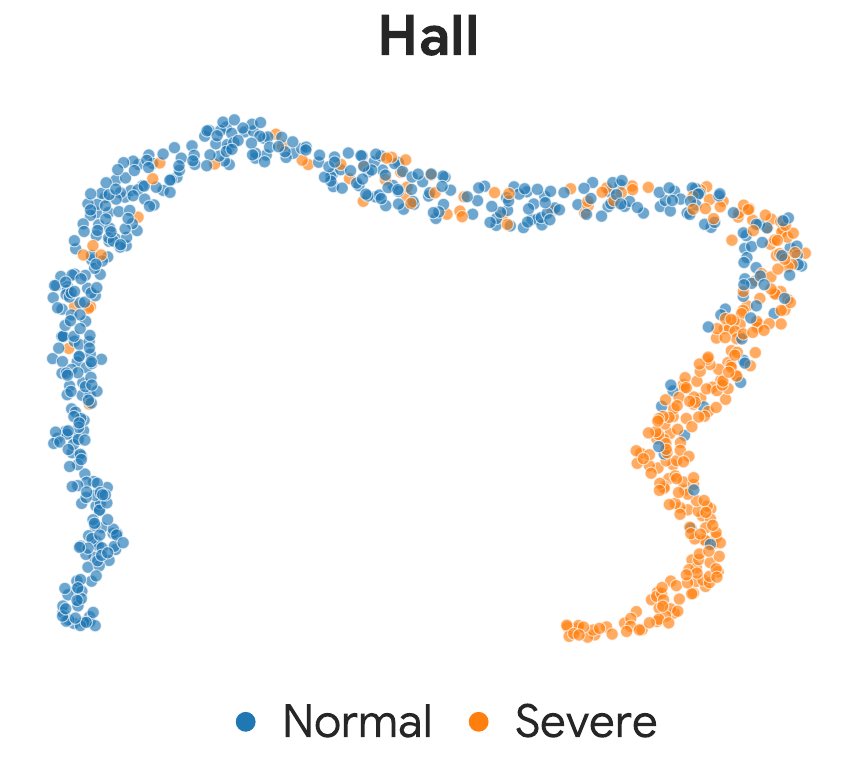}
        \caption{Glucotype}
        \label{fig:umap_glucotype}
    \end{subfigure}
    \hfill
    \begin{subfigure}[t]{0.19\textwidth}
        \centering
        \includegraphics[width=\linewidth]{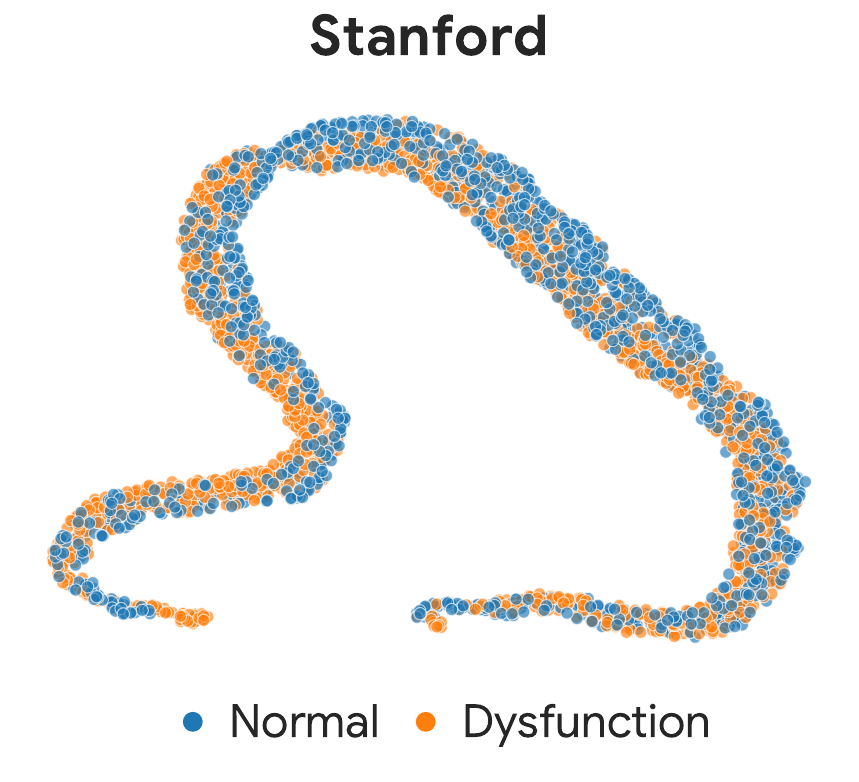}
        \caption{$\beta$-cell}
        \label{fig:umap_betacell_unsup}
    \end{subfigure}
    \hfill
    \begin{subfigure}[t]{0.19\textwidth}
        \centering
        \includegraphics[width=\linewidth]{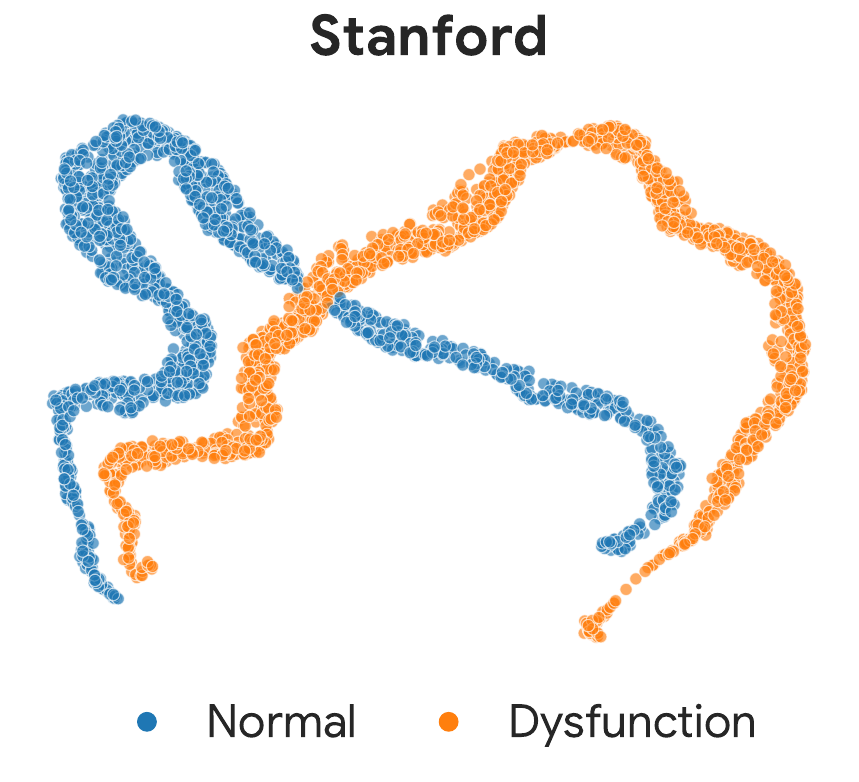}
        \caption{$\beta$-cell (sup.)}
        \label{fig:umap_betacell_sup}
    \end{subfigure}
    \caption{UMAP visualizations of frozen \name embeddings. Panels (a)--(d) show unsupervised UMAP projections; panel (e) shows a supervised UMAP for $\beta$-cell dysfunction.}
    \label{fig:umap_all_tasks}
\end{figure*}

\textbf{Frozen Representation Visualization.}
We use UMAP~\cite{mcinnes2020umapuniformmanifoldapproximation} to qualitatively inspect task-related structure in frozen \name embeddings before downstream adaptation. As shown in Figure~\ref{fig:umap_all_tasks}(a--d), embeddings form continuous manifolds rather than clearly separated clusters, consistent with the gradual nature of glucose regulation and metabolic dysfunction~\cite{becht2019dimensionality,wolf2019paga}. Even without label supervision, several labels are enriched in different regions, especially for diabetes risk, insulin resistance, and glucotype patterns, suggesting that \name captures physiologically relevant variation. More subtle outcomes such as $\beta$-cell dysfunction remain more entangled in the unsupervised view, while the supervised UMAP in Figure~\ref{fig:umap_all_tasks}(e) shows that label-guided projections can reveal more separable structure from the same frozen embeddings. These visualizations provide qualitative support for task-related organization in \name representations.

\textbf{Same-window GMI Comparison.}
We compare \name with a same-window 7-day threshold baseline based on a glucose management indicator (GMI)-equivalent value~\cite{bergenstal2018glucose} for diabetes classification. The GMI-equivalent value is computed from the mean glucose over the same selected seven CGM days used by \name:
$\mathrm{GMI}=3.31+0.02392\times\mathrm{mean\ glucose}_{\mathrm{mg/dL}}$.
For \stanford, diabetes is treated as a binary task, where subjects with $\mathrm{GMI}\geq5.7$ are assigned to the risk-positive class and those with $\mathrm{GMI}<5.7$ to the negative class.
For \cgmacros, diabetes is treated as a three-class task using thresholds at 5.7 and 6.4:
$\mathrm{GMI}<5.7$, $5.7\leq\mathrm{GMI}<6.4$, and $\mathrm{GMI}\geq6.4$.
In the fused \cgmacros setting, GMI is computed from combined Dexcom and Libre readings within the matched seven-day window. Since this baseline is deterministic, we use macro-F1 as the primary metric.

\input{tables/wraptable_GMI}
As shown in Table~\ref{tab:gmi_glucofm_f1}, \name consistently outperforms the same-window 7-day GMI-equivalent threshold baseline across \stanford and \cgmacros sensor settings. The gains are largest on \cgmacros-Dexcom and the fused setting, indicating stronger diabetes category classification than a rule based only on seven-day mean glucose. Since the GMI baseline reduces each seven-day window to a single average-glucose-derived value, these results show that \name retains predictive CGM information beyond seven-day mean glucose. The smaller but positive gain on \cgmacros-Libre further shows that the benefit of representation learning depends on the sensor setting and baseline strength.

\textbf{Pretraining Scalability Analysis.}
We evaluate how \name scales with unlabeled data by sampling 20--100\% of subjects from each pretraining cohort using five random seeds, while keeping the model 
\begin{wrapfigure}{r}{0.48\textwidth}
\centering
\includegraphics[width=0.48\textwidth]{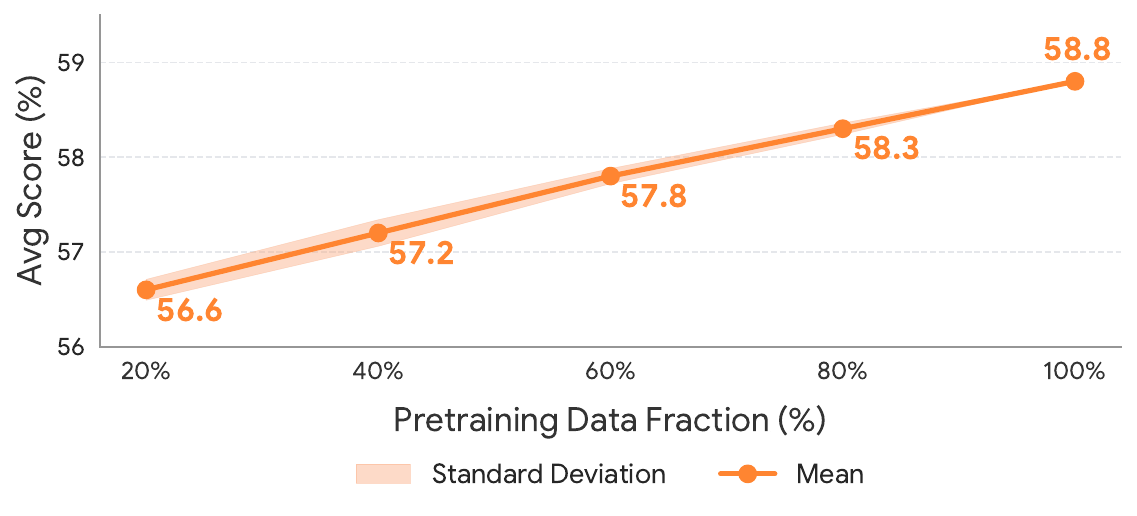} 
\caption{Pretraining data scaling analysis.}
\label{fig:scale}
\vspace{-0.5em}
\end{wrapfigure}
architecture and downstream protocol fixed. As shown in Figure~\ref{fig:scale}, average PR-AUC increases steadily as more pretraining data are used. 
Notably, \name trained with only 20\% of the corpus already matches CGM-specific foundation model baselines trained on the full corpus, suggesting strong data efficiency, consistent with the inductive biases introduced by state--event modeling and JEPA-style pretraining. Performance continues to improve with additional unlabeled data, indicating that \name is not saturated at the current scale. The narrow shaded region shows that the trend is stable across subject subsampling seeds, suggesting that the improvements are not driven by a particular subset of pretraining subjects.

%% file: tables/wraptable_dataset.tex
\begin{wraptable}{r}{0.5\textwidth} 
\vspace{-1em}
  \caption{Composition of \name datasets.}
  \label{tab:pretrain_data}
  \centering
  \resizebox{\linewidth}{!}{ 
  \begin{tabular}{rccc}
    \toprule
    \textbf{Dataset}  & \textbf{Sample Rate} & \textbf{\# Subjects} & \textbf{Duration (h)}  \\
    \midrule
    
    \rowcolor{gray!10}
    \multicolumn{4}{l}{\textit{Pre-train Datasets}} \\
    \wearcgm   & 5 & 192 & 75,330  \\
    \shanghai~\cite{Zhao2023}  & 15 &44 &12,414 \\
    \stanford~\cite{metwally2025prediction}  & 5 &19 &8,761 \\
    \bigideas~\cite{PhysioNet-big-ideas-glycemic-wearable-1.1.2}  & 5 &16 &3,017 \\
    \colas~\cite{Colas2019} & 5 & 206 &9,544 \\
    \midrule
    
    \rowcolor{gray!10}
    \multicolumn{4}{l}{\textit{Downstream Datasets}} \\
    
    \cgmacros~\cite{Das2025}  & 5 | 15  & 45  & 10,376 | 10,998 \\
    \shanghai~\cite{Zhao2023}  & 15 &65 &15,634 \\
    \stanford~\cite{metwally2025prediction} & 5 &37 &27,571 \\
    \hall~\cite{hall2018glucotypes} & 5 &56 & 7,090 \\
    \bottomrule
  \end{tabular}
  }
\end{wraptable}

%% file: tables/table_linear_probe.tex
\begin{table}[!t]
\centering
\caption{Linear-probe performance. The best result is shown in \textbf{bold}, and the second-best result is \underline{underlined}. All values report mean performance over 10 iterations of 5-fold subject-grouped cross-validation. \hlfirst{Core} denotes key CGM-based metabolic phenotyping tasks.}
\label{tab:downstream_results}
\resizebox{\textwidth}{!}{
\begin{tabular}{l|c|l | >{\columncolor{gray!10}}c>{\columncolor{gray!10}}c>{\columncolor{gray!10}}c>{\columncolor{gray!10}}c ccc >{\columncolor{gray!10}}c>{\columncolor{gray!10}}c>{\columncolor{gray!10}}c cccc | >{\columncolor{gray!10}} c}
\toprule

\multirow{2}{*}{\raisebox{-3em}{\textbf{Method}}} & \multirow{2}{*}{\raisebox{-4.5em}{\rotatebox{75}{\textbf{Params}}}} & \multirow{2}{*}{\raisebox{-4.5em}{\rotatebox{75}{\textbf{Metrics}}}} &
\multicolumn{4}{c}{\cellcolor{gray!10}\textbf{\cgmacros}} & 
\multicolumn{3}{c}{\textbf{\shanghai}} & 
\multicolumn{3}{c}{\cellcolor{gray!10}\textbf{\stanford}} & 
\multicolumn{4}{c|}{\textbf{\hall}} & \\

\cmidrule(lr){4-7} \cmidrule(lr){8-10} \cmidrule(lr){11-13} \cmidrule(lr){14-17}

& &   & 
\rotatebox{75}{\hlfirst{Diabetes}} & \rotatebox{75}{\hlfirst{IR}} & \rotatebox{75}{Hyperlip.} & \rotatebox{75}{Obesity} & 
\rotatebox{75}{\hlfirst{IR}} & \rotatebox{75}{Hyperlip.} & \rotatebox{75}{Hypogly.} & 
\rotatebox{75}{\hlfirst{Diabetes}} & \rotatebox{75}{\hlfirst{$\beta$-cell Dys.}} & \rotatebox{75}{\hlfirst{IR}} & 
\rotatebox{75}{\hlfirst{Diabetes}} & \rotatebox{75}{\hlfirst{IR}} & \rotatebox{75}{Hyperlip.} & \rotatebox{75}{Glucotype} & \multirow{-5}{*}{\textbf{Avg}} \\

\midrule
\midrule

\rowcolor{gray!10}
\multicolumn{18}{l}{\textit{General Time-series Foundation Models with Large-scale Pretraining}} \\

\multirow{3}{*}{\begin{tabular}{@{}l@{}}\chronos \\ (small)\end{tabular}} & \multirow{3}{*}{28M} 
& PR   & 51.6 & 82.7 & 30.5 & 56.0 & 65.0 & 34.8 & 16.0 & 69.2 & 58.6 & 62.5 & 51.7 & 44.9 & 24.2 & 54.2 & 50.1 \\
& & AUC & 68.0 & 67.5 & 52.1 & 53.7 & 55.0 & 50.6 & 49.1 & 65.4 & 57.3 & 64.2 & 64.9 & 57.7 & \textbf{61.3} & 62.6 & 59.2 \\
& & F1  & 49.4 & 61.7 & 50.1 & 52.7 & 52.6 & 51.5 & 50.5 & 61.3 & 54.8 & 59.6 & 60.9 & 55.0 & 52.5 & 58.6 & 55.1 \\

\midrule

\multirow{3}{*}{\chronos} & \multirow{3}{*}{120M} 
& PR   & 54.3 & 84.1 & \underline{34.0} & 58.1 & 59.8 & 34.7 & 18.9 & 68.5 & 58.5 & 59.2 & 48.6 & 51.3 & 20.6 & 54.4 & 50.4 \\
& & AUC & 69.7 & 69.9 & \textbf{55.8} & 56.6 & 49.5 & 49.7 & 53.0 & 65.6 & 57.8 & 60.2 & 60.4 & 62.7 & 50.3 & 63.0 & 58.9 \\
& & F1  & 50.2 & 63.3 & \textbf{53.7} & 54.2 & 49.0 & 50.4 & 52.8 & 61.9 & 55.5 & 56.6 & 58.4 & 58.2 & 50.9 & 56.2 & 55.1 \\

\midrule

\multirow{3}{*}{\begin{tabular}{@{}l@{}}\moment \\ (small)\end{tabular}} & \multirow{3}{*}{40M}  
& PR   & 54.0 & 83.8 & 30.3 & 60.9 & 58.3 & 38.3 & 16.3 & 70.4 & 57.2 & 63.0 & 59.2 & 53.1 & 19.4 & 55.6 & 51.4 \\
& & AUC & 69.1 & 70.6 & 51.5 & 58.7 & 49.0 & 55.9 & 52.9 & 67.6 & 55.5 & 65.1 & 69.7 & 62.1 & 48.7 & 61.7 & 59.9 \\
& & F1  & 49.8 & 63.6 & 50.9 & 55.7 & 49.3 & 52.9 & 51.4 & 62.5 & 53.8 & 60.1 & 64.5 & 59.0 & 50.4 & 59.3 & 55.9 \\

\midrule

\multirow{3}{*}{\begin{tabular}{@{}l@{}}\moment \\ (large)\end{tabular}} & \multirow{3}{*}{385M} 
& PR   & 50.3 & 87.4 & 32.0 & 61.8 & 59.6 & \textbf{42.4} & 14.3 & 71.9 & 61.8 & 58.1 & 52.1 & 44.9 & 18.4 & 55.8 & 50.8 \\
& & AUC & 66.1 & 75.2 & 53.2 & 59.9 & 47.8 & \textbf{60.8} & 49.6 & 69.1 & 58.9 & 59.4 & 65.0 & 56.6 & 51.4 & 63.2 & 59.7 \\
& & F1  & 46.8 & 67.2 & \underline{52.5} & 56.3 & 48.6 & \textbf{57.2} & 49.7 & 63.7 & 55.7 & 56.1 & 60.5 & 53.3 & 47.7 & 57.8 & 55.2 \\

\midrule

\multirow{3}{*}{\mantis} & \multirow{3}{*}{8M} 
& PR   & 61.8 & 90.9 & 30.1 & 63.4 & 59.2 & 30.6 & \textbf{22.9} & 75.1 & 63.5 & \underline{67.1} & 54.2 & 55.7 & \underline{24.8} & 77.5 & 55.5 \\
& & AUC & 75.5 & \underline{80.3} & 49.8 & 60.8 & 48.2 & 46.0 & 58.6 & \underline{71.4} & 61.5 & \underline{68.2} & 67.1 & 66.2 & 57.9 & 82.6 & 63.9 \\
& & F1  & 56.0 & \textbf{71.6} & 49.8 & 57.2 & 48.4 & 47.5 & \textbf{53.8} & \underline{65.9} & 58.5 & \underline{62.3} & 59.6 & 59.8 & \textbf{53.7} & 73.3 & 58.4 \\

\midrule

\multirow{3}{*}{\mantisv} & \multirow{3}{*}{4.2M} 
& PR   & \underline{63.6} & \underline{91.2} & 30.3 & 62.8 & 64.1 & 33.3 & \underline{22.0} & 75.2 & \underline{65.7} & 64.6 & 57.7 & \underline{59.3} & \textbf{25.0} & 81.5 & \underline{56.9} \\
& & AUC & 76.7 & \underline{80.3} & 49.1 & 60.3 & 52.7 & 48.5 & \textbf{59.6} & 71.2 & \underline{64.4} & 65.5 & 67.5 & 66.7 & \underline{58.8} & 84.7 & \underline{64.7} \\
& & F1  & 56.2 & 68.9 & 49.2 & 55.9 & 51.2 & 48.1 & \underline{53.3} & 65.7 & \underline{58.6} & 60.4 & 61.1 & \underline{60.8} & 52.6 & 76.7 & \underline{58.5} \\

\midrule
\midrule

\rowcolor{gray!10}
\multicolumn{18}{l}{\textit{Externally Pretrained CGM Foundation Model}} \\

\multirow{3}{*}{\cgmformer} & \multirow{3}{*}{0.85M} 
& PR   & 63.3 & 89.9 & 31.8 & \textbf{68.1} & 54.8 & \underline{40.6} & 11.9 & \underline{75.9} & 62.2 & 61.8 & 51.1 & 48.0 & 18.0 & 79.6 & 54.1 \\
& & AUC & \underline{77.1} & 78.0 & 54.2 & \textbf{66.0} & 43.6 & \underline{58.4} & 42.1 & \underline{71.4} & 59.2 & 63.2 & 66.6 & 58.2 & 46.5 & 84.3 & 62.1 \\
& & F1  & \underline{57.2} & 67.6 & 52.1 & \textbf{60.5} & 45.5 & \underline{55.1} & 45.3 & \textbf{66.2} & 55.7 & 59.0 & 59.3 & 56.3 & 48.4 & 77.1 & 57.5 \\

\midrule
\midrule

\rowcolor{gray!10}
\multicolumn{18}{l}{\textit{CGM Foundation Models Retrained on Our Pretraining Corpus}} \\

\multirow{3}{*}{\cgmjepa} & \multirow{3}{*}{0.52M} 
& PR   & 63.0 & 86.2 & 28.7 & 55.5 & \textbf{69.1} & 37.4 & 17.8 & 66.4 & 58.7 & 61.5 & \underline{59.9} & 56.8 & 17.3 & 87.6 & 54.7 \\
& & AUC & 75.9 & 73.8 & 47.6 & 53.8 & \textbf{60.8} & 53.6 & 56.8 & 61.2 & 55.4 & 61.6 & \underline{73.5} & \underline{68.1} & 40.5 & \textbf{90.7} & 62.4 \\
& & F1  & 55.4 & 66.3 & 48.2 & 53.2 & \textbf{57.2} & 51.3 & 50.4 & 58.7 & 53.9 & 58.1 & \textbf{65.0} & 59.4 & 42.8 & \underline{82.2} & 57.3 \\

\midrule

\multirow{3}{*}{\xcgm} & \multirow{3}{*}{0.52M} 
& PR   & \underline{63.6} & 86.6 & 29.8 & 55.3 & 66.9 & 35.5 & 16.9 & 67.4 & 60.0 & 61.9 & 59.3 & 56.2 & 17.5 & \underline{87.7} & 54.6 \\
& & AUC & 76.6 & 73.6 & 48.0 & 53.2 & \underline{58.4} & 52.1 & 55.7 & 61.8 & 56.4 & 62.0 & 73.0 & 67.7 & 40.0 & \underline{90.6} & 62.1 \\
& & F1  & 56.5 & 65.1 & 48.2 & 52.5 & \underline{55.4} & 49.9 & 49.0 & 59.1 & 54.4 & 58.1 & \underline{64.8} & 59.2 & 42.8 & 82.1 & 56.9 \\

\midrule

\multirow{3}{*}{\begin{tabular}{@{}l@{}}\gluformer \\ (tiny) \end{tabular}} & \multirow{3}{*}{0.65M} 
& PR   & 59.4 & 86.1 & 28.2 & 60.2 & 58.1 & 33.9 & 17.9 & 74.3 & 63.3 & 64.5 & 48.9 & 50.9 & 21.6 & 75.4 & 53.0 \\
& & AUC & 73.9 & 72.5 & 47.9 & 57.9 & 47.6 & 51.1 & 53.3 & 68.9 & 61.9 & 65.4 & 63.3 & 63.6 & 58.0 & 81.7 & 61.9 \\
& & F1  & 51.7 & 65.4 & 48.0 & 55.4 & 47.8 & 50.3 & 49.1 & 63.0 & 58.5 & 61.7 & 57.9 & 59.4 & \underline{53.1} & 72.7 & 56.7 \\

\midrule

\multirow{3}{*}{\begin{tabular}{@{}l@{}}\gluformer \\ (base) \end{tabular}} & \multirow{3}{*}{135M} 
& PR   & 62.9 & 87.2 & 31.4 & 61.4 & 56.5 & 31.5 & 11.5 & 69.8 & 61.1 & 56.1 & 51.9 & 51.6 & 16.4 & 82.7 & 52.3 \\
& & AUC & 76.7 & 74.8 & 53.1 & 60.5 & 44.9 & 46.7 & 43.9 & 63.1 & 58.0 & 56.3 & 66.0 & 61.6 & 46.9 & 87.9 & 60.0 \\
& & F1  & 55.7 & 67.8 & \underline{52.5} & 57.2 & 46.3 & 48.1 & 44.8 & 59.2 & 55.4 & 54.8 & 58.7 & 58.3 & 47.7 & 80.7 & 56.2 \\

\midrule
\midrule

\multirow{3}{*}{\textbf{\name}} & \multirow{3}{*}{0.72M} 
& PR   & \textbf{65.9} & \textbf{91.9} & \textbf{36.1} & \underline{64.9} & \underline{67.0} & 33.5 & 21.1 & \textbf{77.3} & \textbf{69.0} & \textbf{67.6} & \textbf{66.2} & \textbf{60.2} & 14.4 & \textbf{88.3} & \textbf{58.8} \\
& & AUC & \textbf{78.7} & \textbf{81.2} & \underline{54.7} & \underline{62.6} & 57.8 & 50.5 & \underline{59.2} & \textbf{72.8} & \textbf{68.7} & \textbf{69.1} & \textbf{75.9} & \textbf{70.7} & 41.6 & \textbf{90.7} & \textbf{66.7} \\
& & F1  & \textbf{58.3} & \underline{69.6} & 50.2 & \underline{59.4} & \underline{55.4} & 49.1 & 50.7 & \textbf{66.2} & \textbf{63.3} & \textbf{64.0} & 64.5 & \textbf{62.0} & 43.1 & \textbf{82.4} & \textbf{59.9} \\

\bottomrule
\end{tabular}
}
\end{table}

%% file: tables/table_ppgr.tex
\begin{table*}[!t]
    \centering
    \caption{PPGR across cumulative context, averaged over Dexcom and Libre evaluations (MAE$\downarrow$).}
    \label{tab:ppgr_multimetric_progression}
    \tablestyle{2.2pt}{0.92}\scriptsize
    \begin{tabular*}{\textwidth}{@{\extracolsep{\fill}}llccccc>{\columncolor{blue!8}}c@{}}
        \toprule
        Context & Endpoint
        & \shortstack{\gluformer\\[-1pt](tiny)}
        & \shortstack{\gluformer\\[-1pt](base)}
        & \cgmformer
        & \cgmjepa
        & \shortstack{\texttt{X-CGM-}\\[-1pt]\texttt{JEPA}}
        & \textbf{\name} \\
        \midrule

        \multirow{4}{*}{Frozen representation}
        & Trajectory (mg/dL)
        & 26.88 & 27.28 & 27.13 & \underline{26.50}
        & \textbf{26.19} & 26.58 \\

        & Positive iAUC (mg/dL$\cdot$h)
        & 37.20 & 37.82 & 37.81 & \underline{36.89}
        & \textbf{35.78} & 37.08 \\

        & Peak rise (mg/dL)
        & 32.12 & 31.97 & 32.51 & \underline{31.36}
        & \textbf{31.24} & 32.62 \\

        & Peak time (min)
        & \underline{32.13} & 33.70 & 32.28 & 32.27
        & 32.19 & \textbf{31.27} \\

        \addlinespace[2pt]
        \midrule

        \multirow{4}{*}{\shortstack[l]{+ 1 h pre-meal\\CGM}}
        & Trajectory (mg/dL)
        & 25.61 & 26.01 & 26.17 & \underline{24.82}
        & 25.14 & \textbf{24.52} \\

        & Positive iAUC (mg/dL$\cdot$h)
        & 35.34 & 36.10 & 36.30 & \textbf{33.73}
        & 34.09 & \underline{34.01} \\

        & Peak rise (mg/dL)
        & 30.75 & 30.79 & 31.20 & \textbf{29.15}
        & \underline{29.96} & 30.28 \\

        & Peak time (min)
        & \underline{31.06} & 32.96 & 32.12 & 32.66
        & 32.85 & \textbf{30.14} \\

        \addlinespace[2pt]
        \midrule

        \multirow{4}{*}{+ Meal nutrition}
        & Trajectory (mg/dL)
        & 24.45 & 25.33 & 24.75 & 23.42
        & \underline{23.00} & \textbf{22.87} \\

        & Positive iAUC (mg/dL$\cdot$h)
        & 33.13 & 34.83 & 33.54 & 31.11
        & \textbf{30.22} & \underline{30.76} \\

        & Peak rise (mg/dL)
        & 28.66 & 29.80 & 28.81 & \underline{27.04}
        & \textbf{26.64} & 27.14 \\

        & Peak time (min)
        & \underline{31.86} & 32.88 & 32.95 & 32.01
        & 32.50 & \textbf{30.55} \\

        \addlinespace[2pt]
        \midrule

        \multirow{4}{*}{+ Fasting glucose}
        & Trajectory (mg/dL)
        & 24.14 & 25.27 & 24.27 & \underline{22.80}
        & 22.80 & \textbf{22.35} \\

        & Positive iAUC (mg/dL$\cdot$h)
        & 32.64 & 34.69 & 32.76 & 30.36
        & \underline{30.13} & \textbf{29.87} \\

        & Peak rise (mg/dL)
        & 28.30 & 29.80 & 28.08 & \textbf{26.40}
        & 26.58 & \underline{26.47} \\

        & Peak time (min)
        & 31.86 & 33.00 & 32.46 & \underline{31.07}
        & 31.36 & \textbf{30.12} \\

        \addlinespace[2pt]
        \midrule

        \multirow{4}{*}{\shortstack[l]{+ BMI + diabetes\\status}}
        & Trajectory (mg/dL)
        & 24.23 & 25.32 & 24.15 & \underline{22.90}
        & 23.10 & \textbf{21.88} \\

        & Positive iAUC (mg/dL$\cdot$h)
        & 32.56 & 34.68 & 32.17 & \underline{30.24}
        & 30.56 & \textbf{28.50} \\

        & Peak rise (mg/dL)
        & 28.10 & 29.70 & 27.93 & \underline{25.97}
        & 26.59 & \textbf{25.40} \\

        & Peak time (min)
        & 31.81 & 32.62 & 32.06 & \underline{31.38}
        & 31.47 & \textbf{30.31} \\

        \bottomrule
    \end{tabular*}
\end{table*}

%% file: tables/table_cross_dataset.tex
\begin{table}[!t]
    \centering
    \caption{Cross-dataset transfer results. ``\gup{+}'' denotes \name's gain over the second-best method; ``\gdown{-}'' denotes its gap to the best baseline.}
    \label{tab:cross_dataset}
    \resizebox{\textwidth}{!}{%
    \begin{tabular}{l cccc cccc}
        \toprule
        & \multicolumn{4}{c}{\stanford $\rightarrow$ \hall} & \multicolumn{4}{c}{\hall $\rightarrow$ \stanford} \\
        \cmidrule(lr){2-5} \cmidrule(lr){6-9}
        & \multicolumn{2}{c}{\textit{Diabetes Risk Assessment}} & \multicolumn{2}{c}{\textit{Insulin Resistance}} & \multicolumn{2}{c}{\textit{Diabetes Risk Assessment}} & \multicolumn{2}{c}{\textit{Insulin Resistance}} \\
        \cmidrule(lr){2-3} \cmidrule(lr){4-5} \cmidrule(lr){6-7} \cmidrule(lr){8-9}
        Metrics & PR-AUC$^{\uparrow}$ & ROC-AUC$^{\uparrow}$ & PR-AUC$^{\uparrow}$ & ROC-AUC$^{\uparrow}$ & PR-AUC$^{\uparrow}$ & ROC-AUC$^{\uparrow}$ & PR-AUC$^{\uparrow}$ & ROC-AUC$^{\uparrow}$ \\
        \midrule
        \cgmjepa   & 53.0\pho & 73.4\pho & 55.0\pho & 67.9\pho & 68.1\pho & 64.3\pho & 61.3\pho & 61.8\pho \\
        \xcgm & 55.4\pho & 73.7\pho & 57.8\pho & 68.2\pho & 68.2\pho & 64.7\pho & 61.7\pho & 62.4\pho \\
        \gluformer (tiny)  & 58.6\pho & 70.4\pho & 50.8\pho & 62.2\pho & 67.6\pho & 65.6\pho & 59.7\pho & 58.3\pho \\
        
        \rowcolor{gray!10}
        \textbf{\name} & \textbf{61.6}\gup{(+3.0)} & \textbf{74.7}\gup{(+1.0)} & \textbf{61.6}\gup{(+3.8)} & \textbf{72.1}\gup{(+3.9)} & \textbf{73.4}\gup{(+5.2)} & \textbf{69.7}\gup{(+4.1)} & \textbf{67.7}\gup{(+6.0)} & \textbf{69.2}\gup{(+6.8)} \\
        \midrule\midrule

        & \multicolumn{4}{c}{\cgmacros $\rightarrow$ \hall} & \multicolumn{4}{c}{\hall $\rightarrow$ \cgmacros} \\
        \cmidrule(lr){2-5} \cmidrule(lr){6-9}
        & \multicolumn{2}{c}{\textit{Diabetes Risk Assessment}} & \multicolumn{2}{c}{\textit{Insulin Resistance}} & \multicolumn{2}{c}{\textit{Diabetes Risk Assessment}} & \multicolumn{2}{c}{\textit{Insulin Resistance}} \\
        \cmidrule(lr){2-3} \cmidrule(lr){4-5} \cmidrule(lr){6-7} \cmidrule(lr){8-9}
        Metrics & PR-AUC$^{\uparrow}$ & ROC-AUC$^{\uparrow}$ & PR-AUC$^{\uparrow}$ & ROC-AUC$^{\uparrow}$ & PR-AUC$^{\uparrow}$ & ROC-AUC$^{\uparrow}$ & PR-AUC$^{\uparrow}$ & ROC-AUC$^{\uparrow}$ \\
        \midrule
        \cgmjepa   & 50.6\pho & 69.6\pho & 53.4\pho & 67.1\pho & 87.5\pho & 78.1\pho & 89.1\pho & 77.0\pho \\
        \xcgm & 49.5\pho & 69.1\pho & 53.6\pho & 67.2\pho & 87.3\pho & 77.8\pho & 89.2\pho & 77.1\pho \\
        \gluformer (tiny)   & 55.9\pho & 70.2\pho & 50.0\pho & 65.1\pho & 83.9\pho & 74.1\pho & 84.8\pho & 71.0\pho \\
        
        \rowcolor{gray!10}
        \textbf{\name}  & \textbf{63.3}\gup{(+7.4)} & \textbf{78.2}\gup{(+8.0)} & \textbf{61.7}\gup{(+8.1)} & \textbf{72.5}\gup{(+5.3)} & \textbf{88.8}\gup{(+1.3)} & \textbf{81.3}\gup{(+3.2)} & \textbf{90.0}\gup{(+0.8)} & \textbf{78.8}\gup{(+1.7)} \\
        \midrule\midrule

        & \multicolumn{4}{c}{\cgmacros $\rightarrow$ \stanford} & \multicolumn{4}{c}{\stanford $\rightarrow$ \cgmacros} \\
        \cmidrule(lr){2-5} \cmidrule(lr){6-9}
        & \multicolumn{2}{c}{\textit{Diabetes Risk Assessment}} & \multicolumn{2}{c}{\textit{Insulin Resistance}} & \multicolumn{2}{c}{\textit{Diabetes Risk Assessment}} & \multicolumn{2}{c}{\textit{Insulin Resistance}} \\
        \cmidrule(lr){2-3} \cmidrule(lr){4-5} \cmidrule(lr){6-7} \cmidrule(lr){8-9}
        Metrics & PR-AUC$^{\uparrow}$ & ROC-AUC$^{\uparrow}$ & PR-AUC$^{\uparrow}$ & ROC-AUC$^{\uparrow}$ & PR-AUC$^{\uparrow}$ & ROC-AUC$^{\uparrow}$ & PR-AUC$^{\uparrow}$ & ROC-AUC$^{\uparrow}$ \\
        \midrule
        \cgmjepa   & 64.1\pho & 60.9\pho & 59.8\pho & 60.3\pho & 87.8\pho & 78.4\pho & 85.9\pho & 74.0\pho \\
        \xcgm & 65.4\pho & 62.3\pho & 59.7\pho & 60.2\pho & \textbf{88.3}\pho & 78.5\pho & 83.6\pho & 70.7\pho \\
        \gluformer (tiny)   & 68.5\pho & 64.0\pho & 64.0\pho & \textbf{65.2}\pho & 85.2\pho & 76.2\pho & \textbf{87.9}\pho & \textbf{76.3}\pho \\
        
        \rowcolor{gray!10}
        \textbf{\name} & \textbf{77.1}\gup{(+8.6)} & \textbf{73.6}\gup{(+9.6)} &\textbf{65.4}\gup{(+1.4)} &64.8\gdown{(-0.4)} 
        & \textbf{88.3}\gup{(+0.5)} & \textbf{79.9}\gup{(+1.4)} &87.3\gdown{(-0.6)} & 73.3\gdown{(-3.0)} \\
        \bottomrule
    \end{tabular}%
    }
\end{table}

%% file: tables/wraptable_GMI.tex
\begin{wraptable}{r}{0.5\textwidth} 
\vspace{-1em}
\caption{GMI-equivalent comparison on Macro-F1.}
\label{tab:gmi_glucofm_f1}
  \centering
  \resizebox{\linewidth}{!}{ 
\begin{tabular}{lccc}
\toprule
Dataset & GMI Rule & \name & $\Delta$ \\
\midrule
\stanford & 59.6 & \textbf{67.0} & \ggup{+07.4} \\
\cgmacros-Dexcom & 36.3 & \textbf{53.7} & \ggup{+17.4} \\
\cgmacros-Libre & 63.7 & \textbf{65.6} & \ggup{+01.9}\\
\cgmacros-Fused & 56.9 & \textbf{65.9} & \ggup{+09.0} \\
\bottomrule
\end{tabular}
  }
\end{wraptable}

%% file: sections/5_ablation.tex
\section{Ablation}
\label{sec:ablation}

We ablate key components of \name under identical pretraining and evaluation protocols. The main text reports task-averaged results, with protocols and additional architecture results in Appendix~\ref{apd:ablation}.

\input{tables/table_prefusion_content_probes}

\textbf{Dual Stream vs. Single Stream.}
We compare a raw-input tokenizer, state-only and event-only single-stream encoders, and the proposed dual-stream state--event encoder. For fairness, all variants receive the corresponding patch-level statistics and temporal-difference features; they differ only in how glucose signals and auxiliary features are organized. As shown in Table~\ref{tab:stream_ablation}, the dual-stream encoder achieves the strongest task-averaged performance across all three downstream metrics, supporting the value of organizing slow-varying glucose trends and short-term deviations in separate streams before fusion.

We further probe frozen pre-fusion state and event tokens on 5-min Dexcom recordings from \cgmacros, enabling interpolation-free measurement of short-term change. State tokens retain substantially more information about hourly glucose level, whereas event tokens retain more information about 5-min absolute change. Paired subject-cluster bootstrap intervals exclude zero for both directional gaps. Together, the downstream ablation and content probes show that the two branches preserve distinct temporal emphasis and that their integration improves downstream representation quality.
Protocol details are provided in Appendix~\ref{app:prefusion_content_probes}.

\begin{wrapfigure}{r}{0.48\textwidth}
\vspace{-1em}
\centering
\includegraphics[width=0.48\textwidth]{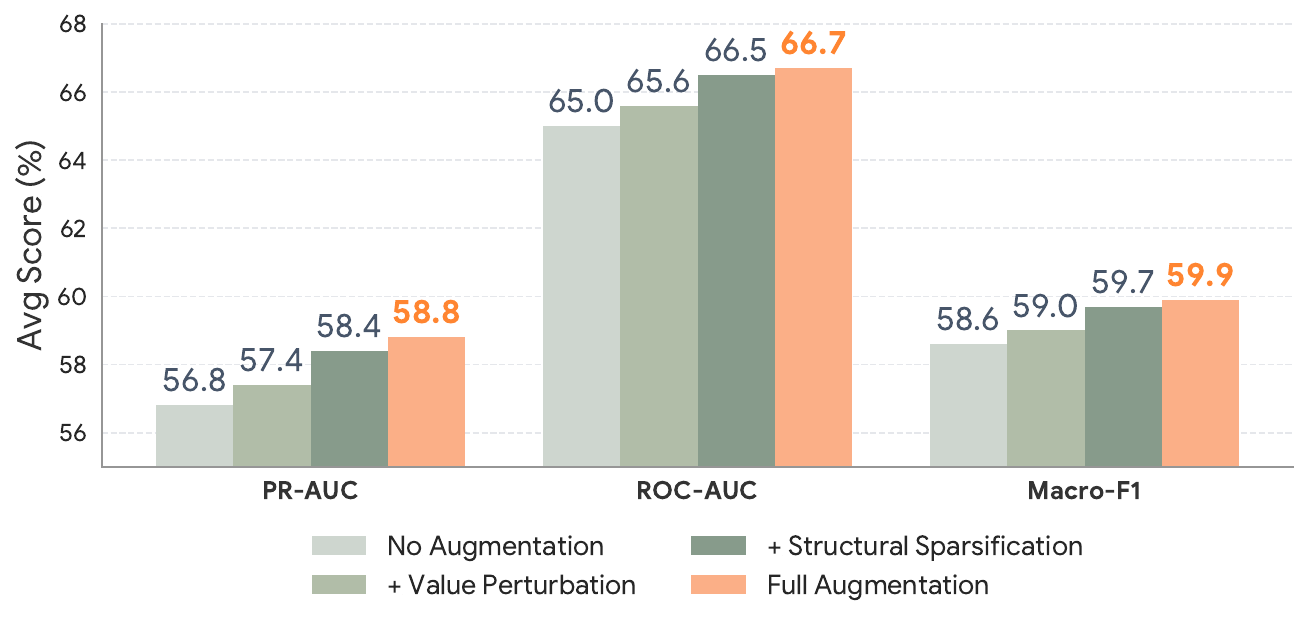} 
\caption{Data augmentation ablation.}
\label{fig:abl_aug}
\vspace{-0.5em}
\end{wrapfigure}
\textbf{Training-time Data Augmentation.}
We evaluate whether \name benefits from the proposed CGM-aware augmentation pipeline. As shown in Figure~\ref{fig:abl_aug}, augmentation improves task-averaged downstream performance across all metrics. Value-level perturbations provide modest gains by exposing the model to amplitude shifts and transient artifacts, while structural sparsification contributes more by simulating realistic missingness and heterogeneous sampling rates. This suggests that robustness to observation-pattern variation is particularly important for CGM representation learning. Overall, augmentation improves downstream representation quality but complements rather than replaces the representation design.

\begin{wrapfigure}{r}{0.48\textwidth}
\vspace{-1em}
\centering
\includegraphics[width=0.48\textwidth]{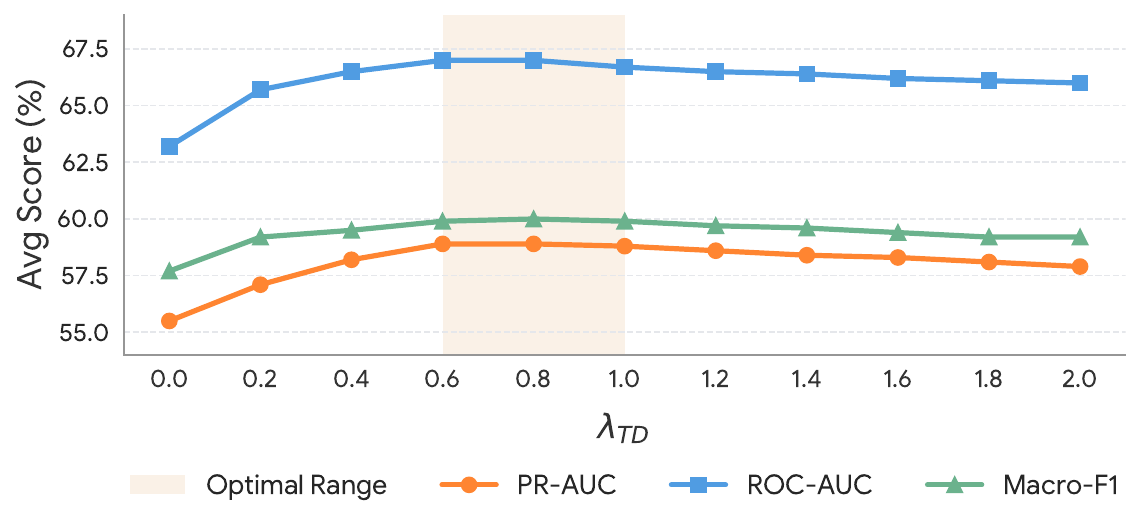} 
\caption{Temporal dynamics weight ablation.}
\label{fig:abl_lambda}
\vspace{-0.5em}
\end{wrapfigure}
\textbf{Temporal Dynamics Weight.}
Figure~\ref{fig:abl_lambda} evaluates the weight of the temporal dynamics objective. Removing this objective ($\lambda_{\mathrm{TD}}=0$) reduces task-averaged performance across all metrics, indicating that masked contextual latent prediction alone does not fully capture clinically useful glucose evolution. Performance improves as $\lambda_{\mathrm{TD}}$ increases and reaches a broad optimum around $0.6$--$1.0$, then gradually declines when the dynamics term becomes too dominant. This supports a balanced pretraining objective: contextual prediction captures global daily structure, while temporal dynamics modeling provides complementary constraints on local state--event transitions.

\begin{wrapfigure}{r}{0.48\textwidth}
\vspace{-1em}
\centering
\includegraphics[width=0.48\textwidth]{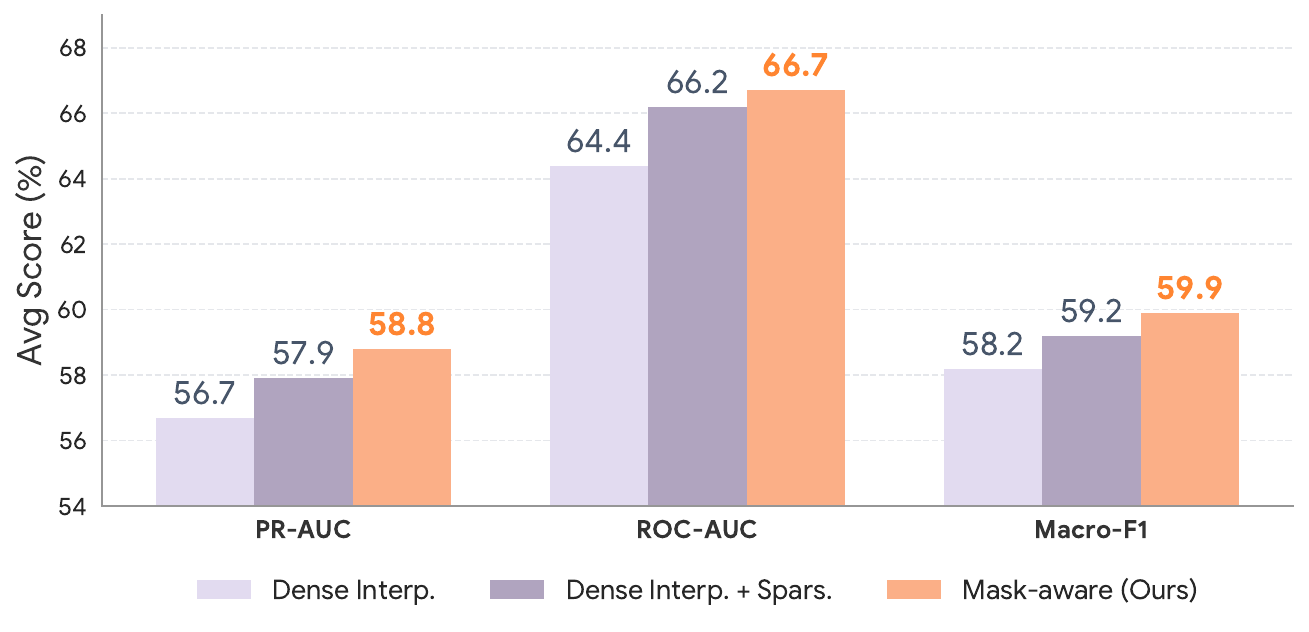} 
\caption{Dense interpolation ablation.}
\label{fig:abl_intp}
\vspace{-0.5em}
\end{wrapfigure}
\textbf{Dense Interpolation.}
We test whether \name requires the dense interpolation preprocessing commonly used by CGM-specific baselines. We compare dense interpolation alone with dense interpolation plus structural sparsification. As shown in Figure~\ref{fig:abl_intp}, both interpolated variants underperform the default mask-aware setting, although they remain competitive with CGM-specific foundation model baselines on average. Adding structural sparsification narrows the gap, suggesting that exposure to structured missingness can partially reduce interpolation-induced shortcuts. Overall, preserving the observation mask is more effective for \name than treating imputed values as fully observed measurements, especially when CGM recordings contain heterogeneous sampling rates and sensor dropouts.

%% file: tables/table_prefusion_content_probes.tex
\begin{wraptable}{r}{0.48\textwidth}
\vspace{-1em}
\centering
\caption{Dual-stream ablation and probes.}
\label{tab:stream_ablation}
\resizebox{\linewidth}{!}{%
\begin{tabular}{@{}lccc@{}}
\toprule
\rowcolor{gray!10}
\multicolumn{4}{@{}l}{\textit{Downstream utility (\%)}} \\
\textbf{Encoder} & \textbf{PR-AUC} & \textbf{ROC-AUC} & \textbf{Macro-F1} \\
\midrule
Raw input & 55.3 & 63.7 & 58.3 \\
State only & 55.6 & 63.7 & 57.9 \\
Event only & 51.8 & 60.1 & 55.6 \\
\textbf{Dual stream} & \textbf{58.8} & \textbf{66.7} & \textbf{59.9} \\
\midrule
\rowcolor{gray!10}
\multicolumn{4}{@{}l}{\textit{Pre-fusion content probes ($R^2$)}} \\
\textbf{Target} & \textbf{State} & \textbf{Event} & \textbf{Gap (95\% CI)} \\
\midrule
Glucose level & \textbf{0.962} & 0.401 & S$-$E: 0.561 [0.518, 0.630] \\
5-min abs. change & 0.832 & \textbf{0.888} & E$-$S: 0.056 [0.040, 0.073] \\
\bottomrule
\end{tabular}
}
\vspace{-0.5em}
\end{wraptable}

%% file: sections/6_discussion.tex
\section{Discussion}
\label{sec:discussion}

\name provides reusable daily CGM representations for both metabolic phenotyping and event-conditioned PPGR. The pre-fusion probes and encoder ablations further connect this downstream utility to complementary slow-trend and short-term-deviation representations.

\textbf{Limitations.}
Despite these findings, \name is limited by the scale and diversity of available CGM data. Large-scale CGM pretraining remains challenging because data collection requires physical sensors, multi-day participant compliance, and often additional laboratory or clinical assessments for downstream labels. Although we use subject-grouped splits, identical downstream protocols, and re-pretrain available CGM-specific baselines on the same corpus when code permits, the pretraining population remains modest and may not fully capture demographic, device, lifestyle, and disease heterogeneity. Our evaluation is restricted to retrospective phenotype classification and meal-aligned PPGR prediction and does not assess longitudinal outcomes, treatment response, or prospective clinical deployment. Moreover, although multiday embedding aggregation improves subject-level prediction, the encoder itself still models 24-hour windows independently, leaving longer-range temporal modeling to simple downstream aggregation. \name is a research prototype, has not been cleared or approved by any regulatory authority, and is not intended to diagnose, treat, cure, or prevent any disease, nor should it be used as a substitute for professional medical advice.

\textbf{Future Work.}
Future work should scale pretraining to larger and more diverse CGM cohorts, including broader device coverage and more real 15-minute recordings. Extending \name from post-hoc multiday aggregation to native multi-day context modeling may better capture longitudinal metabolic state, treatment response, and slower changes over weeks or months. Another important direction is real-time CGM modeling under evolving missingness, device changes, and behavioral context. The PPGR results also motivate free-living prediction under uncertain meal timing and evolving activity, insulin, sleep, and sensor context.

\textbf{Conclusion.}
We introduced \name, a lightweight CGM foundation model that preserves observation patterns while modeling slow-varying trends and short-term deviations. Across multiple cohorts and seven phenotype-classification tasks, \name achieves strong frozen linear probing, low-label adaptation, cross-dataset transfer, and multiday subject-level prediction. The same frozen representation achieves the lowest full-context two-hour PPGR error across both sensors and all four response endpoints. These results suggest that CGM foundation models benefit from preserving sensing structure and explicitly modeling the multiscale nature of glucose dynamics, rather than treating CGM as a generic one-dimensional time series. By learning transferable representations from unlabeled CGM, \name may reduce reliance on large labeled clinical cohorts for metabolic phenotyping; however, it should be validated within the intended population and clinical workflow before real-world use. We will release the code and reproducibility scripts to support transparent evaluation and future CGM foundation-model research.

%% file: sections/7_appendix.tex
\section{Datasets}
\label{apd:dataset}

\subsection{CGM Segmentation and Missingness Handling}
\label{apd:cgm_segmentation}

Before extracting 24-hour windows, we split each subject's raw CGM trace into continuous recording segments based on timestamp gaps. Gaps of at most 1 hour are treated as short interruptions within the same segment. After aligning the trace to the target 5-minute grid, these missing positions are retained as unobserved entries with NaN glucose values and mask value 0, rather than being treated as real measurements. In contrast, gaps longer than 1 hour are treated as segment boundaries, and the trace is split into separate segments.

As a result, each 24-hour training window is extracted from a continuous recording segment and does not contain any single real-data gap longer than 1 hour. This preserves realistic short missingness while avoiding windows that span long non-wear or sensor-disconnection periods. For fair comparison, all applicable baselines use the same pretraining inputs and identical downstream fold splits as \name.

\subsection{Pretraining Datasets}
\label{apd:pretraining_datasets}

To increase the diversity of unlabeled pretraining data, we extract 24-hour windows from each continuous CGM segment using a fixed-seed random overlapping strategy. For each segment, we randomly sample overlapping 24-hour windows with a coverage ratio between 20\% and 80\%, rather than using only fixed non-overlapping windows. This exposes the model to diverse daily start times and temporal contexts while keeping sampling reproducible. Since pretraining is self-supervised and label-free, overlapping windows improve temporal coverage without introducing label leakage; downstream evaluation instead uses non-overlapping windows with subject-grouped splits.

\textbf{\wearcgm.}
The \wearcgm dataset is a non-public multimodal CGM dataset collected as part of Google/Fitbit research studies across two sequential research phases with non-overlapping participants. The protocols for the \wearcgm datasets were approved by Advarra (Institutional Review Board (IRB) no. Pro00059582 and Pro00069880). All participants provided written informed consent prior to data collection, which included permission for the use of de-identified data in secondary research and algorithm development. It contains CGM recordings from healthy, non-diabetic adults in the United States, collected using Dexcom G6 Pro sensors at 5-minute intervals. Phase 1 included 105 participants in an observational protocol lasting approximately 4 weeks, with 2--3 sensors used per participant. Phase 2 included 87 participants over up to 15 days and incorporated standardized meal challenges designed to elicit metabolic responses. Phase 2 also collected clinical measurements, including fasting glucose and insulin, comprehensive metabolic panels, lipid panels, HbA1c, C-reactive protein, and up to two 2-hour oral glucose tolerance tests after overnight fasting. Together, the two phases provide 75,330 hours of CGM recordings. Although the dataset also contains wearable, nutrition, and blood-pressure measurements, only CGM data are used in this work.

\textbf{\bigideas.}
The \bigideas dataset~\cite{PhysioNet-big-ideas-glycemic-wearable-1.1.2} contains CGM recordings from 16 subjects, collected with Dexcom sensors at 5-minute intervals. After preprocessing, it provides 3,017 monitoring hours, corresponding to an average duration of approximately 7.9 days per subject.

\textbf{\colas.}
The \colas dataset~\cite{Colas2019} contains CGM recordings from 206 subjects after preprocessing, collected with iPro sensors at 5-minute intervals. It provides 9,544 monitoring hours, corresponding to an average duration of approximately 1.9 days per subject.

\textbf{\stanford.}
The \stanford dataset~\cite{metwally2025prediction} contains CGM recordings from 56 subjects, collected with Dexcom sensors at 5-minute intervals. Among them, 19 subjects lack the clinical metadata required for downstream label construction and are therefore used only for unlabeled pretraining. This pretraining subset contains 8,761 monitoring hours, corresponding to an average duration of approximately 19.2 days per subject. The remaining 37 subjects have complete clinical profiles and are used for downstream evaluation.

\textbf{\shanghai.}
The \shanghai dataset~\cite{Zhao2023} contains CGM recordings from patients with type 2 diabetes, collected with FreeStyle Libre sensors at 15-minute intervals. Most participants have one recording session, while seven participants provide 2--3 separate sessions. Because clinical measurements may vary across visits separated by long intervals, we follow the baseline protocol and treat each recording visit as a distinct subject-entry. Among these subject-entries, 44 lack sufficient clinical metadata for downstream label construction and are used only for unlabeled pretraining. An identity audit found one biological participant with an earlier unlabeled visit in this pretraining subset and a later labeled visit in the downstream subset. The \shanghai pretraining/downstream separation is therefore defined at the subject-entry (recording-visit) level. This pretraining subset contains 12,414 monitoring hours, corresponding to an average duration of approximately 11.8 days per subject-entry.

\subsection{Downstream Datasets}
\label{apd:downstream_datasets}

The clinical thresholds below are used to define downstream prediction labels consistently across cohorts and are not intended as standalone diagnostic criteria. For all downstream datasets, CGM traces are divided into non-overlapping 1-day segments, which avoids duplicated temporal evidence across samples and reduces leakage or over-counting from highly correlated overlapping windows.

\input{tables/appendix_table_dataset}

\textbf{\cgmacros.}
The \cgmacros dataset~\cite{Das2025} contains multimodal physiological data from 45 subjects, each wearing both Dexcom and FreeStyle Libre sensors for approximately 10 days. The Dexcom and Libre recordings provide 10,376 and 10,998 monitoring hours, respectively. To avoid subject leakage, recordings from both sensor brands for the same subject are always assigned to the same training or testing split. We define four downstream tasks: (1) diabetes risk, with three classes: normoglycemic, prediabetes, and type 2 diabetes; (2) insulin resistance, where a subject is labeled positive if $\mathrm{HOMA\text{-}IR}=(\mathrm{Insulin}_{\mu U/mL}\times\mathrm{Fasting~Glucose}_{mg/dL})/405.0 > 2.9$; (3) obesity, defined as $\mathrm{BMI}=\mathrm{Weight}_{kg}/(\mathrm{Height}_{m})^2 \ge 30$; and (4) hyperlipidemia, where a positive label is assigned if total cholesterol $\ge 240$, LDL $\ge 160$, or triglycerides $\ge 200$ mg/dL.

\textbf{\hall.}
The \hall dataset~\cite{hall2018glucotypes} originally contains 57 subjects. We retain 56 subjects with sufficient clinical information for constructing diabetes risk, insulin resistance, hyperlipidemia, and glucotype labels. The processed CGM recordings are collected with Dexcom sensors at 5-minute intervals and contain 7,090 monitoring hours, with an average duration of approximately 5.3 days per subject. We define four binary downstream tasks: (1) diabetes risk, where prediabetes or diabetes is grouped as abnormal glucose regulation and compared against normoglycemic status; (2) glucotype, distinguishing severe glucose fluctuation from normal profiles, where low and moderate glucotype categories are grouped as non-severe; (3) insulin resistance, determined by steady-state plasma glucose (SSPG), where a subject is labeled positive if $\mathrm{SSPG}>120$, or, when SSPG is unavailable, by $\mathrm{HOMA\text{-}IR}=(\mathrm{Insulin}_{\mu U/mL}\times\mathrm{FBG}_{mg/dL})/405.0 > 2.9$; and (4) hyperlipidemia, where a positive label is assigned if total cholesterol $\ge 240$, LDL $\ge 160$, or triglycerides $\ge 200$ mg/dL.

\textbf{\stanford.}
The downstream \stanford subset~\cite{metwally2025prediction} contains 37 subjects with complete clinical profiles, providing 27,571 monitoring hours and an average duration of approximately 31 days per subject. We define three binary downstream tasks: (1) insulin resistance, mapping insulin-sensitive subjects to 0 and insulin-resistant subjects to 1 based on SSPG-derived classes; (2) $\beta$-cell dysfunction, where subjects are categorized as dysfunction versus normal based on the median disposition index (DI); and (3) diabetes risk, where subjects are labeled as abnormal glucose regulation if $\mathrm{HbA1c} \ge 5.7\%$ and normoglycemic otherwise.

\textbf{\shanghai.}
The downstream \shanghai subset~\cite{Zhao2023} contains 65 labeled sessions from 58 biological participants, providing 15,634 monitoring hours and an average duration of approximately 10 days per session. Each session is evaluated as one subject-entry, and all 24-hour windows from that entry remain in the same cross-validation fold. We define three binary downstream tasks: (1) hypoglycemia, mapped directly from clinical records, where ``yes'' indicates the positive class and ``no'' indicates the negative class; (2) insulin resistance, computed using HOMA-IR after converting fasting insulin from pmol/L to $\mu$U/mL by dividing by 6.945, where a session is labeled positive if $\mathrm{HOMA\text{-}IR}=(\mathrm{Insulin}_{\mu U/mL}\times\mathrm{Glucose}_{mg/dL})/405.0 > 2.9$; and (3) hyperlipidemia, where lipid profiles are converted from mmol/L to mg/dL and a positive label is assigned if total cholesterol $\ge 240$ ($\mathrm{TC}_{mmol/L}\times 38.67$), LDL $\ge 160$ ($\mathrm{LDL}_{mmol/L}\times 38.67$), or triglycerides $\ge 200$ ($\mathrm{TG}_{mmol/L}\times 88.57$).

\section{Baselines}
\label{apd:baseline}

We compare \name with three groups of baselines: general-purpose time-series foundation models, an open-weight CGM-specific foundation model pretrained on an external cohort, and CGM-specific foundation models retrained on our pretraining corpus. We use author-released checkpoints for the general-purpose models and \cgmformer. For \cgmjepa, \xcgm, and \gluformer, official pretrained checkpoints were unavailable at the time of evaluation, so we reproduced the published preprocessing, architecture, and objectives where applicable and retrained the models on our pretraining corpus. \gluformer-base follows the published model scale, whereas \gluformer-tiny is a parameter-matched variant introduced in this study. For all representation-based evaluations, encoders are frozen and the same downstream logistic regression classifier is trained unless otherwise stated.

\subsection{General-purpose Time-series Foundation Models}

\textbf{\chronos.}
\chronos~\cite{ansari2025chronos2} is a general-purpose time-series forecasting foundation model for zero-shot univariate, multivariate, and covariate-informed forecasting. It extends \chronosone with group attention to share information across related series, variables, targets, and covariates. We evaluate both \chronos and \chronos-small as general-purpose foundation model baselines for frozen representation extraction. To obtain sequence-level embeddings, we feed each CGM sequence into the model and extract the output hidden states. We aggregate valid hidden states with mean pooling while excluding the EOS token, which performed best in our preliminary pooling comparison.

\textbf{\moment.}
\moment~\cite{goswami2024moment} is an open time-series foundation model pretrained on the Time Series Pile with masked time-series modeling. Following the official pipeline, we remove null values, align each sequence to 288 points using linear interpolation, and extract frozen representations for downstream evaluation. We include \moment to assess whether large-scale general-purpose time-series pretraining transfers to CGM-based clinical prediction.

\textbf{\mantis and \mantisv.}
\mantis~\cite{feofanov2025mantislightweightcalibratedfoundation} is a lightweight time-series foundation model for classification, using a Vision Transformer-style architecture and contrastive pretraining. \mantisv~\cite{feofanov2026mantisv2closingzeroshotgap} improves \mantis for zero-shot time-series classification through stronger synthetic-data pretraining, architectural refinements, and test-time representation strategies. Following the official preprocessing for both models, we linearly interpolate each input sequence to 512 points and extract frozen encoder representations for downstream evaluation.

\subsection{Externally Pretrained CGM Foundation Model}

\textbf{\cgmformer.}
\cgmformer~\cite{lu2025pretrained} is a CGM-specific Transformer pretrained with masked learning on daily CGM profiles. The released checkpoint is pretrained on an external cohort of 1,917 CGM-days from 964 participants and evaluated in the original work on tasks such as diabetes screening, metabolic subtyping, and dietary recommendation. We use it as an externally pretrained CGM-specific baseline. Following the official preprocessing scheme, each CGM window is represented on a 288-point 5-minute grid; missing bins are mapped to \cgmformer's \texttt{<pad>} token; glucose values are discretized using the released vocabulary; and a \texttt{<cls>} token is prepended to form a 289-token input. We keep the checkpoint frozen and use mean-pooled hidden states as window-level embeddings, following the official embedding extraction strategy; in our implementation, the mean is weighted by the attention mask to exclude \texttt{<pad>} positions.

\subsection{CGM Foundation Models Retrained on Our Pretraining Corpus}

\textbf{\cgmjepa.}
\cgmjepa~\cite{muhammad2026cgmjepalearningconsistentcontinuous} is a CGM-specific predictive self-supervised model that predicts masked latent targets instead of reconstructing raw glucose values. We retrain it on the same unlabeled CGM corpus as \name using the official configuration: inputs are aligned to a 5-minute grid, linearly interpolated to 288-point daily sequences, normalized, and encoded by a 3-layer Transformer with hidden dimension 96 and 6 attention heads. It is trained with batch size 128, learning rate $10^{-4}$, mask ratio 0.25, and 101 epochs. Downstream embeddings are extracted from the frozen encoder before the projection head and mean-pooled over sequence tokens.

\textbf{\xcgm.}
\xcgm~\cite{muhammad2026cgmjepalearningconsistentcontinuous} uses the same encoder, preprocessing, and optimization configuration as \cgmjepa, but adds an auxiliary Glucodensity-based cross-view objective. Following the official setting, Glucodensity views are constructed as KDE-based 2D joint density grids over three variable pairs: glucose--speed, glucose--acceleration, and speed--acceleration, where speed and acceleration denote first- and second-order glucose differences. Each density grid has size $32 \times 32$, is divided into non-overlapping spatial patches, and is partially masked during pretraining. We retrain \xcgm on the same corpus as \name and extract downstream embeddings from the frozen encoder using the same mean-pooling protocol as \cgmjepa.

\textbf{\gluformer.}
\gluformer~\cite{lutsker2026foundation} is a generative CGM foundation model trained with self-supervised autoregressive glucose-token prediction. We retrain two variants on the same pretraining corpus as \name. \gluformer (base) follows the original setting: 5-minute alignment, imputation, clipping to $[40,500]$, discretization into shifted glucose tokens, and a 16-layer Transformer with hidden dimension 1024, 16 attention heads, and feed-forward dimension 2048, trained with Adam at learning rate $5\times10^{-5}$ for 76 epochs. \gluformer (tiny) uses the same preprocessing and objective but a smaller 4-layer Transformer with hidden dimension 128, 4 attention heads, and feed-forward dimension 256, chosen to provide a parameter scale comparable to \name and \cgmjepa. It is trained with Adam at learning rate $10^{-4}$, batch size 128, for 100 epochs. Downstream embeddings are extracted from frozen hidden states over valid non-padding tokens using max pooling by default.

\section{Additional Implementation Details}
\label{apd:implementation}
\subsection{CGM Window Construction and Chronological Grid Alignment}
\label{apd:window_alignment}

All inputs to \name are pre-segmented 24-hour CGM windows. For datasets originally containing multi-day recordings, the segmentation strategy is described in Appendix~\ref{apd:dataset}. Given a 24-hour segment with readings $X=\{x_i\}_{i=1}^{N}$ and timestamps $T=\{t_i\}_{i=1}^{N}$, we align it to a fixed 24-hour grid with 5-minute resolution, yielding $L=288$ grid positions. The first timestamp $t_1$ defines the circadian start index
\begin{equation}
s =
\left\lfloor
\frac{60\cdot \mathrm{hour}(t_1)+\mathrm{minute}(t_1)}
{5}
\right\rfloor ,
\end{equation}
and the absolute time-of-day index of grid position $j$ is $a_j=(s+j)\bmod L$.

Each reading is assigned to a grid index based on its elapsed time from the window start:
\begin{equation}
u_i=\frac{t_i-t_1}{\Delta t},
\qquad
j_i=\mathcal{B}(u_i),
\end{equation}
where $\Delta t=5$ minutes and $\mathcal{B}(\cdot)$ denotes the binning rule used during preprocessing. We use floor-based assignment for most datasets and nearest-index rounding only when it better matches the dataset's timestamp convention. Readings outside the valid range $[0,L-1]$ are excluded, and multiple readings assigned to the same grid position are averaged.

The aligned glucose sequence is denoted as $\hat{X}\in\mathbb{R}^{L}$, and the physical observation mask $M\in\{0,1\}^{L}$ indicates whether each grid position contains at least one real CGM reading. Missing positions are filled only for tensor construction; they are distinguished by the observation mask and are never treated as observed measurements. Thus, chronological grid alignment is separated from interpolation or imputation.

\subsection{Mask-aware Physiological Statistics}
\label{apd:mask_stats}

All feature construction is mask-aware: unobserved grid positions do not contribute to statistics or loss terms. \name divides each 24-hour window into $P=24$ non-overlapping one-hour patches, each containing $K=12$ grid positions. For patch $\mathcal{P}_i$, the physical observation support is
\begin{equation}
d_i =
\frac{1}{K}
\sum_{j\in\mathcal{P}_i} M_j .
\end{equation}
This patch density is reused in mask-aware loss weighting.

For the state stream, \name computes patch-level glucose statistics using only observed entries:
\begin{equation}
\mu_i =
\frac{
\sum_{j\in\mathcal{P}_i} M_j \hat{X}_j
}{
\sum_{j\in\mathcal{P}_i} M_j+\epsilon
},
\qquad
\sigma_i =
\left(
\frac{
\sum_{j\in\mathcal{P}_i} M_j(\hat{X}_j-\mu_i)^2
}{
\sum_{j\in\mathcal{P}_i} M_j+\epsilon
}
\right)^{1/2}.
\end{equation}
Empty patches are zeroed by the validity mask.

For the event stream, rate-of-change features are computed only from valid observed positions. The implementation searches backward up to 9 grid steps and uses the closest previous observed value:
\begin{equation}
r_j =
\frac{\hat{X}_j-\hat{X}_{j-b}}{b},
\end{equation}
where $j$ and $j-b$ are both observed and no closer valid pair exists. If no valid previous observation is found, the rate-of-change entry is set to zero and marked invalid. Patch-level event mean and standard deviation are then computed from valid rate-of-change entries only.

\subsection{Causal Gaussian State--Event Decoupling}
\label{apd:state_event_decoupling}

\name separates each aligned CGM sequence into a low-frequency state component and a high-frequency event component using a causal, mask-aware Gaussian filter. The filter is applied after mask-aware normalization. Let $\tilde{X}$ denote the normalized aligned glucose sequence and let $M$ denote the corresponding observation mask used by the branch. For each grid position $j$, the state component is estimated from current and past observed values:
\begin{equation}
\tilde{X}^{\mathrm{state}}_j
=
\frac{
\sum_{r=0}^{R}
K_{\sigma}(r)\,
M_{j-r}\,
\tilde{X}_{j-r}
}{
\sum_{r=0}^{R}
K_{\sigma}(r)\,
M_{j-r}
+\epsilon
},
\end{equation}
where invalid indices are ignored. The one-sided Gaussian kernel is
\begin{equation}
K_{\sigma}(r)
=
\frac{
\exp\left(-r^2/(2\sigma^2)\right)
}{
\sum_{u=0}^{R}
\exp\left(-u^2/(2\sigma^2)\right)
},
\qquad r=0,\ldots,R .
\end{equation}
Using only $r\geq 0$ makes the filter causal, so future glucose values are not used to estimate the current state. In the implementation, the maximum lag is determined by the maximum allowed bandwidth and a truncation factor of 3, giving $R=\lceil 3\sigma_{\max}\rceil=36$ grid steps.

The bandwidth $\sigma$ is learnable and constrained as
\begin{equation}
\sigma
=
\sigma_{\min}
+
(\sigma_{\max}-\sigma_{\min})
\cdot
\mathrm{sigmoid}(\rho),
\end{equation}
where $\rho$ is an unconstrained learnable parameter, $\sigma_{\min}=2$, and $\sigma_{\max}=12$. On the 5-minute grid, this corresponds to a learnable Gaussian scale of approximately 10--60 minutes, initialized at $\sigma=6$.

The event component is defined as the observed residual after removing the state trend:
\begin{equation}
\tilde{X}^{\mathrm{event}}
=
(\tilde{X}-\tilde{X}^{\mathrm{state}})\odot M .
\end{equation}
The state stream therefore receives smoothed baseline dynamics, while the event stream receives short-term deviations supported by real observations.

\subsection{Architecture and Dimensionality}
\label{apd:architecture_dimensionality}

Following the above patching scheme, the state stream encodes each patch using a $64$-dimensional waveform feature, a $16$-dimensional intra-patch trend-difference feature, and a $48$-dimensional projected statistics feature from patch mean and standard deviation. The event stream encodes each patch using a $48$-dimensional residual waveform feature, a $48$-dimensional rate-of-change feature, and a $32$-dimensional projected statistics feature from valid rate-of-change mean and standard deviation. These components are projected into $64$-dimensional state and event tokens, respectively.

The state and event tokens are fused into a unified physiological patch token of dimension $D=128$. Circular time-of-day features from the absolute grid index are projected to the same dimension and combined with patch positional embeddings through a learnable gate. The context and EMA target encoders share a $3$-layer Transformer architecture with $4$ attention heads and feed-forward dimension $256$. The masked-context predictor is a lightweight $1$-layer Transformer, and the temporal dynamics objective uses two lightweight transition heads. During downstream evaluation, only the frozen online branch is retained. \name has $0.72$M trainable parameters and $1.18$M total parameters during pretraining, mainly due to the additional EMA target branch.

\subsection{Pretraining Masking and Transition Loss}
\label{apd:masking_transition}

During masked contextual representation learning, \name applies random patch-level masking to the online branch, with the masking ratio sampled uniformly from $[0.5,0.6]$. Selected patches are hidden from the visible signal used for mask-aware statistics and filtering, and their patch tokens are replaced by a learnable mask token before the context encoder. The EMA target branch uses the full physical observation mask and encodes the complete sequence to provide latent targets. The masked contextual loss is applied only at masked patch positions and is weighted by each patch's physical observation density:
\begin{equation}
w_i=d_i=
\frac{1}{K}
\sum_{j\in\mathcal{P}_i} M_j .
\end{equation}

For temporal dynamics modeling, \name predicts next-patch state and event targets from the current online state and event tokens before Transformer self-attention. Let $m_i^{\mathrm{mask}}\in\{0,1\}$ indicate whether patch $i$ is masked, and let $d_i$ denote its physical observation density. The transition weight is
\begin{equation}
q_i =
(1-m_i^{\mathrm{mask}})\, d_i\, d_{i+1},
\qquad i=1,\ldots,P-1 .
\end{equation}
This excludes transitions starting from masked context patches and down-weights transitions involving sparsely observed adjacent patches. The EMA next-patch targets are also taken before Transformer self-attention, preventing future contextual information from leaking into the transition objective. In the main experiments, \name optimizes the masked contextual representation loss and temporal dynamics loss with both loss weights set to $1.0$.

\subsection{Downstream Feature Extraction}
\label{apd:downstream_feature}

For downstream evaluation, we freeze the pretrained online branch of \name and discard the EMA target branch and pretraining-only prediction heads. Each one-day CGM window is passed through the online state--event preprocessing modules, unified embedder, and context encoder to obtain patch representations $\{z_i\}_{i=1}^{P}$, where $P=24$. We use global average pooling over patches to obtain a fixed-length window representation.

\subsection{Data Augmentation Details}
\label{apd:data_augmentation}

During pretraining, \name uses two augmentation families: \textit{value perturbations} and \textit{structural sparsification}. Value perturbations preserve the observation mask while modifying observed glucose values. Baseline wander is applied with probability $0.25$ by adding a sinusoidal perturbation with amplitude sampled from $[5,15]$ mg/dL and frequency from $[0.5,2.0]$ cycles per window. Compression-like drops are applied with probability $0.10$ by attenuating a contiguous $6$--$12$-step segment using a V-shaped curve with minimum multiplier sampled from $[0.4,0.7]$.

Structural sparsification alters the observation mask itself. Decimation is applied with probability $0.40$ to dense windows with more than $200$ observed positions by keeping every third observation from a random 5-minute offset, producing a 15-minute-like sampling pattern. Disconnection blocks are applied with probability $0.05$ by removing $1$--$3$ contiguous blocks of $2$--$12$ grid steps. Candidate augmentations are evaluated in random order; after one is applied, subsequent probabilities are multiplied by $0.25$ to avoid excessive corruption.

\section{Task Setup and Additional Results}
\label{apd:full_results}

\subsection{Subject-Disjoint Linear Probing Details}
\label{app:linear_probe_details}

\input{tables/appendix_table_linear_probe}
We provide additional details for the subject-grouped linear probing protocol used in Table~\ref{tab:downstream_results}. For all methods, the pretrained encoder is frozen and only a linear classifier is trained on the extracted 24-hour window representations. We use the same scikit-learn logistic regression classifier with L2 regularization for all models, using the \texttt{lbfgs} solver, a maximum of 1000 iterations, and a fixed random seed. This ensures that performance differences mainly reflect the quality of the frozen representations rather than downstream classifier capacity or optimization choices.

For each dataset and task, we perform 5-fold subject-grouped cross-validation and repeat the procedure for 10 iterations with different fold assignments. All windows from the same subject are assigned to the same fold, ensuring that training and test subjects never overlap. The same subject splits are used for all methods within each task, enabling a paired comparison across representations. All preprocessing statistics and downstream classifiers are fit only on the training subjects in each fold, and evaluation metrics are computed over held-out test windows. We report mean performance in the main paper for readability, and provide the full mean $\pm$ standard deviation across repeated folds in Table~\ref{apd:full_linear_probe}.

\paragraph{Paired statistical analysis.}
We quantify uncertainty in the task-averaged linear-probing comparisons against \cgmjepa and \cgmformer using unrounded fold-level scores. For each metric and repeat/fold, we macro-average the 14 task--dataset scores and form the paired difference between \name and the comparator, yielding $n=50$ paired differences from 10 repeats of five-fold cross-validation. We use the two-sided Nadeau--Bengio corrected repeated-cross-validation paired $t$ test~\cite{NIPS1999_7d12b66d}, with $n_{\mathrm{test}}/n_{\mathrm{train}}=1/4$ and 95\% CIs. Holm adjustment is applied across six comparisons: two comparators by three metrics.

\input{tables/appendix_table_linear_probe_significance_overall}

All six corrected CIs exclude zero, and all six tests remain significant after Holm adjustment. The effects are also directionally broad: \name improves over \cgmjepa on 11/14, 12/14, and 11/14 evaluations and over \cgmformer on 11/14, 11/14, and 9/14 evaluations for PR-AUC, ROC-AUC, and Macro-F1, respectively.

\subsection{Meal-Aligned PPGR Details}
\label{app:ppgr_details}

We construct the two-hour postprandial glycemic response (PPGR) task from paired Dexcom and Libre recordings in \cgmacros. After requiring a strictly causal 24-hour input window and no additional logged meal in $(0,120]$ minutes after meal onset, each sensor contributes the same 874 meals from 34 participants. Meal timestamps and nutrition come from \texttt{CGMacros\_Meal}, but its interpolated CGM columns are not used as model inputs or prediction targets. Instead, pre-meal CGM and response targets are constructed from the original raw Dexcom and Libre exports. The meal-start baseline is the latest raw observation at or before meal onset, with a maximum age of 5 minutes for Dexcom and 15 minutes for Libre. Post-meal targets are nearest raw observations within 2.5 minutes of each Dexcom horizon and 7.5 minutes of each Libre horizon, without target interpolation.

Dexcom and Libre are trained and evaluated separately at their native target cadences. The same MLP with two hidden layers predicts 24 glucose changes at 5-minute intervals from 5 to 120 minutes for Dexcom and 8 changes at 15-minute intervals from 15 to 120 minutes for Libre. Each frozen representation is combined cumulatively with the last hour of pre-meal CGM, five nutrition variables, fasting glucose, and BMI and diabetes status. We use five dataset-defined subject-group outer folds and 10 head initializations. One repetition concatenates held-out predictions from all five folds before evaluation; the reported two-sensor results average the sensor-specific endpoint errors with equal weight.

Trajectory MAE is computed over all meals and native horizons within each sensor. Positive iAUC is computed per meal as
\begin{equation}
\operatorname{iAUC}^{+}=\int_{0}^{2\mathrm{h}}\max\!\left(\Delta g(t),0\right)\,dt,
\end{equation}
using meal-start glucose as the baseline and trapezoidal integration after prepending $\Delta g(0)=0$ to each sensor's native grid; its MAE is reported in mg/dL$\cdot$h. Peak-rise MAE compares the maximum predicted and observed glucose changes on the corresponding native grid. Peak-time MAE compares their first argmax times. We use trajectory, positive iAUC, and peak rise as the main endpoints; peak time is reported as a complementary timing measure.

\input{tables/appendix_table_ppgr_by_device}

At the complete-context setting, \name ranks first on all four endpoints for both Dexcom and Libre (Table~\ref{tab:ppgr_by_device}).

\subsection{Few-shot Adaptation Details}
\label{app:few_shot_details}

We provide additional details for the few-shot adaptation protocol used in Figure~\ref{fig:few_shot}. For each pretrained model, we freeze the encoder and extract representations for all labeled 24-hour windows in each dataset. We perform 5-fold subject-grouped cross-validation with 10 repeated iterations, using the same subject splits across all methods. All windows from the same dataset-defined subject are assigned to the same
fold, ensuring that training and test grouping units never overlap.

In the \textit{limited-subject} setting, we sample exactly $K \in \{1,2,3,4,5\}$ labeled support subjects per class from the training fold. The same logistic regression classifier is then trained on all extracted window representations from the selected support subjects. In the \textit{limited-observation} setting, we retain all training subjects but randomly subsample each subject's 24-hour training windows at fractions $\{1\%,5\%,10\%,20\%,30\%,40\%,50\%\}$. For both settings, each fold and support configuration is evaluated with 5 random samplings, and metrics are computed over held-out test windows. Figure~\ref{fig:few_shot} reports the broader PR-AUC comparison, and Table~\ref{tab:few_shot_summary} provides exact task-averaged PR-AUC, ROC-AUC, and Macro-F1 values for representative CGM-specific models.

\input{tables/appendix_table_few_shot_summary}

Across all evaluated budgets, \name achieves the strongest task-averaged performance on all three metrics among the models in Table~\ref{tab:few_shot_summary}.

\subsection{Cross-Dataset Generalization Details}
\label{app:cross_dataset_details}

We provide additional details for the cross-dataset generalization protocol used in Table~\ref{tab:cross_dataset}. Unlike subject-disjoint linear probing, this experiment does not use cross-validation because the goal is to evaluate direct transfer across distinct cohorts. For each transfer direction, we freeze the pretrained encoder, extract representations from all labeled 24-hour windows in the source and target datasets, train a logistic regression classifier on the full source dataset, and evaluate it directly on the target dataset. The target dataset is used only for final evaluation; no target labels are used for training, validation, model selection, or threshold tuning. The same source--target splits, frozen representations, classifier type, and preprocessing protocol are used for all compared methods.

We evaluate diabetes risk assessment and insulin resistance because they are consistently available across \cgmacros, \stanford, and \hall. To ensure label compatibility, we harmonize task definitions before transfer. For diabetes risk assessment, \cgmacros originally includes normoglycemic, prediabetes, and type 2 diabetes categories; we convert it into a binary risk label by grouping prediabetes and type 2 diabetes as the positive class and normoglycemic status as the negative class, matching the binary diabetes-risk labels in \stanford and \hall. For insulin resistance, we use the dataset-provided binary labels while preserving the same positive/negative semantics across cohorts. This protocol evaluates whether frozen representations support transferable clinical decision boundaries across cohorts rather than relying only on dataset-specific label patterns.

\subsection{Multiday Representation Observation Details}
\label{app:multiday_details}

We provide additional details for the multiday representation observation analysis in Figure~\ref{fig:multiday_observation}. The goal is to evaluate whether observing more days of CGM improves subject-level prediction using frozen \name representations. For each subject, CGM recordings are divided into non-overlapping 24-hour windows and aligned to the \name input format, a 288-point 5-minute chronological grid. A day is included only if it satisfies the preprocessing quality criteria, including a maximum consecutive missing interval of less than one hour. For each valid day, we extract one embedding using the frozen pretrained \name encoder. We then select one fixed eligible $K_{\max}$-day anchor episode for each subject. \stanford, \cgmacros, and \shanghai use $K_{\max}=7$, while \hall uses $K_{\max}=4$ due to the available near-continuous recordings. Within each fixed anchor, we enumerate adjacent $K$-day subwindows for $K=1,\ldots,K_{\max}$.
For each $K$-day subwindow, we aggregate the frozen daily embeddings into one subject-level representation. We evaluate two aggregation variants. The first uses mean pooling:
\begin{equation}
\mathbf{z}^{\mathrm{mean}}_{i,K,s}
=
\frac{1}{K}
\sum_{d=s}^{s+K-1}
\mathbf{h}_{i,d},
\end{equation}
where $\mathbf{h}_{i,d}$ is the frozen daily embedding for subject $i$ on day $d$, and $s$ is the subwindow start position. The second uses concat(mean, max) pooling:
\begin{equation}
\mathbf{z}^{\mathrm{mean+max}}_{i,K,s}
=
\left[
\mathrm{mean}_{d=s}^{s+K-1}(\mathbf{h}_{i,d});
\mathrm{max}_{d=s}^{s+K-1}(\mathbf{h}_{i,d})
\right],
\end{equation}
which preserves both the average daily pattern and salient embedding dimensions across the observation window.

\input{tables/appendix_table_multiday_exp}

For each $K$ and start position, every subject contributes exactly one representation, and each test subject receives exactly one prediction. Thus, adjacent subwindows from the same subject are not treated as independent test samples. We train linear probes with 10 repeated iterations of 5-fold stratified subject-level cross-validation. Metrics are computed separately for each start position and averaged within each repeated evaluation. In \cgmacros, Dexcom, Libre, and Fused are all evaluated on the same 42-subject fused-overlap 7-day anchor cohort. Dexcom and Libre are evaluated separately, while the matched fused setting averages same-subject same-day embeddings from both sensors before multiday aggregation. We report the paired PR-AUC change relative to the one-day representation: $\Delta_K=\mathrm{PR\text{-}AUC}(K)-\mathrm{PR\text{-}AUC}(1)$. Deltas are computed within the same repeated evaluation, and confidence intervals are estimated over repeat-level paired deltas.

Table~\ref{tab:multiday_pooling_summary} reports the full multiday PR-AUC results for both aggregation variants; longer observation windows improve subject-level prediction in most settings.

\subsection{Pretraining Data Scalability Details}
\label{app:pretrain_scale_details}

We provide additional details for the pretraining data scalability analysis in Figure~\ref{fig:scale}. To assess how \name benefits from unlabeled CGM data, we construct reduced pretraining corpora by randomly sampling subjects from each pretraining cohort at ratios of $20\%$, $40\%$, $60\%$, and $80\%$. Subject-level sampling is performed within each cohort to preserve the overall cohort composition while varying the amount of available unlabeled data. For each ratio, we repeat the sampling with five random seeds and pretrain a separate \name model from scratch using the same architecture, optimization settings, masking strategy, and downstream evaluation protocol as the full-data model. This isolates the effect of pretraining data scale from changes in model capacity or downstream classifier settings.

Figure~\ref{fig:scale} shows that task-averaged performance improves as the amount of unlabeled pretraining data increases.

\section{Ablation Design Details}
\label{apd:ablation}
\subsection{Encoder Design Ablation Details}
\label{app:encoder_design_ablation}

\input{tables/appendix_table_encoder_design}

Table~\ref{apd:encoder_design} provides task-wise results for the encoder design ablation summarized in Table~\ref{tab:stream_ablation}. All variants use the same pretraining data, optimization settings, masking strategy, and downstream evaluation protocol. To ensure a fair comparison, we do not remove auxiliary patch-level information from the ablated encoders. For each patch, we compute mask-aware physiological statistics consisting of state statistics, i.e., glucose mean and standard deviation, and event statistics, i.e., mean rate-of-change and rate-of-change standard deviation. The rate-of-change features are computed with a mask-aware dynamic backoff and normalized to the 5-minute grid.

For fairness, all variants are provided with the corresponding patch-level statistics and temporal-difference features. The variants differ only in how the waveform and auxiliary patch features are organized. The \textit{Raw Input} variant directly tokenizes the aligned glucose sequence and receives both patch-internal differences and the full set of patch statistics, including glucose mean/std and rate-of-change mean/std. The \textit{State-stream Only} variant uses the filtered low-frequency trend with trend differences and state statistics. The \textit{Event-stream Only} variant uses the residual component with rate-of-change features and event statistics. The full \textit{Dual-stream} model encodes the state and event streams separately before fusing them into a unified daily representation.

As shown in Table~\ref{apd:encoder_design}, the dual-stream encoder achieves the strongest task-averaged performance, supporting the benefit of integrating the two temporal streams.

\subsection{Pre-fusion Stream Content Probe Details}
\label{app:prefusion_content_probes}

We analyze the frozen 64-dimensional state and event tokens immediately before cross-stream fusion from the same pretrained checkpoint, without updating the encoder. The analysis uses 7,435 high-coverage hourly Dexcom patches from 35 \cgmacros participants. Identical Ridge probes predict either mask-aware hourly mean glucose, which measures retained glucose-level content, or mean absolute adjacent 5-minute change, which measures retained local-dynamics content. Both streams use the same five repeated subject-disjoint 5-fold splits, and point estimates are means of repeat-wise out-of-fold $R^2$. The paired directional gaps reported in Table~\ref{tab:stream_ablation} use 2,000 subject-cluster bootstrap resamples while preserving the repeated evaluation structure.

\input{tables/appendix_table_ablation_summary}

\subsection{Temporal Dynamics Weight Ablation Details}
\label{app:temporal_dynamics_weight}

In the temporal dynamics weight ablation in Figure~\ref{fig:abl_lambda}, we keep the architecture, pretraining data, masking strategy, augmentation pipeline, and downstream evaluation protocol fixed, and vary only the weight $\lambda_{\mathrm{TD}}$ of the temporal dynamics objective. The total pretraining loss is
$\mathcal{L}=\mathcal{L}_{\mathrm{MCR}}+\lambda_{\mathrm{TD}}\mathcal{L}_{\mathrm{TD}}$,
where $\mathcal{L}_{\mathrm{MCR}}$ denotes masked contextual latent prediction and $\mathcal{L}_{\mathrm{TD}}$ denotes temporal dynamics modeling. We sweep $\lambda_{\mathrm{TD}}$ from $0$ to $2.0$, with $\lambda_{\mathrm{TD}}=0$ corresponding to masked contextual prediction alone.

Task-averaged performance is strongest across a broad range around $\lambda_{\mathrm{TD}}=0.6$--$1.0$ (Table~\ref{tab:ablation_summary}).

\subsection{Data Augmentation Ablation Details}
\label{app:data_augmentation_ablation}

For the data augmentation ablation in Figure~\ref{fig:abl_aug}, we keep the architecture, pretraining objectives, masking strategy, and downstream evaluation protocol fixed, and vary only the training-time augmentation pipeline. The \textit{No Aug.} setting removes all augmentations. The \textit{Value Perturb.} setting includes value-level perturbations, such as low-frequency baseline wander and compression-like transient drops. The \textit{Struct. Spars.} setting includes structural sparsification, such as decimation and sensor-disconnection blocks. The full setting combines both augmentation families and achieves the strongest task-averaged performance (Table~\ref{tab:ablation_summary}).

\subsection{Dense Interpolation Ablation Details}
\label{app:dense_interpolation_ablation}

For the dense interpolation ablation in Figure~\ref{fig:abl_intp}, we compare three preprocessing designs while keeping the architecture, pretraining objectives, and downstream protocol fixed. The \textit{Dense Interp.} variant linearly interpolates missing grid positions during both pretraining and downstream representation extraction, and treats the resulting sequence as densely observed. The \textit{Dense Interp. + Aug.} variant uses the same dense interpolation protocol but further applies structural sparsification during pretraining to reintroduce missingness-like perturbations. The default \textit{No Interp.} setting preserves the original observation mask and does not treat interpolated values as real measurements, and achieves the strongest task-averaged performance (Table~\ref{tab:ablation_summary}).

%% file: tables/appendix_table_dataset.tex
\begin{table*}[!t]
\centering
\caption{Label distributions for downstream clinical prediction tasks across four cohorts; \shanghai counts denote labeled recording sessions.}
\label{tab:label_distributions}
\small
\renewcommand{\arraystretch}{1.05}
\setlength{\tabcolsep}{3pt}

\begin{minipage}[t]{0.46\textwidth}
\centering
\textbf{\cgmacros (N=45)}\\[0.2em]
\begin{tabular}{llc}
\toprule
\textbf{Task} & \textbf{Class} & \textbf{Count} \\
\midrule
\multirow{3}{*}{Diabetes Risk} 
& Normoglycemic & 16 \\
& Prediabetes & 14 \\
& Type 2 Diabetes & 15 \\
\midrule
\multirow{2}{*}{Insulin Resistance} 
& Sensitive & 13 \\
& Resistant & 32 \\
\midrule
\multirow{2}{*}{Obesity} 
& Non-obese & 22 \\
& Obese & 23 \\
\midrule
\multirow{2}{*}{Hyperlipidemia} 
& Normal & 33 \\
& Hyperlipidemia & 12 \\
\bottomrule
\end{tabular}
\end{minipage}
\hfill
\begin{minipage}[t]{0.46\textwidth}
\centering
\textbf{\hall (N=56)}\\[0.2em]
\begin{tabular}{llc}
\toprule
\textbf{Task} & \textbf{Class} & \textbf{Count} \\
\midrule
\multirow{2}{*}{Diabetes} 
& Normoglycemic & 37 \\
& Abnormal & 19 \\
\midrule
\multirow{2}{*}{Glucotype} 
& Normal & 34 \\
& Severe & 22 \\
\midrule
\multirow{2}{*}{Insulin Resistance} 
& Sensitive & 35 \\
& Resistant & 21 \\
\midrule
\multirow{2}{*}{Hyperlipidemia} 
& Normal & 48 \\
& Hyperlipidemia & 8 \\
\bottomrule
\end{tabular}
\end{minipage}

\vspace{0.5em}

\begin{minipage}[t]{0.46\textwidth}
\centering
\textbf{\stanford (N=37)}\\[0.2em]
\begin{tabular}{llc}
\toprule
\textbf{Task} & \textbf{Class} & \textbf{Count} \\
\midrule
\multirow{2}{*}{Diabetes} 
& Normoglycemic & 17 \\
& Abnormal & 20 \\
\midrule
\multirow{2}{*}{$\beta$-cell Dysfunction} 
& Normal & 17 \\
& Dysfunction & 20 \\
\midrule
\multirow{2}{*}{Insulin Resistance} 
& Sensitive & 17 \\
& Resistant & 20 \\
\bottomrule
\end{tabular}
\end{minipage}
\hfill
\begin{minipage}[t]{0.46\textwidth}
\centering
\textbf{\shanghai (N=65)}\\[0.2em]
\begin{tabular}{llc}
\toprule
\textbf{Task} & \textbf{Class} & \textbf{Count} \\
\midrule
\multirow{2}{*}{Hypoglycemia} 
& No & 56 \\
& Yes & 9 \\
\midrule
\multirow{2}{*}{Insulin Resistance} 
& Sensitive & 24 \\
& Resistant & 41 \\
\midrule
\multirow{2}{*}{Hyperlipidemia} 
& Normal & 45 \\
& Hyperlipidemia & 20 \\
\bottomrule
\end{tabular}
\end{minipage}
\end{table*}

%% file: tables/appendix_table_linear_probe.tex
\begin{table*}[!t]
\centering
\caption{Full subject-disjoint linear-probe performance. All reported values represent the mean $\pm$ std evaluated via 10 iterations of 5-fold subject-grouped cross-validation.}
\label{apd:full_linear_probe}
\resizebox{\textwidth}{!}{
\begin{tabular}{l|l | >{\columncolor{gray!10}}c>{\columncolor{gray!10}}c>{\columncolor{gray!10}}c>{\columncolor{gray!10}}c ccc >{\columncolor{gray!10}}c>{\columncolor{gray!10}}c>{\columncolor{gray!10}}c cccc}
\toprule

\multirow{2}{*}{\textbf{Method}} & \multirow{2}{*}{\textbf{Metrics}} &
\multicolumn{4}{c}{\cellcolor{gray!10}\textbf{\cgmacros}} & 
\multicolumn{3}{c}{\textbf{\shanghai}} & 
\multicolumn{3}{c}{\cellcolor{gray!10}\textbf{\stanford}} & 
\multicolumn{4}{c}{\textbf{\hall}} \\

\cmidrule(lr){3-6} \cmidrule(lr){7-9} \cmidrule(lr){10-12} \cmidrule(lr){13-16}

& &  
\rotatebox{0}{Diabetes} & \rotatebox{0}{IR} & \rotatebox{0}{Hyperlip.} & \rotatebox{0}{Obesity} & 
\rotatebox{0}{IR} & \rotatebox{0}{Hyperlip.} & \rotatebox{0}{Hypogly.} & 
\rotatebox{0}{Diabetes} & \rotatebox{0}{$\beta$-cell Dys.} & \rotatebox{0}{IR} & 
\rotatebox{0}{Diabetes} & \rotatebox{0}{IR} & \rotatebox{0}{Hyperlip.} & \rotatebox{0}{Glucotype} \\

\midrule
\midrule

\rowcolor{gray!10}
\multicolumn{16}{l}{\textit{General Time-series Foundation Models with Large-scale Pretraining}} \\

\multirow{3}{*}{\begin{tabular}{@{}l@{}}\chronos \\ (small)\end{tabular}} 
& PR  & 51.6 $\pm$ 4.5 & 82.7 $\pm$ 6.5 & 30.5 $\pm$ 6.5 & 56.0 $\pm$ 4.8 & 65.0 $\pm$ 5.4 & 34.8 $\pm$ 6.0 & 16.0 $\pm$ 4.3 & 69.2 $\pm$ 6.6 & 58.6 $\pm$ 5.1 & 62.5 $\pm$ 5.0 & 51.7 $\pm$ 10.3 & 44.9 $\pm$ 7.0 & 24.2 $\pm$ 6.8 & 54.2 $\pm$ 8.2 \\
& AUC & 68.0 $\pm$ 3.7 & 67.5 $\pm$ 10.4 & 52.1 $\pm$ 5.9 & 53.7 $\pm$ 5.1 & 55.0 $\pm$ 6.0 & 50.6 $\pm$ 6.6 & 49.1 $\pm$ 8.8 & 65.4 $\pm$ 8.0 & 57.3 $\pm$ 6.3 & 64.2 $\pm$ 5.3 & 64.9 $\pm$ 9.7 & 57.7 $\pm$ 7.7 & \textbf{61.3} $\pm$ 9.3 & 62.6 $\pm$ 6.9 \\
& F1  & 49.4 $\pm$ 4.8 & 61.7 $\pm$ 7.7 & 50.1 $\pm$ 4.1 & 52.7 $\pm$ 4.3 & 52.6 $\pm$ 4.3 & 51.5 $\pm$ 4.6 & 50.5 $\pm$ 3.8 & 61.3 $\pm$ 6.2 & 54.8 $\pm$ 4.8 & 59.6 $\pm$ 3.8 & 60.9 $\pm$ 7.1 & 55.0 $\pm$ 6.1 & 52.5 $\pm$ 6.5 & 58.6 $\pm$ 6.5 \\
\midrule

\multirow{3}{*}{\chronos} 
& PR  & 54.3 $\pm$ 4.7 & 84.1 $\pm$ 5.8 & \underline{34.0} $\pm$ 9.1 & 58.1 $\pm$ 5.4 & 59.8 $\pm$ 5.3 & 34.7 $\pm$ 5.6 & 18.9 $\pm$ 7.8 & 68.5 $\pm$ 6.8 & 58.5 $\pm$ 4.3 & 59.2 $\pm$ 4.2 & 48.6 $\pm$ 9.2 & 51.3 $\pm$ 9.5 & 20.6 $\pm$ 8.5 & 54.4 $\pm$ 9.3 \\
& AUC & 69.7 $\pm$ 3.9 & 69.9 $\pm$ 9.9 & \textbf{55.8} $\pm$ 7.9 & 56.6 $\pm$ 6.9 & 49.5 $\pm$ 7.2 & 49.7 $\pm$ 7.8 & 53.0 $\pm$ 10.8 & 65.6 $\pm$ 7.8 & 57.8 $\pm$ 5.5 & 60.2 $\pm$ 5.3 & 60.4 $\pm$ 9.2 & 62.7 $\pm$ 8.2 & 50.3 $\pm$ 10.7 & 63.0 $\pm$ 8.6 \\
& F1  & 50.2 $\pm$ 4.4 & 63.3 $\pm$ 7.5 & \textbf{53.7} $\pm$ 4.6 & 54.2 $\pm$ 5.9 & 49.0 $\pm$ 5.6 & 50.4 $\pm$ 5.4 & 52.8 $\pm$ 5.6 & 61.9 $\pm$ 5.6 & 55.5 $\pm$ 4.2 & 56.6 $\pm$ 4.2 & 58.4 $\pm$ 7.9 & 58.2 $\pm$ 6.3 & 50.9 $\pm$ 4.9 & 56.2 $\pm$ 7.8 \\

\midrule

\multirow{3}{*}{\begin{tabular}{@{}l@{}}\moment \\ (small)\end{tabular}} 
& PR  & 54.0 $\pm$ 4.8 & 83.8 $\pm$ 5.8 & 30.3 $\pm$ 8.5 & 60.9 $\pm$ 6.3 & 58.3 $\pm$ 5.3 & 38.3 $\pm$ 7.4 & 16.3 $\pm$ 5.3 & 70.4 $\pm$ 7.4 & 57.2 $\pm$ 5.2 & 63.0 $\pm$ 5.6 & 59.2 $\pm$ 9.7 & 53.1 $\pm$ 8.2 & 19.4 $\pm$ 6.7 & 55.6 $\pm$ 10.0 \\
& AUC & 69.1 $\pm$ 4.3 & 70.6 $\pm$ 9.6 & 51.5 $\pm$ 10.3 & 58.7 $\pm$ 6.5 & 49.0 $\pm$ 6.9 & 55.9 $\pm$ 7.6 & 52.9 $\pm$ 9.8 & 67.6 $\pm$ 8.8 & 55.5 $\pm$ 6.5 & 65.1 $\pm$ 6.3 & 69.7 $\pm$ 8.3 & 62.1 $\pm$ 8.8 & 48.7 $\pm$ 10.3 & 61.7 $\pm$ 8.7 \\
& F1  & 49.8 $\pm$ 4.7 & 63.6 $\pm$ 7.7 & 50.9 $\pm$ 6.1 & 55.7 $\pm$ 5.5 & 49.3 $\pm$ 4.8 & 52.9 $\pm$ 5.8 & 51.4 $\pm$ 4.6 & 62.5 $\pm$ 7.5 & 53.8 $\pm$ 5.0 & 60.1 $\pm$ 4.5 & 64.5 $\pm$ 6.9 & 59.0 $\pm$ 6.8 & 50.4 $\pm$ 6.5 & 59.3 $\pm$ 7.7 \\

\midrule

\multirow{3}{*}{\begin{tabular}{@{}l@{}}\moment \\ (large)\end{tabular}} 
& PR  & 50.3 $\pm$ 5.0 & 87.4 $\pm$ 5.1 & 32.0 $\pm$ 7.9 & 61.8 $\pm$ 5.7 & 59.6 $\pm$ 5.9 & \textbf{42.4} $\pm$ 6.7 & 14.3 $\pm$ 3.4 & 71.9 $\pm$ 5.3 & 61.8 $\pm$ 5.2 & 58.1 $\pm$ 4.3 & 52.1 $\pm$ 9.4 & 44.9 $\pm$ 10.1 & 18.4 $\pm$ 5.7 & 55.8 $\pm$ 7.5 \\
& AUC & 66.1 $\pm$ 4.6 & 75.2 $\pm$ 9.7 & 53.2 $\pm$ 8.3 & 59.9 $\pm$ 6.8 & 47.8 $\pm$ 7.5 & \textbf{60.8} $\pm$ 6.3 & 49.6 $\pm$ 8.1 & 69.1 $\pm$ 6.4 & 58.9 $\pm$ 5.4 & 59.4 $\pm$ 4.5 & 65.0 $\pm$ 8.0 & 56.6 $\pm$ 8.8 & 51.4 $\pm$ 12.2 & 63.2 $\pm$ 7.6 \\
& F1  & 46.8 $\pm$ 5.4 & 67.2 $\pm$ 8.2 & \underline{52.5} $\pm$ 4.9 & 56.3 $\pm$ 4.8 & 48.6 $\pm$ 5.5 & \textbf{57.2} $\pm$ 5.3 & 49.7 $\pm$ 3.5 & 63.7 $\pm$ 5.0 & 55.7 $\pm$ 4.1 & 56.1 $\pm$ 3.0 & 60.5 $\pm$ 6.5 & 53.3 $\pm$ 7.5 & 47.7 $\pm$ 5.9 & 57.8 $\pm$ 6.4 \\

\midrule

\multirow{3}{*}{\mantis} 
& PR  & 61.8 $\pm$ 5.9 & 90.9 $\pm$ 5.5 & 30.1 $\pm$ 10.0 & 63.4 $\pm$ 8.1 & 59.2 $\pm$ 5.5 & 30.6 $\pm$ 4.5 & \textbf{22.9} $\pm$ 12.4 & 75.1 $\pm$ 5.4 & 63.5 $\pm$ 9.0 & \underline{67.1} $\pm$ 7.3 & 54.2 $\pm$ 11.5 & 55.7 $\pm$ 12.4 & \underline{24.8} $\pm$ 9.7 & 77.5 $\pm$ 7.6 \\
& AUC & 75.5 $\pm$ 5.1 & \underline{80.3} $\pm$ 10.9 & 49.8 $\pm$ 13.6 & 60.8 $\pm$ 9.8 & 48.2 $\pm$ 7.3 & 46.0 $\pm$ 7.6 & 58.6 $\pm$ 17.5 & \underline{71.4} $\pm$ 5.2 & 61.5 $\pm$ 9.3 & \underline{68.2} $\pm$ 5.9 & 67.1 $\pm$ 10.1 & 66.2 $\pm$ 10.4 & 57.9 $\pm$ 10.5 & 82.6 $\pm$ 5.7 \\
& F1  & 56.0 $\pm$ 6.6 & \textbf{71.6} $\pm$ 8.3 & 49.8 $\pm$ 6.9 & 57.2 $\pm$ 6.9 & 48.4 $\pm$ 5.3 & 47.5 $\pm$ 5.5 & \textbf{53.8} $\pm$ 7.7 & \underline{65.9} $\pm$ 4.8 & 58.5 $\pm$ 6.3 & \underline{62.3} $\pm$ 4.5 & 59.6 $\pm$ 8.9 & 59.8 $\pm$ 8.9 & \textbf{53.7} $\pm$ 6.7 & 73.3 $\pm$ 6.1 \\

\midrule

\multirow{3}{*}{\mantisv}
& PR  & \underline{63.6} $\pm$ 6.2 & \underline{91.2} $\pm$ 4.2 & 30.3 $\pm$ 10.0 & 62.8 $\pm$ 7.1 & 64.1 $\pm$ 5.9 & 33.3 $\pm$ 6.6 & \underline{22.0} $\pm$ 12.8 & 75.2 $\pm$ 6.7 & \underline{65.7} $\pm$ 7.2 & 64.6 $\pm$ 9.1 & 57.7 $\pm$ 14.4 & \underline{59.3} $\pm$ 13.6 & \textbf{25.0} $\pm$ 9.9 & 81.5 $\pm$ 7.8 \\
& AUC & 76.7 $\pm$ 5.2 & \underline{80.3} $\pm$ 9.0 & 49.1 $\pm$ 13.6 & 60.3 $\pm$ 9.4 & 52.7 $\pm$ 6.8 & 48.5 $\pm$ 8.6 & \textbf{59.6} $\pm$ 20.9 & 71.2 $\pm$ 7.9 & \underline{64.4} $\pm$ 7.2 & 65.5 $\pm$ 9.3 & 67.5 $\pm$ 11.8 & 66.7 $\pm$ 11.1 & \underline{58.8} $\pm$ 9.5 & 84.7 $\pm$ 6.0 \\
& F1  & 56.2 $\pm$ 7.0 & 68.9 $\pm$ 9.0 & 49.2 $\pm$ 7.6 & 55.9 $\pm$ 7.2 & 51.2 $\pm$ 5.6 & 48.1 $\pm$ 6.4 & \underline{53.3} $\pm$ 10.1 & 65.7 $\pm$ 7.7 & \underline{58.6} $\pm$ 5.5 & 60.4 $\pm$ 7.5 & 61.1 $\pm$ 9.3 & \underline{60.8} $\pm$ 8.9 & 52.6 $\pm$ 8.0 & 76.7 $\pm$ 7.3 \\

\midrule
\midrule

\rowcolor{gray!10}
\multicolumn{16}{l}{\textit{Externally Pretrained CGM Foundation Model}} \\

\multirow{3}{*}{\cgmformer} 
& PR  & 63.3 $\pm$ 6.4 & 89.9 $\pm$ 4.6 & 31.8 $\pm$ 8.6 & \textbf{68.1} $\pm$ 7.3 & 54.8 $\pm$ 3.9 & \underline{40.6} $\pm$ 9.2 & 11.9 $\pm$ 3.0 & \underline{75.9} $\pm$ 5.9 & 62.2 $\pm$ 8.6 & 61.8 $\pm$ 5.6 & 51.1 $\pm$ 11.4 & 48.0 $\pm$ 10.7 & 18.0 $\pm$ 7.9 & 79.6 $\pm$ 7.5 \\
& AUC & \underline{77.1} $\pm$ 4.7 & 78.0 $\pm$ 9.5 & 54.2 $\pm$ 10.8 & \textbf{66.0} $\pm$ 8.2 & 43.6 $\pm$ 6.3 & \underline{58.4} $\pm$ 10.3 & 42.1 $\pm$ 10.4 & \underline{71.4} $\pm$ 7.2 & 59.2 $\pm$ 8.5 & 63.2 $\pm$ 4.3 & 66.6 $\pm$ 10.1 & 58.2 $\pm$ 10.2 & 46.5 $\pm$ 20.7 & 84.3 $\pm$ 5.3 \\
& F1  & \underline{57.2} $\pm$ 5.4 & 67.6 $\pm$ 7.6 & 52.1 $\pm$ 6.4 & \textbf{60.5} $\pm$ 6.4 & 45.5 $\pm$ 5.1 & \underline{55.1} $\pm$ 6.3 & 45.3 $\pm$ 4.0 & \textbf{66.2} $\pm$ 5.8 & 55.7 $\pm$ 6.6 & 59.0 $\pm$ 3.0 & 59.3 $\pm$ 8.8 & 56.3 $\pm$ 7.9 & 48.4 $\pm$ 8.1 & 77.1 $\pm$ 5.6 \\

\midrule
\midrule

\rowcolor{gray!10}
\multicolumn{16}{l}{\textit{CGM Foundation Models Retrained on Our Pretraining Corpus}} \\

\multirow{3}{*}{\cgmjepa} 
& PR  & 63.0 $\pm$ 7.5 & 86.2 $\pm$ 7.5 & 28.7 $\pm$ 9.6 & 55.5 $\pm$ 7.4 & \textbf{69.1} $\pm$ 7.5 & 37.4 $\pm$ 8.2 & 17.8 $\pm$ 5.5 & 66.4 $\pm$ 4.1 & 58.7 $\pm$ 5.9 & 61.5 $\pm$ 7.6 & \underline{59.9} $\pm$ 12.1 & 56.8 $\pm$ 14.2 & 17.3 $\pm$ 9.8 & \underline{87.6} $\pm$ 5.9 \\
& AUC & 75.9 $\pm$ 6.0 & 73.8 $\pm$ 12.2 & 47.6 $\pm$ 13.4 & 53.8 $\pm$ 10.0 & \textbf{60.8} $\pm$ 8.9 & 53.6 $\pm$ 10.0 & 56.8 $\pm$ 11.6 & 61.2 $\pm$ 6.9 & 55.4 $\pm$ 6.7 & 61.6 $\pm$ 7.5 & \underline{73.5} $\pm$ 9.4 & \underline{68.1} $\pm$ 12.5 & 40.5 $\pm$ 10.5 & \textbf{90.7} $\pm$ 4.5 \\
& F1  & 55.4 $\pm$ 7.5 & 66.3 $\pm$ 8.1 & 48.2 $\pm$ 7.9 & 53.2 $\pm$ 6.9 & \textbf{57.2} $\pm$ 6.9 & 51.3 $\pm$ 6.4 & 50.4 $\pm$ 6.2 & 58.7 $\pm$ 5.4 & 53.9 $\pm$ 5.5 & 58.1 $\pm$ 5.4 & \textbf{65.0} $\pm$ 7.9 & 59.4 $\pm$ 10.8 & 42.8 $\pm$ 5.9 & \underline{82.2} $\pm$ 5.9 \\

\midrule

\multirow{3}{*}{\xcgm} 
& PR  & \underline{63.6} $\pm$ 7.2 & 86.6 $\pm$ 6.9 & 29.8 $\pm$ 10.3 & 55.3 $\pm$ 7.4 & 66.9 $\pm$ 7.4 & 35.5 $\pm$ 8.1 & 16.9 $\pm$ 6.2 & 67.4 $\pm$ 3.8 & 60.0 $\pm$ 6.5 & 61.9 $\pm$ 7.7 & 59.3 $\pm$ 11.9 & 56.2 $\pm$ 14.2 & 17.5 $\pm$ 9.4 & \underline{87.7} $\pm$ 5.9 \\
& AUC & 76.6 $\pm$ 5.8 & 73.6 $\pm$ 11.5 & 48.0 $\pm$ 13.2 & 53.2 $\pm$ 10.1 & \underline{58.4} $\pm$ 9.0 & 52.1 $\pm$ 10.9 & 55.7 $\pm$ 13.7 & 61.8 $\pm$ 6.1 & 56.4 $\pm$ 7.1 & 62.0 $\pm$ 7.0 & 73.0 $\pm$ 9.2 & 67.7 $\pm$ 12.4 & 40.0 $\pm$ 9.4 & \underline{90.6} $\pm$ 4.6 \\
& F1  & 56.5 $\pm$ 7.6 & 65.1 $\pm$ 8.0 & 48.2 $\pm$ 7.9 & 52.5 $\pm$ 7.1 & \underline{55.4} $\pm$ 7.0 & 49.9 $\pm$ 7.4 & 49.0 $\pm$ 6.7 & 59.1 $\pm$ 5.0 & 54.4 $\pm$ 5.4 & 58.1 $\pm$ 5.3 & \underline{64.8} $\pm$ 7.5 & 59.2 $\pm$ 10.4 & 42.8 $\pm$ 5.2 & 82.1 $\pm$ 5.3 \\

\midrule

\multirow{3}{*}{\begin{tabular}{@{}l@{}}\gluformer \\ (tiny) \end{tabular}} 
& PR  & 59.4 $\pm$ 5.5 & 86.1 $\pm$ 6.7 & 28.2 $\pm$ 8.8 & 60.2 $\pm$ 7.6 & 58.1 $\pm$ 3.8 & 33.9 $\pm$ 5.0 & 17.9 $\pm$ 7.5 & 74.3 $\pm$ 6.1 & 63.3 $\pm$ 6.0 & 64.5 $\pm$ 5.0 & 48.9 $\pm$ 11.0 & 50.9 $\pm$ 9.4 & 21.6 $\pm$ 7.2 & 75.4 $\pm$ 7.0 \\
& AUC & 73.9 $\pm$ 4.6 & 72.5 $\pm$ 11.8 & 47.9 $\pm$ 11.4 & 57.9 $\pm$ 8.3 & 47.6 $\pm$ 4.7 & 51.1 $\pm$ 7.7 & 53.3 $\pm$ 13.3 & 68.9 $\pm$ 6.3 & 61.9 $\pm$ 7.2 & 65.4 $\pm$ 5.0 & 63.3 $\pm$ 9.4 & 63.6 $\pm$ 8.2 & 58.0 $\pm$ 12.4 & 81.7 $\pm$ 4.8 \\
& F1  & 51.7 $\pm$ 5.7 & 65.4 $\pm$ 8.1 & 48.0 $\pm$ 6.2 & 55.4 $\pm$ 6.1 & 47.8 $\pm$ 4.2 & 50.3 $\pm$ 5.9 & 49.1 $\pm$ 6.0 & 63.0 $\pm$ 5.4 & 58.5 $\pm$ 5.9 & 61.7 $\pm$ 3.8 & 57.9 $\pm$ 7.6 & 59.4 $\pm$ 7.3 & \underline{53.1} $\pm$ 6.6 & 72.7 $\pm$ 5.1 \\

\midrule

\multirow{3}{*}{\begin{tabular}{@{}l@{}}\gluformer \\ (base) \end{tabular}} 
& PR  & 62.9 $\pm$ 4.6 & 87.2 $\pm$ 6.2 & 31.4 $\pm$ 8.4 & 61.4 $\pm$ 6.5 & 56.5 $\pm$ 4.1 & 31.5 $\pm$ 5.4 & 11.5 $\pm$ 1.9 & 69.8 $\pm$ 6.1 & 61.1 $\pm$ 5.1 & 56.1 $\pm$ 4.4 & 51.9 $\pm$ 12.2 & 51.6 $\pm$ 10.5 & 16.4 $\pm$ 6.0 & 82.7 $\pm$ 7.8 \\
& AUC & 76.7 $\pm$ 3.7 & 74.8 $\pm$ 11.3 & 53.1 $\pm$ 10.0 & 60.5 $\pm$ 7.9 & 44.9 $\pm$ 5.8 & 46.7 $\pm$ 8.8 & 43.9 $\pm$ 10.9 & 63.1 $\pm$ 6.7 & 58.0 $\pm$ 5.0 & 56.3 $\pm$ 4.5 & 66.0 $\pm$ 9.9 & 61.6 $\pm$ 9.0 & 46.9 $\pm$ 9.9 & 87.9 $\pm$ 5.1 \\
& F1  & 55.7 $\pm$ 5.4 & 67.8 $\pm$ 7.3 & \underline{52.5} $\pm$ 6.0 & 57.2 $\pm$ 6.2 & 46.3 $\pm$ 4.7 & 48.1 $\pm$ 5.8 & 44.8 $\pm$ 2.6 & 59.2 $\pm$ 5.2 & 55.4 $\pm$ 3.9 & 54.8 $\pm$ 3.8 & 58.7 $\pm$ 8.7 & 58.3 $\pm$ 7.7 & 47.7 $\pm$ 4.9 & 80.7 $\pm$ 5.6 \\

\midrule
\midrule

\multirow{3}{*}{\textbf{\name}} 
& PR  & \textbf{65.9} $\pm$ 7.5 & \textbf{91.9} $\pm$ 5.3 & \textbf{36.1} $\pm$ 11.2 & \underline{64.9} $\pm$ 11.7 & \underline{67.0} $\pm$ 7.7 & 33.5 $\pm$ 7.8 & 21.1 $\pm$ 10.0 & \textbf{77.3} $\pm$ 7.5 & \textbf{69.0} $\pm$ 9.6 & \textbf{67.6} $\pm$ 8.1 & \textbf{66.2} $\pm$ 13.0 & \textbf{60.2} $\pm$ 15.1 & 14.4 $\pm$ 5.2 & \textbf{88.3} $\pm$ 5.7 \\
& AUC & \textbf{78.7} $\pm$ 6.1 & \textbf{81.2} $\pm$ 12.7 & \underline{54.7} $\pm$ 15.9 & \underline{62.6} $\pm$ 13.2 & 57.8 $\pm$ 8.2 & 50.5 $\pm$ 13.4 & \underline{59.2} $\pm$ 15.9 & \textbf{72.8} $\pm$ 7.4 & \textbf{68.7} $\pm$ 8.9 & \textbf{69.1} $\pm$ 6.6 & \textbf{75.9} $\pm$ 10.2 & \textbf{70.7} $\pm$ 12.2 & 41.6 $\pm$ 11.6 & \textbf{90.7} $\pm$ 4.3 \\
& F1  & \textbf{58.3} $\pm$ 8.5 & \underline{69.6} $\pm$ 11.0 & 50.2 $\pm$ 9.7 & \underline{59.4} $\pm$ 9.9 & \underline{55.4} $\pm$ 6.4 & 49.1 $\pm$ 9.1 & 50.7 $\pm$ 6.9 & \textbf{66.2} $\pm$ 6.6 & \textbf{63.3} $\pm$ 6.9 & \textbf{64.0} $\pm$ 5.8 & 64.5 $\pm$ 9.2 & \textbf{62.0} $\pm$ 10.4 & 43.1 $\pm$ 4.9 & \textbf{82.4} $\pm$ 5.6 \\

\bottomrule
\end{tabular}
}
\end{table*}

%% file: tables/appendix_table_linear_probe_significance_overall.tex
\begin{table*}[!t]
    \centering
    \caption{Paired task-averaged linear-probing comparisons. Differences are \name minus the comparator in percentage points. CIs and $p$-values use the Nadeau--Bengio correction; Holm adjustment is applied across the six model--metric comparisons.}
    \label{tab:linear_probe_significance_overall}
    \tablestyle{5pt}{1.08}
    \begin{tabular}{@{}llccc@{}}
        \toprule
        Comparator & Metric & Task average (GlucoFM / comparator) & $\Delta$ [95\% CI] & Holm-$p$ \\
        \midrule
        \multirow{3}{*}{\cgmjepa}
        & PR-AUC   & 58.8 / 54.7 & $+4.11$ [$2.40$, $5.81$] & $<0.001$ \\
        & ROC-AUC  & 66.7 / 62.4 & $+4.34$ [$2.26$, $6.41$] & $<0.001$ \\
        & Macro-F1 & 59.9 / 57.3 & $+2.57$ [$0.91$, $4.23$] & $0.006$ \\
        \midrule
        \multirow{3}{*}{\cgmformer}
        & PR-AUC   & 58.8 / 54.1 & $+4.74$ [$2.47$, $7.02$] & $<0.001$ \\
        & ROC-AUC  & 66.7 / 62.1 & $+4.67$ [$2.02$, $7.32$] & $0.003$ \\
        & Macro-F1 & 59.9 / 57.5 & $+2.34$ [$0.55$, $4.14$] & $0.012$ \\
        \bottomrule
    \end{tabular}
\end{table*}

%% file: tables/appendix_table_ppgr_by_device.tex
\begin{table*}[!t]
    \centering
    \caption{Two-hour PPGR with complete context at each sensor's native cadence (MAE$\downarrow$).}
    \label{tab:ppgr_by_device}
    \tablestyle{4pt}{1.02}\scriptsize
    \begin{tabular*}{\textwidth}{@{\extracolsep{\fill}}llcccc@{}}
        \toprule
        Sensor & Model & Trajectory (mg/dL) & Positive iAUC (mg/dL$\cdot$h) & Peak rise (mg/dL) & Peak time (min) \\
        \midrule
        \multirow{6}{*}{Dexcom}
        & \gluformer~(tiny) & 25.38 & 34.98 & 30.15 & 31.37 \\
        & \gluformer~(base) & 26.51 & 36.89 & 31.65 & 32.71 \\
        & \cgmformer & 24.71 & 33.11 & 28.86 & 31.70 \\
        & \cgmjepa & \underline{23.60} & \underline{31.27} & \underline{27.76} & \underline{30.74} \\
        & \xcgm & 24.74 & 33.14 & 29.11 & 31.59 \\
        & \texttt{\bfseries GlucoFM} & \textbf{22.63} & \textbf{30.09} & \textbf{26.92} & \textbf{30.17} \\
        \addlinespace[2pt]\midrule
        \multirow{6}{*}{Libre}
        & \gluformer~(tiny) & 23.08 & 30.14 & 26.04 & 32.25 \\
        & \gluformer~(base) & 24.12 & 32.46 & 27.74 & 32.54 \\
        & \cgmformer & 23.59 & 31.22 & 26.99 & 32.43 \\
        & \cgmjepa & 22.21 & 29.21 & 24.19 & 32.01 \\
        & \xcgm & \underline{21.46} & \underline{27.99} & \underline{24.06} & \underline{31.34} \\
        & \texttt{\bfseries GlucoFM} & \textbf{21.12} & \textbf{26.91} & \textbf{23.87} & \textbf{30.46} \\
        \bottomrule
    \end{tabular*}
\end{table*}

%% file: tables/appendix_table_few_shot_summary.tex
\begin{table*}[!t]
    \centering
    \caption{Task-averaged few-shot performance. Each cell reports PR-AUC / ROC-AUC / Macro-F1 (\%).}
    \label{tab:few_shot_summary}
    \tablestyle{3.5pt}{1.02}\scriptsize
    \begin{tabular*}{\textwidth}{@{\extracolsep{\fill}}llcccc@{}}
        \toprule
        Setting & Budget & \cgmjepa & \gluformer~(tiny) & \cgmformer & \texttt{\bfseries GlucoFM} \\
        \midrule
        \multirow{5}{*}{Limited subjects}
        & $K=1$ & 50.2 / 57.0 / 49.3 & 51.6 / 58.8 / 50.2 & 50.4 / 57.5 / 49.7 & \textbf{53.3 / 59.6 / 50.4} \\
        & $K=2$ & 50.8 / 58.3 / 51.8 & 51.9 / 59.5 / 52.4 & 51.3 / 58.8 / 52.2 & \textbf{53.9 / 60.7 / 53.3} \\
        & $K=3$ & 51.1 / 58.7 / 52.7 & 51.9 / 59.7 / 53.1 & 51.8 / 59.2 / 53.1 & \textbf{54.5 / 61.5 / 54.4} \\
        & $K=4$ & 51.5 / 59.1 / 53.2 & 51.9 / 60.0 / 53.6 & 52.1 / 59.7 / 53.5 & \textbf{55.2 / 62.2 / 55.2} \\
        & $K=5$ & 52.1 / 59.8 / 53.9 & 52.0 / 60.3 / 54.0 & 52.5 / 60.2 / 54.2 & \textbf{55.6 / 62.8 / 56.0} \\
        \addlinespace[2pt]\midrule
        \multirow{7}{*}{Limited observations}
        & $1\%$  & 50.1 / 58.4 / 54.2 & 52.9 / 61.6 / 56.5 & 51.8 / 59.8 / 55.6 & \textbf{56.3 / 64.0 / 57.9} \\
        & $5\%$  & 50.1 / 58.3 / 54.4 & 52.6 / 61.2 / 56.2 & 51.8 / 59.8 / 55.6 & \textbf{56.5 / 64.2 / 58.1} \\
        & $10\%$ & 50.9 / 59.2 / 54.9 & 52.7 / 61.2 / 56.2 & 52.2 / 60.3 / 55.9 & \textbf{57.2 / 65.0 / 58.6} \\
        & $20\%$ & 51.8 / 60.0 / 55.6 & 52.2 / 60.7 / 56.0 & 52.9 / 60.9 / 56.4 & \textbf{57.4 / 65.3 / 59.0} \\
        & $30\%$ & 52.9 / 60.7 / 56.1 & 52.1 / 60.7 / 55.9 & 53.1 / 61.1 / 56.5 & \textbf{58.1 / 65.8 / 59.5} \\
        & $40\%$ & 53.5 / 61.2 / 56.4 & 52.3 / 61.0 / 56.0 & 53.2 / 61.3 / 56.8 & \textbf{58.2 / 66.0 / 59.5} \\
        & $50\%$ & 54.0 / 61.7 / 56.7 & 52.3 / 61.0 / 56.1 & 53.3 / 61.3 / 56.8 & \textbf{58.5 / 66.3 / 59.6} \\
        \bottomrule
    \end{tabular*}
\end{table*}

%% file: tables/appendix_table_multiday_exp.tex
\begin{table*}[!t]
\centering
\caption{
Subject-level multiday PR-AUC across observation lengths. Values are mean PR-AUC (\%) with 95\% confidence intervals over repeated subject-level cross-validation. Mean+Max denotes concat(mean, max) pooling. The best value in each row is bolded. \hall is evaluated only for $K=1$--$4$.
}
\label{tab:multiday_pooling_summary}
\scriptsize
\setlength{\tabcolsep}{2.4pt}
\renewcommand{\arraystretch}{1.05}
\resizebox{\textwidth}{!}{%
\begin{tabular}{@{}lllccccccc@{}}
\toprule
Dataset& Task & Pooling & K1 & K2 & K3 & K4 & K5 & K6 & K7 \\
\midrule
\cgmacros-Dexcom & Diabetes & Mean
& 57.6 [57.0, 58.2] & 58.5 [57.8, 59.4] & 58.2 [57.4, 59.0] & 58.5 [57.6, 59.4] & 58.5 [57.5, 59.4] & 58.6 [57.3, 59.7] & \textbf{59.8} [58.5, 60.9] \\

\cgmacros-Dexcom & Diabetes & Mean+Max
& 57.0 [56.3, 57.7] & 57.9 [57.0, 58.7] & 57.1 [55.7, 58.4] & 58.0 [56.2, 59.8] & 57.3 [54.9, 59.6] & 57.3 [55.1, 59.3] & \textbf{59.5} [57.5, 61.7] \\

\cgmacros-Dexcom & IR & Mean
& 92.5 [91.7, 93.2] & 93.3 [92.3, 94.0] & 94.4 [93.3, 95.2] & 94.9 [93.8, 95.7] & 95.3 [94.2, 96.1] & \textbf{95.6} [94.5, 96.5] & \textbf{95.6} [94.7, 96.3] \\

\cgmacros-Dexcom & IR & Mean+Max
& 92.3 [91.4, 92.9] & 92.5 [91.5, 93.2] & 93.3 [92.0, 94.3] & 93.8 [92.3, 94.9] & 94.6 [93.1, 95.7] & 95.5 [94.1, 96.6] & \textbf{97.8} [97.2, 98.3] \\

\midrule
\cgmacros-Libre & Diabetes & Mean
& 63.7 [62.1, 65.1] & 65.5 [63.5, 67.2] & 66.7 [64.5, 68.3] & 67.0 [64.7, 68.8] & \textbf{67.8} [65.5, 69.6] & 67.5 [65.0, 69.5] & 67.4 [64.9, 69.6] \\

\cgmacros-Libre & Diabetes & Mean+Max
& 62.9 [61.1, 64.4] & 63.9 [61.7, 65.6] & 63.3 [61.1, 64.9] & 63.3 [60.8, 65.0] & \textbf{65.7} [62.9, 68.0] & 63.3 [60.3, 65.9] & 64.5 [61.2, 67.5] \\

\cgmacros-Libre & IR & Mean
& 90.6 [90.1, 91.1] & 91.8 [91.5, 92.2] & 92.1 [91.8, 92.5] & 92.1 [91.8, 92.5] & 92.4 [92.0, 92.8] & 92.5 [92.1, 92.9] & \textbf{92.7} [92.3, 93.1] \\

\cgmacros-Libre & IR & Mean+Max
& 89.7 [88.9, 90.3] & 91.6 [91.2, 92.0] & 91.5 [91.0, 91.9] & 91.5 [91.1, 92.1] & 92.5 [92.1, 93.0] & \textbf{92.7} [92.3, 93.1] & 92.2 [91.8, 92.7] \\

\midrule
\cgmacros-Fused & Diabetes & Mean
& 64.0 [63.0, 65.1] & 65.6 [64.4, 66.7] & 66.3 [65.0, 67.5] & 67.0 [65.6, 68.3] & \textbf{67.6} [66.1, 68.9] & 67.4 [65.7, 69.1] & 67.0 [65.0, 68.8] \\

\cgmacros-Fused & Diabetes & Mean+Max
& 63.2 [62.2, 64.3] & 64.0 [62.8, 65.3] & 64.2 [62.6, 65.7] & 65.7 [63.7, 67.6] & \textbf{66.1} [63.9, 68.3] & 63.9 [61.6, 66.1] & 63.6 [61.6, 65.5] \\

\cgmacros-Fused & IR & Mean
& 92.7 [92.4, 93.1] & 93.2 [92.7, 93.8] & 93.6 [92.9, 94.2] & 93.8 [93.1, 94.6] & 94.2 [93.5, 94.9] & 94.4 [93.7, 95.1] & \textbf{94.6} [93.9, 95.2] \\

\cgmacros-Fused & IR & Mean+Max
& 92.3 [91.9, 92.7] & 93.5 [92.9, 94.0] & 93.5 [92.9, 94.0] & 93.7 [93.1, 94.2] & 93.9 [93.2, 94.6] & 94.4 [93.7, 95.0] & \textbf{95.0} [93.7, 95.8] \\

\midrule
\shanghai & IR & Mean
& \textbf{61.9} [60.7, 63.2] & \textbf{61.9} [59.9, 63.7] & 61.0 [58.4, 63.4] & 61.7 [58.9, 64.4] & 61.2 [58.3, 64.1] & 61.4 [58.4, 64.4] & \textbf{61.9} [59.5, 64.3] \\

\shanghai & IR & Mean+Max
& 63.1 [61.8, 64.4] & 62.5 [60.6, 64.3] & 60.8 [58.4, 63.3] & 62.1 [59.6, 64.6] & 67.8 [65.3, 70.3] & \textbf{75.2} [72.7, 77.6] & \textbf{75.2} [73.6, 76.7] \\

\midrule
\stanford  & Beta-cell & Mean
& 60.3 [59.2, 61.3] & 61.8 [60.5, 63.0] & 64.0 [62.5, 65.4] & 65.4 [63.8, 66.9] & 66.8 [65.0, 68.5] & 68.3 [65.7, 70.7] & \textbf{69.9} [67.1, 72.4] \\

\stanford  & Beta-cell & Mean+Max
& 59.7 [58.8, 60.5] & 63.0 [62.2, 63.9] & 63.7 [62.4, 65.0] & 64.8 [63.7, 66.0] & 63.6 [62.0, 65.1] & 63.5 [61.5, 65.5] & \textbf{65.8} [63.2, 68.3] \\

\stanford & Diabetes & Mean
& 67.6 [66.1, 69.1] & 69.8 [68.3, 71.5] & 70.6 [69.0, 72.3] & 70.3 [68.2, 72.5] & 70.5 [67.9, 73.4] & 71.8 [68.7, 74.9] & \textbf{73.6} [71.2, 75.9] \\

\stanford  & Diabetes & Mean+Max
& 67.9 [66.3, 69.6] & 67.7 [66.1, 69.4] & 68.3 [66.5, 70.1] & 66.9 [64.4, 69.6] & 68.0 [65.1, 71.1] & 69.5 [66.6, 72.5] & \textbf{71.7} [69.5, 74.2] \\

\stanford  & IR & Mean
& 60.3 [58.7, 61.9] & 61.5 [59.4, 63.5] & 62.4 [60.5, 64.2] & 62.9 [60.9, 64.9] & 64.1 [62.1, 66.1] & 65.9 [64.0, 68.0] & \textbf{66.6} [64.6, 68.9] \\

\stanford & IR & Mean+Max
& 60.1 [58.6, 61.4] & 61.8 [60.0, 63.7] & 63.3 [61.4, 65.0] & \textbf{66.2} [64.3, 68.1] & 63.9 [62.6, 65.3] & 64.0 [62.9, 65.1] & 65.3 [63.6, 66.8] \\

\midrule
\hall & Diabetes & Mean
& 59.8 [58.7, 60.9] & 62.8 [60.0, 65.5] & 67.7 [64.9, 70.5] & \textbf{73.8} [71.6, 75.9] & -- & -- & -- \\

\hall  & Diabetes & Mean+Max
& 57.8 [56.8, 58.7] & 65.4 [62.7, 67.5] & 63.6 [60.0, 67.1] & \textbf{70.9} [67.8, 74.1] & -- & -- & -- \\

\hall  & Glucotype & Mean
& 85.9 [84.9, 86.8] & 91.6 [90.9, 92.3] & 96.0 [95.4, 96.5] & \textbf{99.5} [98.9, 99.9] & -- & -- & -- \\

\hall  & Glucotype & Mean+Max
& 85.0 [83.9, 86.0] & 90.9 [90.0, 91.9] & 93.7 [92.5, 94.8] & \textbf{98.8} [98.0, 99.4] & -- & -- & -- \\

\hall  & IR & Mean
& 46.3 [44.2, 48.4] & 44.5 [42.3, 46.5] & 46.3 [42.8, 49.5] & \textbf{49.8} [46.0, 53.0] & -- & -- & -- \\

\hall  & IR & Mean+Max
& 45.0 [42.9, 47.2] & 44.9 [42.8, 46.8] & 41.5 [38.9, 44.2] & \textbf{45.1} [42.1, 48.0] & -- & -- & -- \\
\bottomrule
\end{tabular}%
}
\end{table*}

%% file: tables/appendix_table_encoder_design.tex
\begin{table*}[!t]
\centering
\caption{Performance comparison of different encoder designs. All reported values represent the mean $\pm$ std evaluated via a 10-iteration 5-fold cross-validation.}
\label{apd:encoder_design}
\resizebox{\textwidth}{!}{
\begin{tabular}{l|l | >{\columncolor{gray!10}}c>{\columncolor{gray!10}}c>{\columncolor{gray!10}}c>{\columncolor{gray!10}}c ccc >{\columncolor{gray!10}}c>{\columncolor{gray!10}}c>{\columncolor{gray!10}}c cccc}
\toprule

\multirow{2}{*}{\textbf{Encoder Design}} & \multirow{2}{*}{\textbf{Metrics}} &
\multicolumn{4}{c}{\cellcolor{gray!10}\textbf{\cgmacros}} & 
\multicolumn{3}{c}{\textbf{\shanghai}} & 
\multicolumn{3}{c}{\cellcolor{gray!10}\textbf{\stanford}} & 
\multicolumn{4}{c}{\textbf{\hall}} \\

\cmidrule(lr){3-6} \cmidrule(lr){7-9} \cmidrule(lr){10-12} \cmidrule(lr){13-16}

& &  
\rotatebox{0}{Diabetes} & \rotatebox{0}{IR} & \rotatebox{0}{Hyperlip.} & \rotatebox{0}{Obesity} & 
\rotatebox{0}{IR} & \rotatebox{0}{Hyperlip.} & \rotatebox{0}{Hypogly.} & 
\rotatebox{0}{Diabetes} & \rotatebox{0}{$\beta$-cell Dys.} & \rotatebox{0}{IR} & 
\rotatebox{0}{Diabetes} & \rotatebox{0}{IR} & \rotatebox{0}{Hyperlip.} & \rotatebox{0}{Glucotype} \\

\midrule
\midrule

\multirow{3}{*}{Raw Input} 
& PR  & 61.4 $\pm$ 6.0 & 89.4 $\pm$ 7.0 & 31.1 $\pm$ 9.8 & \textbf{66.0} $\pm$ 8.9 & 60.9 $\pm$ 6.8 & \underline{34.5} $\pm$ 7.0 & \textbf{23.3} $\pm$ 11.1 & \textbf{77.3} $\pm$ 6.7 & \underline{64.4} $\pm$ 8.1 & \underline{63.8} $\pm$ 7.7 & \underline{55.7} $\pm$ 12.9 & 53.2 $\pm$ 10.6 & 16.1 $\pm$ 4.9 & 77.0 $\pm$ 9.9 \\
& AUC & 75.4 $\pm$ 4.9 & 79.0 $\pm$ 13.8 & 49.9 $\pm$ 14.6 & \textbf{64.5} $\pm$ 9.4 & 50.8 $\pm$ 8.4 & 50.1 $\pm$ 10.3 & \textbf{63.2} $\pm$ 16.0 & \textbf{73.0} $\pm$ 7.4 & \underline{64.0} $\pm$ 8.4 & \underline{65.0} $\pm$ 6.7 & 68.1 $\pm$ 10.4 & 63.1 $\pm$ 10.6 & 43.3 $\pm$ 11.6 & 82.1 $\pm$ 7.0 \\
& F1  & \underline{55.3} $\pm$ 5.9 & \underline{70.6} $\pm$ 10.3 & 48.8 $\pm$ 7.6 & \textbf{60.5} $\pm$ 7.1 & 50.5 $\pm$ 6.4 & \textbf{49.7} $\pm$ 6.5 & \textbf{56.3} $\pm$ 8.7 & \textbf{66.4} $\pm$ 6.6 & \underline{59.7} $\pm$ 6.8 & \underline{60.1} $\pm$ 4.5 & 60.1 $\pm$ 7.9 & 58.0 $\pm$ 7.3 & 46.6 $\pm$ 5.8 & 73.4 $\pm$ 6.4 \\

\midrule

\multirow{3}{*}{State-stream Only} 
& PR  & \underline{62.8} $\pm$ 6.7 & \underline{91.0} $\pm$ 5.6 & \underline{34.5} $\pm$ 12.9 & 62.6 $\pm$ 9.1 & 59.2 $\pm$ 6.1 & 33.2 $\pm$ 8.4 & 18.2 $\pm$ 7.1 & 74.5 $\pm$ 4.7 & 62.7 $\pm$ 7.9 & 62.2 $\pm$ 6.9 & 54.1 $\pm$ 11.3 & \underline{58.9} $\pm$ 12.0 & \underline{18.5} $\pm$ 5.1 & \underline{86.2} $\pm$ 6.4 \\
& AUC & \underline{76.0} $\pm$ 5.6 & \underline{80.0} $\pm$ 12.7 & \underline{51.0} $\pm$ 16.7 & 60.0 $\pm$ 11.0 & 47.8 $\pm$ 7.8 & 48.5 $\pm$ 13.2 & 55.1 $\pm$ 12.6 & 69.5 $\pm$ 6.7 & 60.8 $\pm$ 8.9 & 62.7 $\pm$ 5.1 & \underline{68.5} $\pm$ 8.7 & \underline{69.0} $\pm$ 11.0 & \textbf{54.3} $\pm$ 13.5 & \underline{89.2} $\pm$ 4.4 \\
& F1  & 54.1 $\pm$ 7.0 & \textbf{70.7} $\pm$ 10.7 & \underline{50.0} $\pm$ 8.3 & 56.6 $\pm$ 8.0 & 48.4 $\pm$ 6.6 & 48.7 $\pm$ 8.5 & 50.5 $\pm$ 6.6 & 63.9 $\pm$ 5.7 & 57.2 $\pm$ 7.0 & 59.5 $\pm$ 4.0 & \underline{61.3} $\pm$ 7.4 & \textbf{62.7} $\pm$ 9.0 & \underline{47.3} $\pm$ 6.7 & \underline{80.2} $\pm$ 6.0 \\
\midrule

\multirow{3}{*}{Event-stream Only} 
& PR  & 60.1 $\pm$ 4.8 & 84.7 $\pm$ 9.7 & 28.6 $\pm$ 8.2 & 60.5 $\pm$ 7.4 & \underline{62.2} $\pm$ 6.5 & \textbf{34.7} $\pm$ 5.9 & 13.9 $\pm$ 5.8 & \underline{76.9} $\pm$ 5.9 & 62.0 $\pm$ 8.1 & 57.2 $\pm$ 6.5 & 53.1 $\pm$ 13.2 & 52.0 $\pm$ 11.5 & \textbf{19.4} $\pm$ 7.7 & 59.4 $\pm$ 9.7 \\
& AUC & 74.5 $\pm$ 4.4 & 72.0 $\pm$ 16.1 & 46.9 $\pm$ 11.1 & 58.7 $\pm$ 9.9 & \underline{53.5} $\pm$ 8.2 & \textbf{51.3} $\pm$ 9.3 & 46.3 $\pm$ 9.8 & 72.3 $\pm$ 7.0 & 61.9 $\pm$ 7.8 & 60.5 $\pm$ 7.2 & 62.3 $\pm$ 12.4 & 62.6 $\pm$ 9.4 & \underline{52.6} $\pm$ 11.4 & 65.9 $\pm$ 9.8 \\
& F1  & 54.5 $\pm$ 5.9 & 65.8 $\pm$ 10.0 & 46.8 $\pm$ 6.6 & 56.2 $\pm$ 8.0 & \underline{53.0} $\pm$ 5.6 & \underline{49.2} $\pm$ 6.6 & 46.4 $\pm$ 4.7 & 65.3 $\pm$ 6.4 & 56.4 $\pm$ 6.6 & 57.3 $\pm$ 4.6 & 57.4 $\pm$ 7.5 & 57.7 $\pm$ 6.9 & \textbf{49.4} $\pm$ 7.4 & 62.6 $\pm$ 7.5 \\

\midrule
\multirow{3}{*}{\textbf{Dual-stream}} 
& PR  & \textbf{65.9} $\pm$ 7.5 & \textbf{91.9} $\pm$ 5.3 & \textbf{36.1} $\pm$ 11.2 & \underline{64.9} $\pm$ 11.7 & \textbf{67.0} $\pm$ 7.7 & 33.5 $\pm$ 7.8 & \underline{21.1} $\pm$ 10.0 & \textbf{77.3} $\pm$ 7.5 & \textbf{69.0} $\pm$ 9.6 & \textbf{67.6} $\pm$ 8.1 & \textbf{66.2} $\pm$ 13.0 & \textbf{60.2} $\pm$ 15.1 & 14.4 $\pm$ 5.2 & \textbf{88.3} $\pm$ 5.7 \\
& AUC & \textbf{78.7} $\pm$ 6.1 & \textbf{81.2} $\pm$ 12.7 & \textbf{54.7} $\pm$ 15.9 & \underline{62.6} $\pm$ 13.2 & \textbf{57.8} $\pm$ 8.2 & \underline{50.5} $\pm$ 13.4 & \underline{59.2} $\pm$ 15.9 & \underline{72.8} $\pm$ 7.4 & \textbf{68.7} $\pm$ 8.9 & \textbf{69.1} $\pm$ 6.6 & \textbf{75.9} $\pm$ 10.2 & \textbf{70.7} $\pm$ 12.2 & 41.6 $\pm$ 11.6 & \textbf{90.7} $\pm$ 4.3 \\
& F1  & \textbf{58.3} $\pm$ 8.5 & 69.6 $\pm$ 11.0 & \textbf{50.2} $\pm$ 9.7 & \underline{59.4} $\pm$ 9.9 & \textbf{55.4} $\pm$ 6.4 & 49.1 $\pm$ 9.1 & \underline{50.7} $\pm$ 6.9 & \underline{66.2} $\pm$ 6.6 & \textbf{63.3} $\pm$ 6.9 & \textbf{64.0} $\pm$ 5.8 & \textbf{64.5} $\pm$ 9.2 & \underline{62.0} $\pm$ 10.4 & 43.1 $\pm$ 4.9 & \textbf{82.4} $\pm$ 5.6 \\

\bottomrule
\end{tabular}
}
\end{table*}

%% file: tables/appendix_table_ablation_summary.tex
\begin{table*}[!t]
\centering
\caption{Task-averaged results for supporting ablations.}
\label{tab:ablation_summary}
\tablestyle{5.5pt}{1.02}\small
\begin{tabular*}{\textwidth}{@{\extracolsep{\fill}}llccc@{}}
\toprule
Ablation & Setting & PR-AUC & ROC-AUC & Macro-F1 \\
\midrule
\multirow{11}{*}{Temporal dynamics weight}
& $\lambda_{\mathrm{TD}}=0.0$ & 55.5 & 63.2 & 57.7 \\
& $\lambda_{\mathrm{TD}}=0.2$ & 57.1 & 65.7 & 59.2 \\
& $\lambda_{\mathrm{TD}}=0.4$ & 58.2 & 66.5 & 59.5 \\
& $\lambda_{\mathrm{TD}}=0.6$ & 58.9 & 67.0 & 59.9 \\
& $\lambda_{\mathrm{TD}}=0.8$ & 58.9 & 67.0 & 60.0 \\
& $\lambda_{\mathrm{TD}}=1.0$ (default) & 58.8 & 66.7 & 59.9 \\
& $\lambda_{\mathrm{TD}}=1.2$ & 58.6 & 66.5 & 59.7 \\
& $\lambda_{\mathrm{TD}}=1.4$ & 58.4 & 66.4 & 59.6 \\
& $\lambda_{\mathrm{TD}}=1.6$ & 58.3 & 66.2 & 59.4 \\
& $\lambda_{\mathrm{TD}}=1.8$ & 58.1 & 66.1 & 59.2 \\
& $\lambda_{\mathrm{TD}}=2.0$ & 57.9 & 66.0 & 59.2 \\
\midrule
\multirow{4}{*}{Data augmentation}
& None & 56.8 & 65.0 & 58.6 \\
& Value perturbation & 57.4 & 65.6 & 59.0 \\
& Structural sparsification & 58.4 & 66.5 & 59.7 \\
& Full (default) & 58.8 & 66.7 & 59.9 \\
\midrule
\multirow{3}{*}{Interpolation}
& Dense interpolation & 56.7 & 64.4 & 58.2 \\
& Dense + sparsification & 57.9 & 66.2 & 59.2 \\
& Mask-aware (default) & 58.8 & 66.7 & 59.9 \\
\bottomrule
\end{tabular*}

\end{table*}